%% file: MAIN_DC.tex
\documentclass[a4paper,fleqn]{cas-dc}

\usepackage[numbers]{natbib}
\usepackage{subcaption}
\usepackage{siunitx}
\usepackage{caption}

\usepackage{siunitx}
\def\tsc#1{\csdef{#1}{\textsc{\lowercase{#1}}\xspace}}
\tsc{WGM}
\tsc{QE}
\tsc{EP}
\tsc{PMS}
\tsc{BEC}
\tsc{DE}
\begin{document}

\let\WriteBookmarks\relax
\def\floatpagepagefraction{1}
\def\textpagefraction{.001}
\shorttitle{FlowLet}
\shortauthors{D. Danese~et al.}

\title [mode = title]{FlowLet: Conditional 3D Brain MRI Synthesis using Wavelet Flow Matching}                      

\author[1]{Danilo Danese}
\ead{danilo.danese@poliba.it}
\cormark[1]

\author[1]{Angela Lombardi}
\ead{angela.lombardi@poliba.it}
\cormark[1]

\author[1, 2]{Matteo Attimonelli}
\ead{matteo.attimonelli@poliba.it}

\author[1]{Giuseppe Fasano}
\ead{giuseppe.fasano@poliba.it}

\author[1]{Tommaso Di~Noia}
\ead{tommaso.dinoia@poliba.it}

\affiliation[1]{organization={Politecnico di Bari},
                city={Bari},
                country={Italy}}

\affiliation[2]{organization={Sapienza University of Rome},
                city={Rome},
                country={Italy}}

\cortext[cor1]{Corresponding author}

\begin{abstract}
Generative modeling for 3D brain MRI is challenged by a trade-off between anatomical fidelity, sample diversity, and computational efficiency. Diffusion-based approaches achieve strong visual quality but typically require hundreds to thousands of sampling steps, while latent-space compression can introduce reconstruction artifacts and degrade fine-grained anatomy.
We introduce \emph{FlowLet}, a conditional generative framework that performs Flow Matching in an invertible 3D wavelet domain. This representation enables multi-scale generation without learned latent compression, while deterministic ODE sampling allows fast inference. Age conditioning is modeled through complementary feature-wise modulation and spatially adaptive cross-attention, enabling explicit control over age-related morphological variation.
Across multi-site neuroimaging datasets, FlowLet achieves competitive and, in several settings, superior global fidelity compared to diffusion-based baselines using as few as 10 sampling steps. Region-based evaluation across 95 cortical and subcortical brain regions demonstrates improved local anatomical plausibility beyond what is captured by global similarity metrics alone. In a downstream brain age prediction study, models augmented with FlowLet-generated data consistently reduce prediction error relative to real-only training and other generative baselines.
Rather than focusing on a single dominant metric improvement, these results highlight a consistent trade-off between efficiency, controllability, and anatomically meaningful 3D brain MRI generation.
The proposed framework is released as open-source to support reproducibility.
\end{abstract}

% \includegraphics[width=\textwidth]{FlowLet_Graphical_abstract.pdf}

% \begin{highlights}
% \item Wavelet-based flow matching model synthesizes 3D brain MRI
% \item Age conditioning controls localized morphology changes in synthetic volumes
% \item Deterministic sampling allows generation in 10 steps without additional latent compression
% \item Region-and task-based tests show gains in brain age prediction accuracy
% \item Methodology, code, and evaluation protocols are released free and open-source for reproducibility
% \end{highlights}


\maketitle

\input{corpus}

%%%%%%%%%%%%%%
\appendix

\input{supplementary_corpus}

\input{acknowledgments}

\bibliographystyle{cas-model2-names}

\bibliography{cas-refs}

\end{document}

%% file: corpus.tex
\begin{abstract}
Brain Magnetic Resonance Imaging (MRI) plays a central role in studying neurological development, aging, and diseases. One key application is Brain Age Prediction (BAP), which estimates an individual’s biological brain age from MRI data. Effective BAP models require large, diverse, and age‑balanced datasets, whereas existing 3D MRI datasets are demographically skewed, limiting fairness and generalizability. Acquiring new data is costly and ethically constrained, motivating generative data augmentation. Current generative methods are often based on latent diffusion models, which operate in learned low dimensional latent spaces to address the memory demands of volumetric MRI data. However, these methods are typically slow at inference, may introduce artifacts due to latent compression, and are rarely conditioned on age, thereby affecting the BAP performance. In this work, we propose FlowLet, a conditional generative framework that synthesizes age‑conditioned 3D MRIs by leveraging flow matching within an invertible 3D wavelet domain, helping to avoid reconstruction artifacts and reducing computational demands. Experiments show that FlowLet generates high‑fidelity volumes with few sampling steps. Training BAP models with data generated by FlowLet improves performance for underrepresented age groups, and region‑based analysis confirms preservation of anatomical structures.
\end{abstract}

\section{Introduction}

Magnetic Resonance Imaging (MRI) provides a non-invasive, high-resolution representation of brain anatomy and is routinely used to study structural changes associated with development, aging, and neurological disorders \cite{bacon2024neuroimage}. In recent years, learning-based analysis of brain MRI has enabled quantitative biomarkers that capture subtle morphological variations beyond traditional volumetric measures \cite{rahman2025understanding, fu2025synthesizing}.

A prominent example is \emph{Brain Age Prediction} (BAP), a data-driven framework that estimates an individual’s biological brain age from structural MRI by learning population-level associations between anatomical features and chronological age \cite{cole2017predicting, baecker2021machine, lombardi2021explainable, peng2021accurate}. The discrepancy between predicted and chronological age, commonly referred to as the brain-age gap, has been linked to cognitive decline, neurodegenerative disease, and altered aging trajectories, making BAP a clinically relevant biomarker \cite{gaser2024perspective}.

The performance and reliability of BAP models strongly depend on the availability of large, diverse, and age-balanced MRI datasets spanning the full lifespan. However, in practice, publicly available 3D brain MRI datasets exhibit substantial demographic imbalances. Certain age ranges, particularly young and middle-aged adults, are overrepresented, while pediatric and elderly populations are often under-sampled \cite{bashyam2020mri}. Although large-scale cohorts exist \cite{Sudlow2015-fd}, restricted access and acquisition costs limit their widespread use. These data limitations reduce generalizability, introduce age-related biases, and hinder the deployment of BAP models in clinical and epidemiological settings \cite{dinsdale2021learning}.

Given the cost, logistical complexity, and ethical constraints associated with acquiring additional MRI data, synthetic data generation has emerged as a complementary strategy for dataset enrichment. In this context, \emph{generative modeling} aims to learn the underlying distribution of brain MRIs and synthesize anatomically plausible volumes that can be used for augmentation, bias mitigation, and robustness improvement \cite{Chintapalli2024-hy}. When properly designed, generative models offer a scalable mechanism to increase data diversity, particularly for underrepresented age groups.

However, generating realistic 3D brain MRIs remains technically challenging. Volumetric neuroimaging data are characterized by extremely high dimensionality and complex multi-scale anatomical structure \cite{fernandez2024generating, yu2025hifi}. State-of-the-art generative approaches, including autoencoding architectures and diffusion models, often face trade-offs between computational efficiency and anatomical fidelity when applied to full-resolution 3D volumes \cite{DBLP:conf/nips/DhariwalN21}. To alleviate computational burden, many methods rely on learned latent representations, which can introduce reconstruction artifacts and obscure fine-grained anatomical details that are critical for age-related analysis \cite{Muller-Franzes2023-ju}.

Beyond realism, effective use of synthetic MRI in BAP requires \emph{controlled generation}. In particular, models must be conditioned on age information such that synthesized volumes reflect meaningful and localized morphological changes associated with aging \cite{peng2024metadata}. Designing conditioning mechanisms that influence generation at multiple spatial scales while preserving anatomical coherence remains an open problem. Insufficient or overly coarse conditioning often results in weak age specificity or entangled anatomical effects, limiting downstream utility.

These challenges are closely related to a fundamental limitation in modern generative modeling, commonly referred to as the \emph{generative modeling trilemma} \cite{DBLP:conf/iclr/XiaoKV22}. The trilemma formalizes the trade-off between sample quality, diversity, and sampling efficiency, stating that improving one objective typically degrades at least one of the others. In the context of medical imaging, where high anatomical fidelity, population-level variability, and practical inference times are all required, mitigating this trade-off is particularly critical.

In this work, we introduce \emph{FlowLet}, a conditional generative framework for 3D brain MRI synthesis designed to address these challenges. FlowLet combines Flow Matching with an invertible wavelet-based representation, enabling multi-scale modeling of volumetric brain anatomy without relying on learned latent compression. This design supports anatomically structured generation while remaining computationally tractable. Age information is incorporated through complementary feature-wise and spatial conditioning mechanisms, providing explicit control over age-related morphological variations. By leveraging Flow Matching, FlowLet further supports efficient sample generation and offers a practical alternative to the diffusion-based baseline implementations considered in this study.

We evaluate FlowLet on three neuroimaging datasets spanning over 12 multi-site sources to assess robustness to acquisition variability and demographic heterogeneity. Beyond standard global similarity metrics (FID, MMD, MS-SSIM), which are known to be limited in volumetric neuroimaging due to background dominance, we develop a comprehensive evaluation protocol. This includes (i) a downstream BAP task to assess functional utility as a clinically meaningful biomarker, and (ii) region-based anatomical metrics across 95 brain regions to quantify localized structural fidelity. This combined evaluation reveals cases where favorable global metrics fail to capture anatomically relevant errors.
To support reproducibility and practical adoption, we release the complete open-source implementation of FlowLet.

Our main contributions are:
\begin{enumerate}
    \item A publicly available wavelet-based Flow Matching framework for efficient and anatomically faithful 3D brain MRI synthesis;
    \item A multi-level age-conditioning design that combines feature-wise and spatial mechanisms for localized morphological control;
    \item A region- and task-aware evaluation protocol for generative brain MRI, integrating anatomical fidelity and downstream brain age prediction.
\end{enumerate}

\section{Related Work} \label{Related}
\subsection{Augmentation}
Despite an increase in general-purpose data availability, 3D neuroimaging remains constrained by small cohorts and high-dimensional voxel data, limiting population diversity and statistical reliability \cite{powerfailure}. Deep learning models, which require large and diverse datasets to generalize well, are particularly affected. Data augmentation is thus essential for improving model robustness by artificially expanding training sets. Traditional methods such as adding noise, cropping, flipping or elastic transform, preserve labels but limit variability and, more importantly, they raise the risk of introducing distorted anatomical structures or amplify biases \cite{medicaldareview,DBLP:journals/jbd/ShortenK19}. These risks are especially pronounced in medical imaging, where anatomical fidelity is critical and validation often relies on expert assessment. Moreover, class imbalance remains a persistent challenge: oversampling strategies, though common, scale poorly in high-dimensional spaces and may fail to capture minority class variation \cite{DBLP:journals/ijkesdp/Nguyenck11, DBLP:journals/bmcbi/BlagusL13a}. These limitations require specialised methods, i.e., synthetic data generation.

\subsection{Generative models for 3D synthesis}
Generative modeling has become a key approach for synthetic data generation in medical imaging, enabling the synthesis of anatomically plausible samples beyond simple geometric or intensity-based augmentations. Early deep generative models, such as Generative Adversarial Networks (GANs)~\cite{DBLP:conf/nips/GoodfellowPMXWOCB14} and Variational Autoencoders (VAEs)~\cite{DBLP:journals/corr/Kingmaw13}, laid the foundation for data-driven image synthesis. GANs are known for producing sharp samples but often suffer from training instability and mode collapse \cite{DBLP:conf/nips/SalimansGZCRCC16}, while VAEs offer stable training and efficient sampling at the expense of overly smooth reconstructions \cite{DBLP:journals/corr/Kingmaw13}. These limitations become particularly pronounced when extending such models to high-resolution 3D neuroimaging data.

Denoising Diffusion Models (DDMs)~\cite{DBLP:conf/nips/HoJA20} have recently achieved state-of-the-art performance across a wide range of imaging domains, including medical and neuroimaging applications \cite{DBLP:conf/cvpr/WyattLSW22,pinaya2022brain,DBLP:conf/midl/WuFFZYXL023,DBLP:conf/midl/DurrerWBSWSGYC23}. These models formulate generation as the reversal of a gradual noising process, enabling high-quality and diverse sample synthesis. However, diffusion-based generation typically requires solving stochastic differential equations through hundreds or thousands of iterative steps \cite{DBLP:conf/iclr/SongME21,DBLP:conf/iccv/DavtyanSF23}, resulting in substantial computational overhead. This iterative sampling becomes a major bottleneck when scaling to high-resolution 3D volumes.

To improve sampling efficiency, several approaches operate in compressed or discrete latent spaces. Vector-Quantized VAEs (VQ-VAEs)~\cite{DBLP:conf/nips/OordVK17} and Latent Diffusion Models (LDMs)~\cite{DBLP:conf/cvpr/RombachBLEO22} reduce computational costs by performing diffusion in a lower-dimensional representation. While effective in improving efficiency and global coherence, these methods introduce additional sources of complexity, such as learned compression artifacts or increased training overhead, which may degrade fine anatomical details. Extensions targeting ultra-high-resolution synthesis rely on specialized scheduling and hierarchical strategies \cite{DBLP:conf/cvpr/ZhangHLG025}, further increasing system complexity.

Flow Matching (FM) has recently emerged as an alternative paradigm for generative modeling that addresses the inefficiency of diffusion-based sampling \cite{DBLP:conf/iclr/LipmanCBNL23,DBLP:conf/iclr/AlbergoV23,DBLP:conf/iclr/LiuG023,DBLP:conf/icml/NeklyudovB0M23}. Instead of learning a stochastic denoising process, FM models a continuous-time velocity field that transports samples from a simple prior to the target data distribution by solving an ordinary differential equation. By encouraging straighter transport trajectories \cite{DBLP:conf/icml/LeeKY23}, FM substantially reduces the number of inference steps required for sample generation, making it particularly attractive for high-resolution volumetric synthesis. Recent works have further explored conditional control and efficiency improvements within the FM framework \cite{gagneux2025visual}.

Wavelet Diffusion Models (WDMs) provide a learning-free alternative to latent diffusion for dimensionality reduction in generative modeling \cite{DBLP:conf/cvpr/PhungDT23,DBLP:conf/miccai/FriedrichWBDC24,DBLP:journals/corr/abs-2503-18352}. By applying a fixed wavelet transform, WDMs decompose images into multi-resolution frequency components and perform diffusion directly in the wavelet domain. This approach significantly reduces memory requirements while preserving spatial structure and avoiding artifacts introduced by learned compression.

Despite these advantages, existing 3D WDM formulations~\cite{DBLP:conf/miccai/FriedrichWBDC24} considered here rely on iterative diffusion sampling, in which inference requires hundreds to thousands of sequential denoising evaluations along the reverse trajectory. Each evaluation carries the full computational cost of a forward pass through a 3D U-Net, making the cumulative inference cost a significant bottleneck in high-resolution volumetric settings. The integration of more efficient generative paradigms, such as Flow Matching, within a wavelet-based representation remains largely unexplored. A conceptually related direction combining wavelets with normalizing flows has been investigated for high-resolution 2D image synthesis \cite{DBLP:conf/nips/YuDB20}, but extensions to conditional 3D neuroimaging remain limited.

Building on the literature discussed above, FlowLet addresses several open gaps in generative modeling for 3D neuroimaging. While diffusion-based models achieve high sample quality, their iterative sampling limits scalability to high-resolution volumes. Latent and wavelet-based approaches mitigate computational costs but often remain tied to diffusion sampling or introduce compression artifacts. In parallel, Flow Matching offers efficient generation but has seen limited adoption in volumetric medical imaging and has not been explored in conjunction with wavelet representations. FlowLet bridges these directions by integrating Flow Matching within an invertible wavelet domain, enabling efficient, anatomically faithful, and controllable 3D brain MRI synthesis. In addition, FlowLet explicitly targets age-conditioned generation and introduces a task- and region-aware evaluation protocol, addressing limitations in both conditioning strategies and assessment practices in prior work.

\section{Proposed method} 

\subsection{System overview}
\begin{figure*}[t!]
  \centering
  \includegraphics[width=\textwidth]{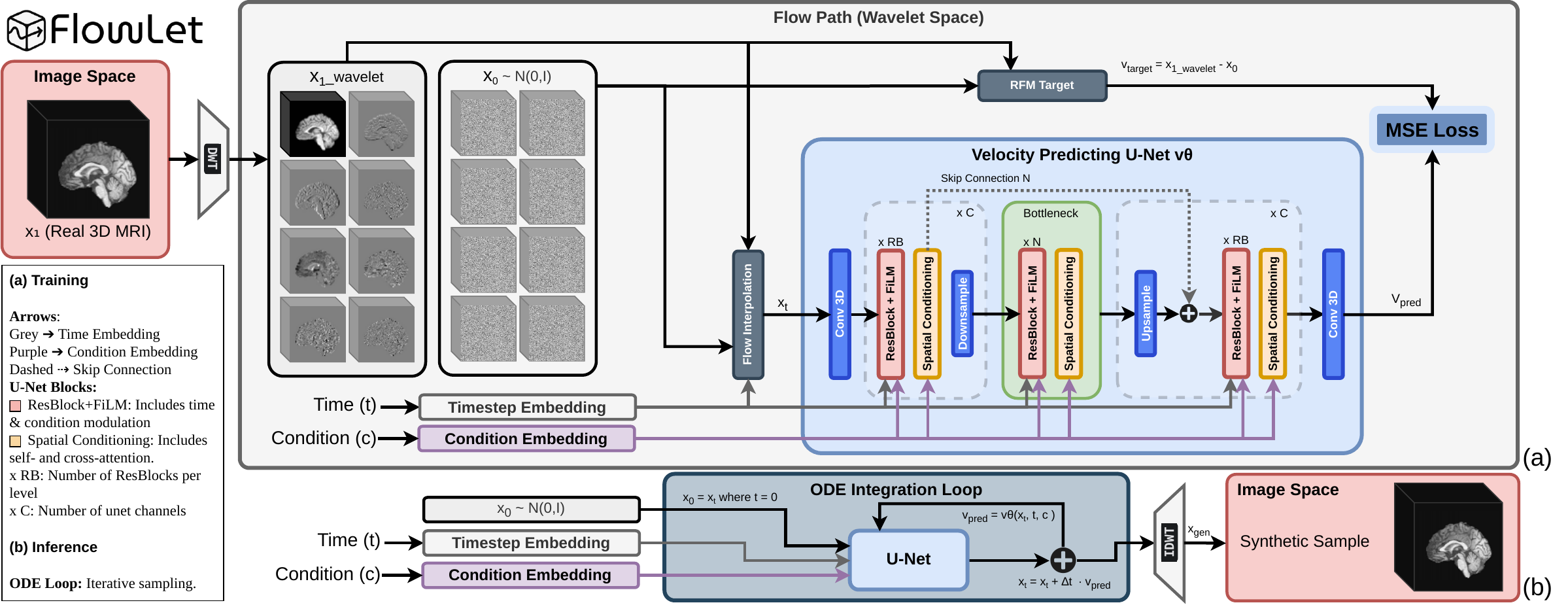}

\caption{\textbf{FlowLet overview.} \textbf{(a) Training:} a preprocessed 3D MRI volume $x$ is mapped to the invertible 3D Haar wavelet domain via the DWT, producing eight subbands (LLL + seven detail bands). The conditional 3D U-Net $v_\theta(x_t, t, c)$ (condition $c$: age) is trained to predict the FM velocity field in wavelet space along the interpolation path $x_t$ between noise $x_0$ and data $x_1$. \textbf{(b) Inference:} starting from Gaussian noise $x_0$ in wavelet space and a target condition $c$, samples are generated by integrating the learned ODE $\partial_t x_t = v_\theta(x_t, t, c)$ with a small number of solver steps; the final wavelet coefficients are mapped back to voxel space with the inverse DWT (IDWT) to obtain the synthesized MRI.}
  \label{fig:FlowletArchitecture}
\end{figure*}

FlowLet is a conditional generative framework for 3D brain MRI synthesis that operates entirely in the wavelet domain. The overall, single-stage architecture pipeline illustrated in Figure~\ref{fig:FlowletArchitecture}, consists of three main blocks: (i) wavelet-based decomposition of volumetric MRI data, (ii) flow matching–based generative modeling in the multi-scale frequency domain, and (iii) reconstruction of the synthesized volume via inverse wavelet transform.

During training, each input MRI volume is first transformed using an invertible 3D Haar discrete wavelet transform (DWT), yielding one low-frequency subband capturing coarse anatomical structure and seven high-frequency subbands encoding localized fine details. A conditional neural network is then trained to predict continuous-time velocity fields that transport samples from Gaussian noise to the target data distribution in this wavelet space. Conditioning variables, such as age, are injected into the network to explicitly control age-dependent morphological characteristics during generation.

At inference time, synthesis begins from Gaussian noise in the wavelet domain and proceeds by solving an ordinary differential equation defined by the learned velocity field. The resulting wavelet coefficients are finally mapped back to the voxel domain through the inverse DWT, producing a full-resolution 3D brain MRI. This design enables efficient sampling, multi-scale anatomical modeling, and explicit control over clinically relevant attributes, while avoiding the artifacts and overhead associated with learned latent compression.

\subsection{Preliminaries}
\label{Wavelet_background}

To achieve tractable learning while maintaining anatomical fidelity, the orthonormal Haar Discrete Wavelet Transform (DWT)~\cite{DBLP:books/daglib/0098272} is adopted. This perfectly invertible transform decomposes 3D volumes into frequency components, reducing dimensionality while preserving structural information up to numerical precision~\cite{BULLMORE2004S234}.

Given a 3D volume $x \in \mathbb{R}^{D \times H \times W}$, where $D$, $H$, and $W$ denote the three spatial dimensions of the volume, the Haar DWT applies 1D low-pass and high-pass filters sequentially along each axis. Following the preprocessing convention adopted in this work, which aligns all volumes to a common reference space, these axes correspond to the anterior–posterior, superior–inferior, and left–right directions, respectively.
Specifically, the Haar analysis filters are
\begin{equation}
\mathbf{L} = \frac{1}{\sqrt{2}}\begin{bmatrix}1 & 1\end{bmatrix},
\qquad
\mathbf{H} = \frac{1}{\sqrt{2}}\begin{bmatrix}-1 & 1\end{bmatrix}.
\end{equation}
This produces eight frequency subbands,
\begin{equation}
\mathcal{B} = \{\mathrm{LLL}, \mathrm{LLH}, \mathrm{LHL}, \mathrm{LHH}, \mathrm{HLL}, \mathrm{HLH}, \mathrm{HHL}, \mathrm{HHH}\},
\end{equation}
where each letter indicates whether a Low- or High-pass filter is applied along the corresponding axis. The subbands are then concatenated into an 8-channel tensor,
\begin{equation}
 x_w = \mathcal{W}(x)
 \in \mathbb{R}^{8 \times \frac{D}{2} \times \frac{H}{2} \times \frac{W}{2}},
\end{equation}
where $\mathcal{W}$ denotes the forward DWT. The transform is perfectly invertible via the inverse DWT,
\begin{equation}
IDWT = \mathcal{W}^{-1}(x_w) = x.
\end{equation}
Due to the orthonormality of the Haar basis (Parseval’s theorem for orthonormal wavelets~\cite{DBLP:journals/pami/Mallat89, DBLP:books/siam/92/D1992}), energy preservation and perfect reconstruction are guaranteed across subbands.

\subsection{FlowLet framework}
FlowLet builds upon the wavelet representation by explicitly modeling generation in the multi-scale frequency domain. As shown in Figure~\ref{fig:FlowletArchitecture} (left), the transformed 3D MRI volume is factorized into a low-frequency approximation subband (LLL) and seven high-frequency detail subbands (LLH, LHL, ..., HHH). The LLL subband captures the dominant anatomical structure, while the remaining components isolate fine-grained, spatially localized details. As the signal energy is predominantly concentrated in the LLL subband and total energy is preserved across subbands by Parseval’s theorem, interpolation trajectories in this coarse subspace are inherently smoother, supporting stable learning dynamics and reconstruction~\cite{DBLP:conf/iclr/LipmanCBNL23}. At the same time, neural networks can specialize across subbands, enhancing stability and convergence in generative frameworks~\cite{DBLP:conf/nips/HoJA20, DBLP:conf/nips/VahdatKK21}. This frequency-based factorization also reduces global variability and overfitting by promoting spatial locality~\cite{DBLP:books/daglib/0098272}. 
In practice, this representation allows the generative network to operate jointly over the eight wavelet subbands at reduced spatial resolution, yielding a compact multi-scale parameterization for volumetric synthesis. Although this reduces spatial resolution, the transform is invertible and no information is discarded, as the eight subbands jointly encode the original signal. In addition, the effect of this representation on low- and high-frequency components is explicitly analyzed in Appendix \ref{sec:lll_only_ablation_appendix} and \ref{sec:high_frequency_wavelet_analysis_appendix}.
Code is available at \url{https://github.com/sisinflab/FlowLet}.

\subsubsection{Flow Matching formulations}
The framework supports multiple flow matching formulations to enable flexible and adaptable modeling, all operating in the wavelet domain, where samples \(x_t\), noise \(x_0\), data \(x_1\), and velocity fields \(v\) (the instantaneous time-derivative \(\partial_t x_t\) along the path) belong to \(\mathbb{R}^{8 \times \frac{D}{2} \times \frac{H}{2} \times \frac{W}{2}}\). The data sample \(x_1\) is the wavelet transform \(\mathcal{W}(x_1^{\mathrm{voxel}})\) of an original sample \(x_1^{\mathrm{voxel}} \sim p_{\mathrm{data}}\), and noise \(x_0\) is standard Gaussian in wavelet space. FlowLet supports a modular family of FM strategies that define different continuous-time interpolation paths \(x_t\) from noise to data in the wavelet domain. Each variant specifies a target velocity field \(v_{\mathrm{target}}(x_t, t)\). The model learns a parameterized velocity network \( v_\theta(x_t, t, c) \), conditioned on signal \( c \), by minimizing the MSE loss:
{\small
\begin{equation}
\mathcal{L}_{\mathrm{FM}} = \mathbb{E}_{x_t, t, v_{\mathrm{target}}} \left[ \left\| v_\theta(x_t, t, c) - v_{\mathrm{target}}(x_t, t) \right\|^2 \right].
\end{equation}
} 
The learned velocity field governs a deterministic ordinary differential equation (ODE) on \( t \in [0,1] \), solved via Euler integration. 

The selected formulations, chosen for their foundational role in the FM literature and their progressively increasing trajectory curvature, are ordered to reflect differences in the geometry of the interpolation path, an aspect known to influence the expressiveness and stability of the learned velocity field~\cite{DBLP:conf/icml/LeeKY23}. This setup enables a systematic evaluation of how curvature impacts training stability and synthesis quality. An overview of the implemented FM variants is provided.

\paragraph{Rectified Flow Matching (RFM).}
RFM~\cite{DBLP:conf/iclr/LiuG023} performs linear interpolation between a Gaussian noise sample and a data sample in the wavelet domain:
{\small
\begin{equation}
x_t = (1 - t) x_0 + t x_1, ~v_{\mathrm{target}} = x_1 - x_0.
\end{equation}}The target velocity field is constant along the straight line path, resulting in zero path curvature, which promotes stable training and yields low-variance gradient estimates.

\paragraph{Conditional Flow Matching (CFM).} 
CFM~\cite{DBLP:conf/iclr/LipmanCBNL23, DBLP:conf/iclr/AlbergoV23} constructs a linear path between a data sample \(x_1\) and a sampled noise \(x_0\) in wavelet space for each training instance. The target velocity field is then defined as the instantaneous direction from \(x_t\) to \(x_1\), scaled by the remaining time:
{\small
\begin{equation}
x_t = (1 - t) x_0 + t x_1, ~ v_{\mathrm{target}} = \frac{x_1 - x_t}{1 - t + \epsilon},
\end{equation}
}where a small \( \epsilon > 0 \) is added to prevent divergence as \( t \to 1 \). While the underlying path is a straight line, the target velocity field is explicitly dependent on the current state \(x_t\) and time \(t\), making it more dynamic than the constant velocity of RFM. This introduces non-zero curvature and increases sensitivity as \(t\) approaches \(1\).

\paragraph{Variance-Preserving Diffusion Matching (VP).}
Inspired by Variance-Preserving diffusion and the corresponding deterministic probability-flow formulation~\cite{DBLP:conf/iclr/0011SKKEP21, DBLP:conf/iclr/SongME21}, VP defines a nonlinear interpolation from data to noise governed by a linear variance schedule \(\beta(t) = \beta_{\min} + t(\beta_{\max} - \beta_{\min})\), \(t \in [0, 1]\). Signal and noise scaling coefficients are:
{\small
\begin{equation}
\bar{\alpha}(t) = \exp\left(-\frac{1}{2} \int_0^t \beta(s) \, ds\right), \quad \sigma(t) = \sqrt{1 - \bar{\alpha}(t)^2}.
\end{equation}
}The forward noising process generates intermediate samples \(x_t\) via interpolation between a data sample \(x_1\) and standard Gaussian noise \(\xi \sim \mathcal{N}(0, \mathbf{I})\), while the corresponding target velocity field \(v_{\mathrm{target}}(x_t, t)\), governing the reverse-time dynamics, is defined by the gradient (score) of the marginal distribution \(\nabla_{x_t} \log p_t(x_t)\):
{\small
\begin{equation}
x_t = \bar{\alpha}(t)\, x_1 + \sigma(t)\, \xi, ~
v_{\mathrm{target}}(x_t, t) = -\frac{1}{2}\beta(t)\, x_t - \beta(t)\,\nabla_{x_t} \log p_t(x_t).
\end{equation}
}This nonlinear velocity field leads to curved reverse trajectories characteristic of diffusion models. A small positive constant is typically added to denominators during training for numerical stability.

\paragraph{Trigonometric Flow.} 
Trigonometric~\cite{DBLP:conf/icml/NicholD21} uses a circular interpolation on the unit half-circle in wavelet space:
{\small
\begin{equation}
x_t = \cos\left(\frac{\pi}{2} t\right) x_0 + \sin\left(\frac{\pi}{2} t\right) x_1,
\end{equation}
with time derivative velocity
\begin{equation}
v_{\mathrm{target}} = \frac{\pi}{2} \left[- \sin\left(\frac{\pi}{2} t\right) x_0 + \cos\left(\frac{\pi}{2} t\right) x_1 \right].
\end{equation}
}This formulation maintains constant norm \(\|x_t\|\) and has constant curvature  $\frac{\pi^2}{4}$, introducing smooth curved trajectories with stable, non-straight, velocity fields.

Implementation-level details for each FM variant (including the exact code correspondence for $x_t$ and $v_{\text{target}}$) are provided in Appendix \ref{A_fm_implementation}, while Appendix \ref{A_VP_derivation} gives the full VP derivation linking the SDE to the deterministic probability-flow ODE and the exact conditional target velocity used in training.

\subsubsection{Conditional U-Net architecture}
\label{U-NetArchitecture}

FlowLet employs a conditional 3D U-Net, \( v_\theta \), designed to predict the velocity field within the 8-channel wavelet domain, as illustrated in Figure~\ref{fig:FlowletArchitecture}(a). The model \emph{input} consists of interpolated wavelet coefficients
$x_t \in \mathbb{R}^{8 \times \frac{D}{2} \times \frac{H}{2} \times \frac{W}{2}}$,
a timestep $t$, and a conditioning vector $c$; the \emph{output} is the predicted velocity field $v_{\mathrm{pred}}$ in the same wavelet domain, which is integrated via Euler ODE and mapped back to the volume domain through the inverse discrete wavelet transform (IDWT).

The U-Net backbone follows encoder, bottleneck and decoder stages with skip connections, enabling hierarchical feature extraction for 3D data. The primary building blocks are residual blocks (\emph{ResBlocks}) incorporating normalization layers and SiLU activations. Temporal conditioning is intrinsic to the flow matching formulation: the timestep $t$ is embedded via sinusoidal positional encoding~\cite{DBLP:conf/nips/VaswaniSPUJGKP17}, followed by a multi-layer perceptron (MLP), yielding a time embedding $e_{\mathrm{time}}$ that conditions feature computation throughout the network.

\subsubsection{Unified conditional embedding}
Beyond temporal conditioning, FlowLet is explicitly designed to synthesize brain MRIs according to clinically relevant attributes, with age being the focus of this work. Scalar conditioning variables are first normalized to a consistent numerical range; in the case of age, values are scaled to $[0,1]$ based on the minimum and maximum observed in the training data. Each normalized scalar is then projected into a high-dimensional latent space using a dedicated two-layer MLP with SiLU activations, transforming the 1-dimensional normalized input into a 512-dimensional vector. When multiple conditioning variables are present, their projected representations are combined via element-wise summation, yielding a unified conditioning embedding $e_{\mathrm{cond}}$.

This unified representation is injected into the U-Net through two complementary mechanisms: depth-wise feature modulation and spatially adaptive attention.

\subsubsection{Depth-wise conditioning via FiLM}
To ensure the conditioning information influences feature computation at every level of the network, we employ Feature-wise Linear Modulation (FiLM)~\cite{DBLP:conf/aaai/PerezSVDC18} within every residual block of the U-Net. FiLM applies an affine transformation to intermediate feature maps, allowing the network to dynamically adjust the scale and bias of activations on a per-instance basis. In our model, both the flow matching timestep embedding ($e_{\mathrm{time}}$) and the condition embedding ($e_{\mathrm{cond}}$) generate independent scale ($\gamma$) and bias ($b$) parameters. Specifically, given intermediate features $h$, conditioning is applied sequentially as:
\begin{align}
h' &= \mathrm{Norm}(h) \cdot (1 + \gamma_{\mathrm{time}}) + b_{\mathrm{time}}, \label{eq:film_time} \\
h_{\mathrm{mod}} &= h' \cdot (1 + \gamma_{\mathrm{cond}}) + b_{\mathrm{cond}}. \label{eq:film_cond}
\end{align}
The first transformation (\ref{eq:film_time}) adapts features according to the current point along the flow trajectory, while the second (\ref{eq:film_cond}) refines them based on the conditioning variable. By integrating FiLM into every residual block, the conditional signal is made pervasively available, influencing both low-level texture formation and high-level anatomical structure across encoder and decoder paths.

\subsubsection{Spatially adaptive conditioning via cross-attention}
While FiLM enables robust global modulation, it applies identical transformations across all spatial locations within a feature channel. To capture localized, region-specific anatomical changes associated with aging, FlowLet further incorporates spatial conditioning modules inspired by transformer architectures~\cite{DBLP:conf/cvpr/RombachBLEO22}. These modules are inserted at lower spatial resolutions of the U-Net, where feature maps encode abstract, semantically meaningful representations with large receptive fields.

Each spatial conditioning block first applies self-attention to model long-range spatial dependencies within the feature map. This is followed by a cross-attention operation, in which spatial features act as queries and the unified conditioning embedding provides both keys and values:
\begin{align}
h_{\mathrm{attn}} &= \mathrm{Attention}(Q, K, V), \\
Q &= W_Q\, h, \quad
K = W_K\, e_{\mathrm{cond}}, \quad
V = W_V\, e_{\mathrm{cond}}.
\end{align}
This mechanism enables the network to selectively modulate specific anatomical regions as a function of age, refining spatially localized morphological patterns such as ventricular enlargement or cortical thinning.

While FiLM and cross-attention are individually well-established conditioning 
techniques, their roles in this framework are complementary and specifically motivated by the 
wavelet-domain formulation. Because the network input consists of eight subbands encoding 
distinct frequency components, FiLM provides channel-wise modulation that can differentially 
scale and shift each subband as a function of the conditioning variable, effectively controlling 
the relative emphasis on low- versus high-frequency content at each layer. Cross-attention 
complements this by introducing spatially selective conditioning: age-related morphological 
changes (e.g., ventricular enlargement, cortical thinning) are localized phenomena, and 
cross-attention allows the network to attend to different spatial regions depending on the 
conditioning input. The combination thus provides both frequency-selective and spatially-selective 
control, which we argue is particularly relevant when operating on a multi-band wavelet 
representation rather than on single-channel image data.

\subsubsection{Inference}
Sampling, illustrated in Figure~\ref{fig:FlowletArchitecture}(b), is performed by solving an ordinary differential equation in the wavelet domain, starting from Gaussian noise $x_0 \sim \mathcal{N}(0, I)$. Wavelet coefficients are iteratively updated using the learned velocity field $v_{\mathrm{pred}}$. After integration, the final MRI volume is reconstructed by applying the IDWT. Additional training and inference benchmarks are provided in Appendix~\ref{Efficiency_benchmarks}.

\section{Datasets and preprocessing}
\label{sec:datasets}
This section describes the multi-site datasets and the unified preprocessing pipeline used to build a full range age-spanning cohort. This setup aims to provide consistent training inputs and to enable fair evaluation of both synthesis quality and downstream BAP performance.

\label{sec:materials}
The training cohort was constructed by integrating T1-weighted MRI scans from three publicly available datasets: OpenBHB\footnote{\url{https://baobablab.github.io/bhb/dataset}}, ADNI\footnote{\url{https://adni.loni.usc.edu/}}, and OASIS-3\footnote{\url{https://sites.wustl.edu/oasisbrains/}}.
Only cognitively normal subjects were retained. This integration was motivated by the pronounced age imbalance of OpenBHB, which predominantly represents younger populations, and by the need to ensure adequate coverage of older age ranges for age-conditioned synthesis and downstream BAP evaluation.

An additional external dataset (DLBS) is used for evaluation purposes and is described in Appendix \ref{sec:dlbs_dataset_appendix}.

\subsection{OpenBHB}
OpenBHB aggregates T1-weighted MRIs from 10 publicly available datasets (e.g., IXI, ABIDE I/II, GSP, CoRR), spanning 62 imaging sites across North America, Europe, and Asia~\cite{dufumier2022openbhb}. The mean age of the subjects is $24.92 \pm 14.29$ years, with a strong concentration around early adulthood. OpenBHB provides a predefined training--validation split obtained via stratified sampling based on age, sex, and acquisition site; in this work, the predefined validation subset was reserved exclusively for testing downstream tasks.

\subsection{ADNI}
From the Alzheimer's Disease Neuroimaging Initiative (ADNI-1, ADNI-2, and ADNI-3), 769 T1-weighted scans were selected from cognitively normal individuals aged 60–91 years (mean: $76.97 \pm 4.99$). To avoid repeated measures, only one scan per subject was retained, and all individuals with any cognitive impairment were excluded.

\subsection{OASIS-3}
The Open Access Series of Imaging Studies (OASIS-3) is a longitudinal neuroimaging dataset spanning the adult lifespan. From an initial pool of 1,314 scans, 1,041 T1-weighted MRIs of cognitively normal subjects were retained after filtering, covering ages 42–95 years (mean: $71.10 \pm 8.93$).

\begin{center}
\includegraphics[width=\columnwidth]{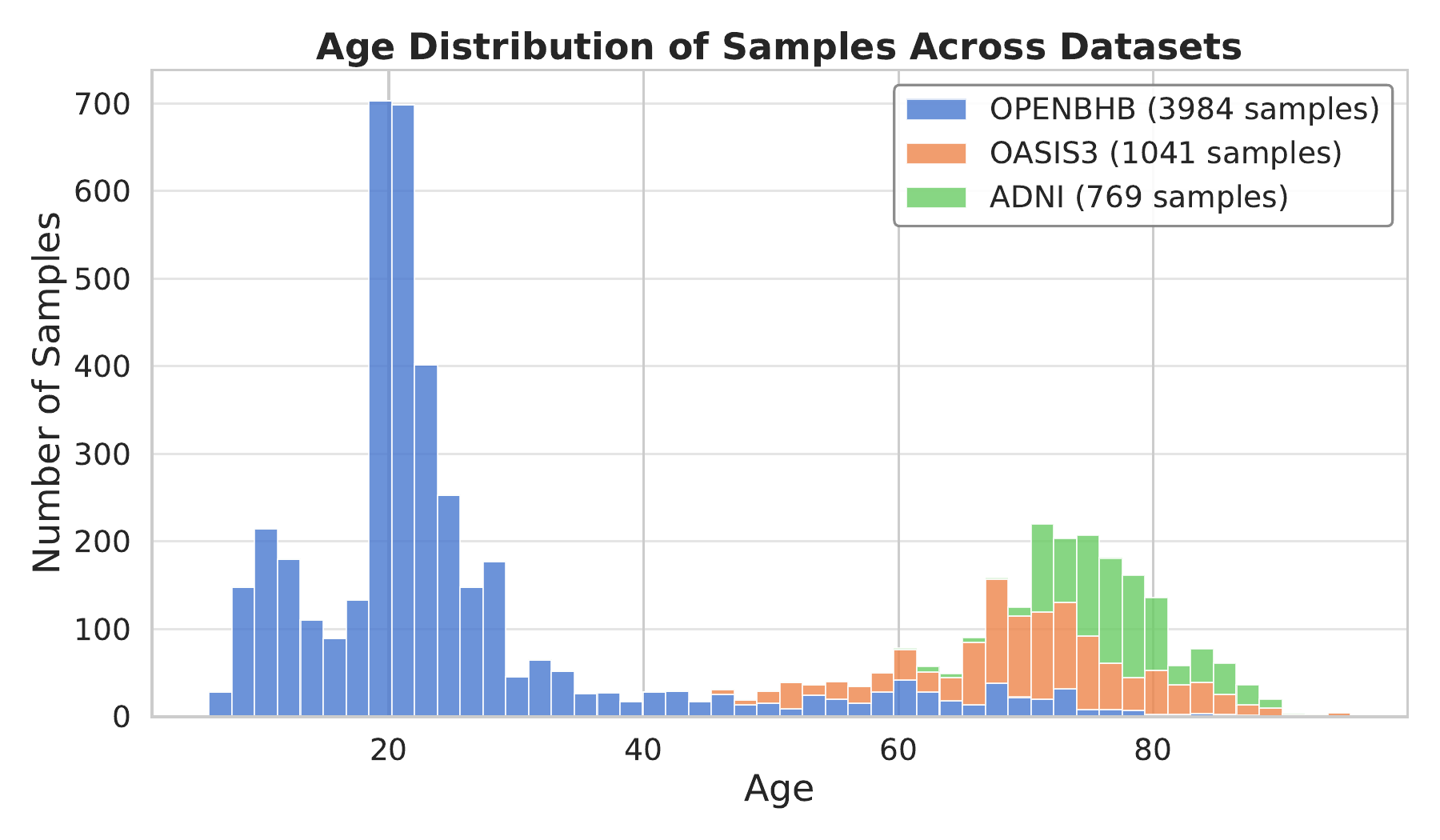}
\captionof{figure}{Age distribution across datasets. Overlaid histograms show the number of samples per age for OpenBHB, OASIS-3, and ADNI. OpenBHB is concentrated in younger adults (strong peak around the early 20s), whereas OASIS-3 and ADNI primarily cover older adults (50–90 years), with highest density around the 70–80 age range.}
\label{fig:Age_Distribution}
\end{center}

\subsection{Age distribution and data splits}
\label{sec:age_distribution}

Figure~\ref{fig:Age_Distribution} illustrates the combined age distribution of the integrated training cohort. OpenBHB contributes the majority of samples, producing a pronounced peak in younger age groups (approximately 10–30 years). The inclusion of ADNI and OASIS-3 substantially enriches representation in older age ranges (60–95 years), resulting in a more balanced coverage across adulthood. This broader distribution is essential for learning age-dependent anatomical variability and for evaluating age-conditioned synthesis.

For generative evaluation, test samples were drawn from a 20\% split of the integrated training distribution, preserving the same age profile. In contrast, BAP evaluation was performed on the independent OpenBHB validation set, which follows the original OpenBHB age distribution and provides an unbiased benchmark for downstream prediction performance.

\subsection{Preprocessing pipeline}
\label{sec:preprocessing}

All MRI volumes were processed using a standardized preprocessing pipeline applied uniformly across datasets.
In particular, skull stripping plays a critical role in preventing the model from exploiting non-brain structures (e.g., skull, scalp, or neck regions) that may correlate with age but are unrelated to brain morphology. Such spurious correlations can induce a \emph{Clever Hans} effect~\cite{cleverhans}, where predictions rely on superficial cues rather than meaningful neuroanatomical patterns. By isolating brain tissue, the model is forced to focus on age-relevant neuroanatomical patterns only.

The following preprocessing steps were applied uniformly using tools from ANTs~\cite{Tustison2021-sc} and FSL:

\begin{itemize}
    \item \textbf{Bias field correction:} N4ITK correction~\cite{Tustison2010-sx} was applied using ANTs to remove smooth, low-frequency intensity variations that commonly affect MRI scans;
    \item \textbf{Spatial normalization:} Affine registration to the MNI152 template~\cite{FONOV2009S102} using FSL FLIRT~\cite{JENKINSON2002825};
    \item \textbf{Skull stripping:} Brain extraction was conducted on registered images using FSL BET~\cite{Smith2002-lm};
    \item \textbf{Resampling:} Volumes were resampled to a common isotropic resolution of $91 \times 109 \times 91$, selected for compatibility with the downstream BAP model;
    \item \textbf{Intensity normalization:} Final volumes were normalized using z-score normalization (zero mean, unit variance).
\end{itemize}

Scans that failed automated preprocessing were excluded from the final dataset. No additional manual curation was performed. A complete list of retained samples and associated metadata is provided in the codebase, along with the full preprocessing implementation.

\section{Experimental setup}
This section describes the experimental protocol, baseline models, and training configuration. Evaluation metrics and downstream validation tasks are described in the following sections.

\subsection{Baselines} 
We compare FlowLet against the following seven state-of-the-art generative models for 3D brain MRI synthesis:
\begin{itemize}
    \item \textbf{WDM}: unconditional Wavelet Diffusion Model operating in the 3D Haar wavelet domain~\cite{DBLP:conf/miccai/FriedrichWBDC24};
    \item \textbf{MD}: Medical Diffusion, a DDPM trained on a VQ-GAN backbone~\cite{DBLP:journals/corr/abs-2211-03364};
    \item \textbf{MLDM}: MONAI Latent Diffusion Model~\cite{pinaya2022brain};
    \item \textbf{BS}: BrainSynth, a VQ-VAE and Transformer-based model for brain MRI synthesis~\cite{DBLP:journals/natmi/TudosiuPCDPBFGGNOC24};
    \item \textbf{MOTFM}: a recent Flow Matching–based framework~\cite{yazdani2025flow}.
\end{itemize}

Several of these baselines are unconditional in their original public implementations. To enable a fair comparison on the age-conditioned synthesis and downstream BAP task, we introduce age-conditioned variants of WDM and MOTFM, denoted as \textbf{WDMa} and \textbf{MOTFMa}, respectively. Conditioning is implemented using the same strategy adopted in FlowLet: the scalar age value is normalized, projected into a 512-dimensional embedding, and injected into the U-Net backbone via cross-attention. This ensures that all conditional models receive age information in a consistent manner, without introducing architectural advantages specific to FlowLet.

We exclude subject-specific brain aging models~\cite{DBLP:conf/miccai/LitricoGGRB24,DBLP:journals/mia/PomboGCORAN23,yeganeh2025latent}, which aim to transform an existing MRI of a given subject to a different age. These approaches address a fundamentally different task, as they rely on identity preservation and do not perform synthesis from noise, making them not directly comparable to our setting.

\subsection{Training protocol and implementation details}
\label{sec:implementation}
FlowLet is implemented in PyTorch, using the AdamW optimizer with cosine annealing and mixed precision. For more memory-efficient attention,  \texttt{xformers}\footnote{\url{https://github.com/facebookresearch/xformers}} is optionally available. All experiments were run on an NVIDIA A6000 GPU (48 GB VRAM); notably, while baseline models required this large memory capacity, FlowLet can be trained in 24 GB, enhancing accessibility. To compare different flow formulations (RFM, CFM, VP, and Trigonometric), all variants were implemented and trained under identical architecture, hyperparameters, and optimization settings. All external baselines were likewise trained from scratch on the same dataset, using the provided default hyperparameters except for input channels and padding to match our volumes. Although faster diffusion samplers such as DDIM exist \cite{DBLP:conf/iclr/SongME21}, they are not part of the original released implementations of the considered baselines. In this work, we retain the standard sampling procedures provided by each method to ensure a consistent and reproducible comparison across models. We note that FlowLet does not rely on accelerated sampling heuristics but directly benefits from the deterministic ODE formulation, which enables efficient generation without requiring additional sampling approximations. Complete training details and hyperparameters are provided in Appendix \ref{sec:hyperparams}. 
Detailed efficiency benchmarks (VRAM, training time, inference throughput, and scaling with resolution) are reported in Appendix \ref{Efficiency_benchmarks}.

\subsection{Ablation study design}
To assess the contribution of individual design choices within FlowLet, we conduct a series of controlled ablation experiments. Specifically, we analyze the impact of: (i) the wavelet basis used for the invertible representation, (ii) the number of inference steps and the associated sampling efficiency, (iii) the conditioning mechanisms, including FiLM and spatial conditioning via cross-attention, and (iv) the numerical solver employed for ODE integration.

In each ablation, a single component is modified or removed while keeping the remaining architecture, training procedure, and evaluation protocol unchanged. All ablation variants are trained and evaluated under the same experimental setup as the full model to ensure a fair and isolated assessment of each design choice. Quantitative and qualitative results of these ablations are reported in Section \ref{sec:ablation_studies}.

\section{Evaluation metrics}
\label{sec:evaluation_metrics}
This section describes the quantitative metrics used to evaluate generative performance in terms of image fidelity, sample diversity, and anatomical plausibility, as well as their relevance to downstream clinical tasks.

\subsection{Image fidelity and diversity}
Image fidelity and distributional alignment between synthetic and real data were evaluated using the Fréchet Inception Distance (FID)~\cite{DBLP:conf/nips/HeuselRUNH17} and the Gaussian kernel-based Maximum Mean Discrepancy (MMD)~\cite{DBLP:journals/jmlr/GrettonBRSS12}. Both metrics were computed on feature representations extracted from a ResNet-50 pretrained on medical images~\cite{DBLP:journals/corr/abs-1904-00625}, following established evaluation practices for 3D medical image synthesis~\cite{DBLP:conf/miccai/FriedrichWBDC24}. Lower values of FID and MMD indicate closer alignment between the distributions of synthetic and real samples.

Sample diversity was assessed using an \emph{intra-set} Multi-Scale Structural Similarity Index (MS-SSIM)~\cite{1292216}, computed as the average pairwise similarity among generated samples. In this setting, higher intra-set MS-SSIM values indicate reduced inter-sample variability and may suggest mode collapse, whereas lower values correspond to increased diversity. Statistical significance of pairwise comparisons was assessed using two-sided Wilcoxon rank-sum tests with Bonferroni correction (\(\alpha = 0.05\)).

Although MS-SSIM is traditionally used to compare a generated image with a reference target, it has been widely adopted in 3D brain MRI synthesis as an intra-set metric to quantify structural similarity among generated samples~\cite{pinaya2022brain,DBLP:conf/miccai/FriedrichWBDC24}. When used in this manner, MS-SSIM should be interpreted as a relative measure under consistent evaluation conditions rather than against an absolute threshold. A well-performing generative model is therefore expected to preserve anatomically coherent structures while maintaining realistic inter-subject variability.

Because intra-set MS-SSIM values are influenced by the conditioning range used during generation, diversity metrics depend on the age interval over which samples are synthesized. To account for this effect, we perform an additional age-stratified evaluation in which fidelity and diversity metrics are computed independently within restricted age ranges (non-overlapping age bins 15–30, 40–55, and 65–80 years). The detailed age-stratified protocol and corresponding quantitative results are reported in Appendix~\ref{age_stratified_quantitative_evaluation}.

\subsection{Region-based anatomical plausibility}
\label{sec:regional_metrics}

Global similarity metrics are effective at assessing overall image quality and distributional alignment, but they may overlook fine-grained anatomical inconsistencies that are critical in clinical neuroimaging \cite{wu2025medical}. In volumetric MRI in particular, metrics such as FID and MS-SSIM can be dominated by non-informative voxels, potentially producing favorable scores even when clinically relevant brain structures differ \cite{jafrasteh2025wasabi}. To complement the global evaluation, we therefore perform a region-based anatomical analysis that explicitly focuses on local realism and morphological consistency across anatomically defined brain regions.

This evaluation leverages \texttt{FastSurfer}\footnote{\url{https://deep-mi.org/research/fastsurfer/} (v2.4.2)}, a deep-learning pipeline for automated segmentation and parcellation of structural MRIs. Each volume is segmented into 95 cortical and subcortical regions of interest (ROIs). Let $\mathcal{R}$ denote the set of ROIs. For each region $r \in \mathcal{R}$, let $V_r^{(R)}$ and $V_r^{(S)}$ denote the sets of voxels belonging to region $r$ in the segmentations of the real reference volume $R$ and the synthetic volume $S$, respectively. Region-based metrics are computed independently for each ROI and subsequently averaged across $\mathcal{R}$, yielding summary scores that reflect anatomical plausibility across the full brain rather than a limited subset of structures.

A key aspect of this evaluation is the pairing strategy between synthetic and real volumes. Each synthetic brain volume is compared against a \emph{different age-matched} real subject, and metrics are computed independently for each ROI prior to averaging across $\mathcal{R}$. This one-to-one, age-matched evaluation prevents a mode-collapsed model from appearing artificially strong: a single repeated anatomy cannot simultaneously match the diverse set of real age-matched references. Consequently, favorable scores reflect genuine anatomical plausibility and alignment with natural inter-subject variability rather than artifacts stemming from reduced diversity.

\paragraph{Models and sample generation.}
The region-based evaluation includes all baseline models (WDM, WDMa, MD, MLDM, MOTFM, MOTFMa, BS) as well as all FlowLet variants (RFM, CFM, VP, and Trigonometric). Unless otherwise specified, FlowLet variants are configured for 10-step generation. For each model, 500 synthetic brain volumes are generated with age conditions linearly spanning the full age range of the training set. To ensure comparability across FlowLet variants despite differences in the underlying flow formulation, the same random seed is used for all generations. All synthetic volumes are segmented into 95 anatomical classes using \texttt{FastSurfer}. A reference set of 500 real brain MRIs, spanning the same age range, is processed through the identical segmentation pipeline and used as anatomical ground truth.

\paragraph{Region-wise metric definitions.}
Due to natural anatomical variability and potential segmentation mismatches, ROI segmentations extracted from synthetic and real volumes may not perfectly coincide, resulting in different voxel supports for a given region. Using only one of the two supports could therefore bias region-wise comparisons. To ensure a robust and unbiased evaluation, we define the comparison domain for each region as the union of voxel sets derived from both the real and synthetic segmentations,
\[
U_r = V_r^{(R)} \cup V_r^{(S)} .
\]
All region-wise metrics are computed over this unified voxel set and subsequently averaged across all regions in $\mathcal{R}$, providing a stable summary of anatomical plausibility relative to the real reference.

We define three complementary summary metrics:
{\small
\begin{equation}
\mathrm{iMAE} =
\frac{1}{|\mathcal{R}|}
\sum_{r \in \mathcal{R}}
\left(
\frac{1}{|U_r|}
\sum_{v \in U_r}
\lvert R_v - S_v \rvert
\right),
\end{equation}
\begin{equation}
\mathrm{KLD} =
\frac{1}{|\mathcal{R}|}
\sum_{r \in \mathcal{R}}
D_{\mathrm{KL}}\!\left(
P_r^{(R)} \,\Vert\, P_r^{(S)}
\right),
\end{equation}
\begin{equation}
\mathrm{DICE} =
\frac{1}{|\mathcal{R}|}
\sum_{r \in \mathcal{R}}
\frac{2\,\lvert V_r^{(R)} \cap V_r^{(S)} \rvert}
{\lvert V_r^{(R)} \rvert + \lvert V_r^{(S)} \rvert}.
\end{equation}
}

The \emph{overall intensity Mean Absolute Error (iMAE)} quantifies local intensity realism by measuring the average absolute voxel-wise difference between $R$ and $S$ samples within each ROI. The \emph{overall Kullback--Leibler divergence (KLD)} evaluates the similarity of regional intensity distributions by comparing normalized histograms of real ($P_r^{(R)}$) and synthetic ($P_r^{(S)}$) intensities computed over $U_r$. Finally, the \emph{overall Dice Similarity Coefficient (DICE)} assesses morphological consistency through the spatial overlap between real and synthetic ROI segmentations. Dice values range from 0 (no overlap) to 1 (perfect overlap), while lower values of iMAE and KLD indicate improved intensity and distributional alignment.

\paragraph{Interpretation alongside global metrics.}
Region-based anatomical fidelity metrics are intended to be interpreted in conjunction with global distributional and diversity measures, including FID, MMD, and intra-set MS-SSIM. A reliable generative model should achieve high anatomical accuracy (e.g., low iMAE and high DICE) while simultaneously preserving sufficient inter-sample variability (low MS-SSIM) and global distributional alignment (low FID and MMD). Importantly, a mode-collapsed model may result in a deceptively strong region-wise score for a single anatomy, but will be revealed by elevated MS-SSIM and degraded global metrics. Considering these measures jointly therefore provides a more comprehensive and reliable assessment of generative quality.

\label{sec:roi}

\subsection{Downstream Brain Age Prediction}
\label{sec:bap}

The clinical usefulness and functional fidelity of the synthetic data were evaluated through their impact on BAP. This task is particularly relevant in our setting because older adults are underrepresented in the training distribution, yet are the population most susceptible to cognitive decline. Establishing a normative trajectory of healthy aging therefore requires that BAP be defined and evaluated exclusively on cognitively normal subjects. Following the protocol of~\cite{de2024explainable}, we assess whether synthetic data generated by each method improves age regression performance when training a 3D BAP predictor and evaluating it on a real, held-out test set.

\paragraph{BAP model and training protocol.}
We employ a 3D DenseNet-121 architecture configured for regression, with a linear output layer predicting chronological age from structural T1-weighted MRI volumes, following~\cite{de2024explainable}. Input intensities are normalized to the $[0,1]$ range using the 5th and 95th percentile values computed from the \emph{training set} to avoid test-set leakage. Models are trained using stochastic gradient descent with a cosine annealing warm restarts scheduler.

For each generative method, the downstream BAP evaluation is performed according to the following protocol:
\begin{enumerate}
    \item \textbf{Synthetic data generation:} 3,000 synthetic brain MRI volumes are generated for each method;
    
    \item \textbf{Age labeling:} since BAP is a supervised task, each synthetic volume is associated with an age label. For conditional generative models, age is used directly as the conditioning variable during generation, with samples spanning the full training age range (5.9--95.5 years). For unconditional generators, age labels are assigned by sampling from the empirical age distribution of the training set to ensure a fair comparison;
    
    \item \textbf{BAP model training:} a separate instance of the BAP network is trained for each generative method using the same architecture and optimization protocol;
    
    \item \textbf{Evaluation dataset:} all trained BAP predictors are evaluated on the independent OpenBHB validation set, restricted to cognitively normal subjects aged 44 years and older;
    
    \item \textbf{Performance metric:} prediction performance is quantified using Mean Absolute Error (MAE) in years, where lower values indicate more accurate estimation of chronological age.
\end{enumerate}

\begin{figure*}[t]
% \centering
\includegraphics[width=\linewidth]{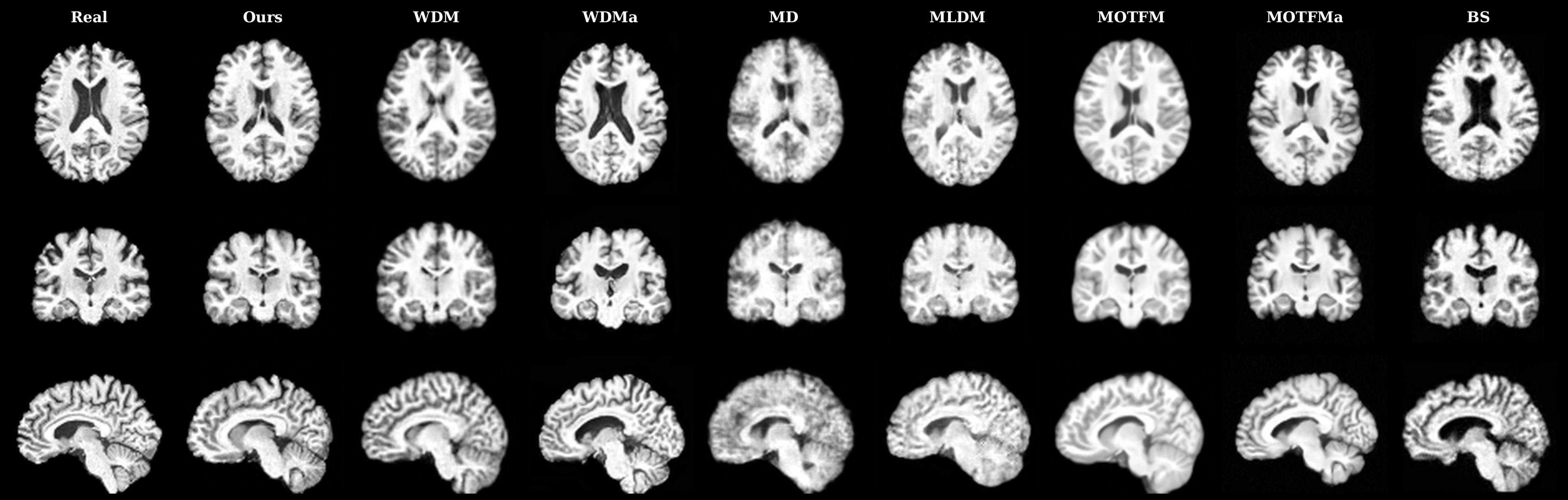}
\caption{{Visual assessment of image fidelity and realism for different 3D brain MRI synthesis models. Each column displays standard Axial, Coronal, and Sagittal views for the real reference scan (Subject of 72 years old) and specified generative method (Ours is RFM 10 steps).}}\label{fig:Samples_comparison}
\end{figure*}

\section{Results}
We evaluate FlowLet against baselines, combining qualitative inspection with quantitative evidence: global metrics, BAP performance, and ROI-based anatomical plausibility.
\subsection{Qualitative evaluation}
\label{sec:qualitative_results}

Figure~\ref{fig:Samples_comparison} presents representative samples generated by each model, providing a first qualitative assessment of image-level fidelity and anatomical realism. 
The FlowLet-generated volumes exhibit anatomically coherent structures and consistent global organization across views. However, visual differences in fine-scale anatomical details (e.g., cortical folding and cerebellar structures) remain subtle when compared to strong baselines such as WDM and MOTFMa. In particular, some degree of blurring is still observable in high-frequency regions across all methods, including FlowLet.

In addition, to isolate the effect of age conditioning, we generate a fixed-seed trajectory in which only the conditioning variable is varied while keeping the initialization constant (Appendix \ref{sec:fixed_seed_age_trajectory_appendix}, Figure \ref{fig:fixed_seed_age_trajectory_appendix}). This controlled setting highlights coherent age-dependent morphological changes.

A more detailed analysis of the wavelet-domain behavior is provided in Appendix \ref{sec:high_frequency_wavelet_analysis_appendix} (Figures \ref{fig:wavelet_tail_metrics_appendix} and \ref{fig:wavelet_abs_montage_appendix}).

\begin{figure*}[t!]
\centering
\includegraphics[width=1\linewidth]{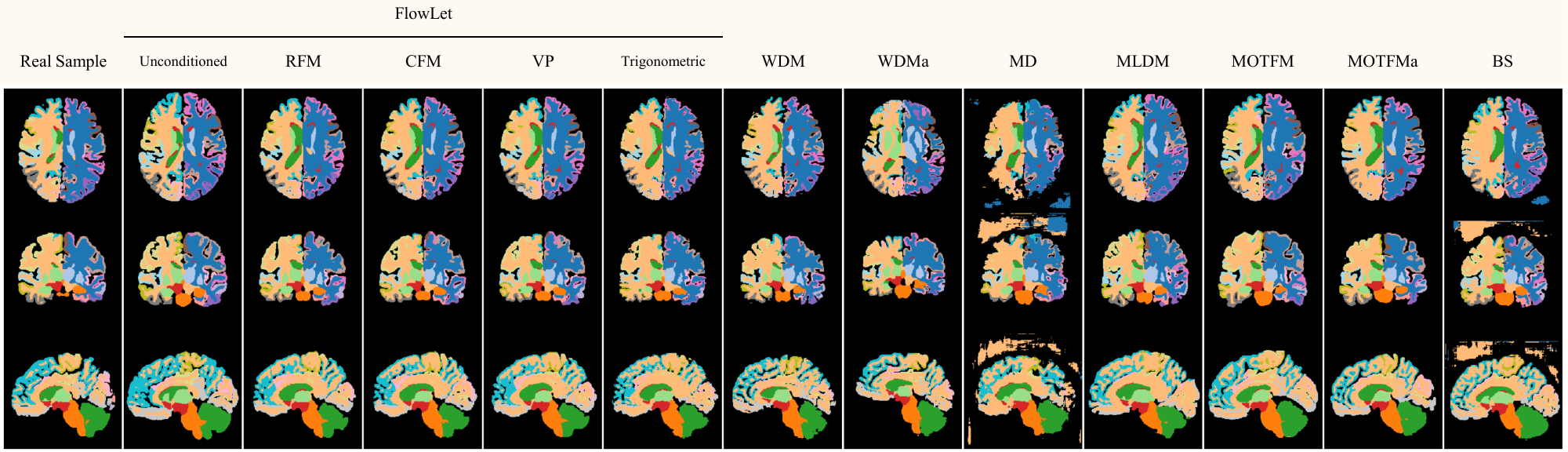}
\caption{Qualitative comparison of FastSurfer segmentations of synthetic data from different models. The leftmost column displays the segmentation from a real 72-year-old subject, while each subsequent column shows axial, coronal, and sagittal views relative to a synthetic sample of the same age generated by a specific model. FlowLet samples are generated at 10 steps.}
\label{fig:SEG_comparison}
\end{figure*}

The segmentation outputs provide an additional qualitative perspective on anatomical consistency across models. Figure~\ref{fig:SEG_comparison} reports representative segmentation outputs for synthetic samples generated at the same target age (72 years), compared against a real reference subject. For unconditional baselines (MOTFM, MD, and WDM), the synthetic sample is selected randomly, whereas FlowLet samples are generated with explicit age conditioning. All FlowLet variants produce broadly consistent and anatomically plausible segmentations, with the Trigonometric formulation exhibiting slightly less regular boundaries.

Visual differences across methods remain subtle in this setting, particularly for models that do not exhibit clear failure modes. As a result, qualitative inspection alone is not sufficient to establish meaningful differences in anatomical fidelity. For this reason, we complement this analysis with region-based quantitative metrics (Section \ref{sec:region_based_res}), which provide a more sensitive and reliable evaluation across anatomical regions.

A step-wise qualitative comparison across flow variants for 1–200 ODE steps is provided in~\ref{Additional_qualitative} for completeness.

\input{table_metrics_complete_overall}

\subsection{Quantitative evaluation}

\subsubsection{Image fidelity and diversity}
\label{sec:fid_div}
We first assess the generative performance of FlowLet in terms of image fidelity and sample diversity, with particular attention to sampling efficiency. Table~\ref{tab:ours_vs_baselines} reports global similarity metrics for FlowLet variants and competing generative models. All FlowLet variants achieve highly competitive Fréchet Inception Distance (FID) and Maximum Mean Discrepancy (MMD) scores using only 10 ODE sampling steps, outperforming diffusion-based baselines such as WDM, MD, and MLDM, which require substantially more sampling iterations. Among the evaluated variants, the Trigonometric formulation attains the lowest FID and MMD values, indicating strong alignment with the real data distribution in feature space.
To contextualize these values, we also report real-to-real baselines computed between disjoint subsets of the held-out data, as well as on the independent test set (Appendix \ref{sec:real_to_real_references_appendix}), providing an empirical reference for the lower bound of each metric.

Notably, improvements in global fidelity are not obtained at the expense of sample diversity. All FlowLet variants maintain intra-set MS-SSIM values that are comparable to, or lower than, those of baseline models, suggesting that enhanced similarity does not arise from mode collapse or reduced inter-sample variability. This indicates that FlowLet preserves a realistic level of diversity while improving global distributional alignment.

Figure \ref{fig:steps_ablation} further illustrates the trade-off between sampling efficiency and image fidelity for FlowLet variants. FID consistently improves as the number of sampling steps increases, with performance saturating at approximately 10 steps for RFM, CFM, and VP. Beyond this point, additional steps do not materially improve these formulations, while the Trigonometric path becomes unstable at 200 steps and degrades in both quantitative and downstream behaviour.

\subsubsection{Brain Age Prediction}
\label{sec:bap_res}
We next evaluate whether synthetic brain MRIs generated by FlowLet improve performance on the downstream BAP task, a clinically relevant benchmark that is particularly sensitive to age-related anatomical variation. Table~\ref{tab2a:brain_age_results_age_44} reports BAP results for cognitively normal subjects aged 44 years and older, comparing models trained on real data alone with models augmented using synthetic samples from different generative methods. To further assess generalization beyond the training distribution, we additionally evaluate the augmentation pipeline on a fully independent external dataset (DLBS), with results reported in Appendix \ref{sec:dlbs_external_validation_appendix}.

BAP models trained with FlowLet-generated data consistently achieve lower test MAE than those trained on real data alone and those augmented with synthetic samples from baseline generators. Among FlowLet variants, the RFM and CFM formulations yield the best test performance, outperforming both unconditional baselines (e.g., WDM and MD) and the strong conditional MLDM baseline. This indicates that explicit age-conditioned synthesis from noise is beneficial for improving downstream age regression in the present setting.

The comparison between MOTFM and its age-conditioned counterpart MOTFMa further highlights the importance of conditioning. While the original MOTFM framework performs poorly on BAP, applying the same conditioning strategy used in FlowLet leads to a substantial reduction in test error. This confirms that the observed performance gains are driven by effective age conditioning rather than by architectural differences alone.

Notably, BAP models trained with FlowLet-RFM and FlowLet-CFM synthetic data achieve lower test error than the model trained exclusively on real data, suggesting that FlowLet-generated samples provide complementary age-relevant information that mitigates dataset imbalance and improves generalization.

\input{TWOSIDED_TABLES_ROI_BAP}

\subsubsection{Region-based anatomical plausibility}
\label{sec:region_based_res}
We further assess whether improvements in global image fidelity correspond to anatomically meaningful realism by evaluating local structural consistency across anatomically defined brain regions. 
Table~\ref{tab:ROI_table} reports region-based metrics computed on FastSurfer segmentations, including intensity-based errors (iMAE, KLD) and morphological overlap (DICE), averaged over 95 cortical and subcortical regions.

FlowLet variants, particularly RFM and VP, rank among the strongest methods on the region-based metrics, with lower iMAE and KLD values and competitive DICE scores relative to most evaluated baselines. In contrast, several baseline models that exhibit competitive global fidelity metrics show degraded regional performance. For example, while the MD baseline achieves relatively low global FID values, its substantially lower DICE score ($0.294$) reveals deficiencies in capturing accurate anatomical shapes at the regional level.

These region-level results also help contextualize the downstream BAP findings. Although the Trigonometric formulation achieves strong global similarity metrics, it underperforms relative to RFM and CFM on both region-based anatomical measures and brain age prediction. This discrepancy indicates that alignment in global feature space does not necessarily guarantee preservation of fine-grained anatomical structure that is critical for age-related tasks.

To complement the overall metrics, we report additional age-stratified results across three non-overlapping age bins in~\ref{age_stratified_quantitative_evaluation}, coupled with the corresponding p-value results and the complete significance-testing protocol for pairwise comparisons in~\ref{sec:pvalues}.

\input{ablation_section}

\paragraph{Flow Matching variants.}
We compare different flow matching formulations within the same architectural and training setup to assess the effect of interpolation geometry on generative performance. Specifically, we evaluate RFM, CFM, VP diffusion matching, and the Trigonometric formulation.

Across evaluations, RFM and CFM consistently provide the most stable and reliable performance, achieving competitive global fidelity while maintaining strong downstream BAP accuracy and regional anatomical plausibility. In contrast, the Trigonometric formulation attains favorable global metrics at low step counts but exhibits degraded regional consistency and downstream performance as the number of steps increases. The VP formulation does not yield a clear advantage over RFM or CFM to offset its additional complexity (Tables\ref{tab:ours_sidebyside}, \ref{tab2a:brain_age_results_age_44}, and \ref{tab:ROI_table}).

These results support RFM as the preferred formulation for FlowLet, offering a robust balance between generative quality, stability, and computational efficiency.

\section{Discussion}

\subsection{Efficient 3D MRI synthesis without latent compression}

FlowLet was designed to address a practical bottleneck in 3D neuroimaging generation: achieving high anatomical fidelity and controllable synthesis while keeping sampling costs tractable. Most state-of-the-art generative models for volumetric MRI rely on diffusion-based sampling~\cite{DBLP:conf/nips/HoJA20, DBLP:conf/iclr/SongME21}, which typically requires hundreds to thousands of iterations to reach high-quality outputs. To mitigate the associated computational burden, many approaches operate in learned latent spaces~\cite{DBLP:conf/cvpr/RombachBLEO22,pinaya2022brain}, trading spatial resolution and anatomical detail for improved efficiency.

Our results show that this trade-off is not unavoidable. By performing Flow Matching directly in an invertible wavelet domain~\cite{DBLP:conf/iclr/LipmanCBNL23}, FlowLet achieves competitive or superior global fidelity compared to latent diffusion and wavelet diffusion baselines~\cite{DBLP:conf/miccai/FriedrichWBDC24}, while requiring an order-of-magnitude fewer sampling steps (see Section \ref{sec:fid_div}). Importantly, this efficiency gain does not rely on learned compression: the wavelet representation preserves exact invertibility and fine-grained spatial detail, avoiding the reconstruction artifacts commonly observed in latent autoencoder-based pipelines~\cite{Muller-Franzes2023-ju}.

The ablation results further clarify the factors enabling this efficiency. Performance saturates at approximately ten ODE steps across all FlowLet variants, indicating that long stochastic trajectories are unnecessary when transport is learned in a multi-scale, invertible representation. In contrast to diffusion-based models, where increasing the number of steps is often essential to reduce noise-induced artifacts~\cite{DBLP:conf/iclr/SongME21}, FlowLet benefits from smooth and stable trajectories already at low step counts.

Compared to wavelet diffusion models, which reduce memory usage but remain constrained by slow iterative sampling~\cite{DBLP:conf/miccai/FriedrichWBDC24}, FlowLet demonstrates that wavelet representations can be effectively combined with deterministic flow-based generation. 

\subsection{Computational efficiency and accessibility}

Beyond sampling efficiency, FlowLet substantially reduces the computational resources required for training and inference. 
As reported in Section \ref{sec:implementation} and Appendix~\ref{Efficiency_benchmarks}, operating in the wavelet domain results in an approximately $8\times$ reduction in memory consumption compared to voxel-space diffusion models, enabling training with batch size 4 using approximately 22 GB of VRAM. In contrast, diffusion-based baselines such as WDM and MLDM require over 40 GB of VRAM under comparable settings. This reduced memory footprint allows FlowLet to be trained and deployed on widely available consumer-grade GPUs (e.g., RTX 3090/4090), lowering the barrier to entry for large-scale 3D MRI synthesis. Inference efficiency further reinforces this advantage: FlowLet generates a full-resolution 3D brain MRI in approximately 1.6 seconds using 10 ODE steps, representing a speedup of over one order of magnitude relative to the original released conditional diffusion baseline implementations used in this study.

These practical considerations are particularly relevant for research environments with limited computational resources and support the scalability of FlowLet to large neuroimaging cohorts.

\subsection{Global metrics are not enough in volumetric neuroimaging}

Global distributional metrics such as FID, MMD, and MS-SSIM are widely adopted to evaluate generative models, including in medical imaging applications~\cite{pinaya2022brain, DBLP:conf/miccai/FriedrichWBDC24}. However, our results highlight that strong performance on these metrics does not necessarily translate into anatomically meaningful realism when synthesizing high-dimensional volumetric data such as 3D brain MRI (see Section \ref{sec:region_based_res}).

Several baseline models achieve competitive global fidelity scores, yet exhibit degraded regional anatomical consistency and weaker downstream performance. This discrepancy is particularly evident in diffusion- and latent-based approaches, which may generate globally plausible intensity distributions while failing to preserve fine-grained structural details in anatomically relevant regions~\cite{DBLP:conf/nips/HoJA20,DBLP:conf/cvpr/RombachBLEO22}.

In volumetric MRI, the predominance of background voxels further amplifies this effect, allowing global metrics to be dominated by non-informative regions and potentially masking localized anatomical errors~\cite{DBLP:conf/miccai/FriedrichWBDC24}. Our region-based evaluation explicitly addresses this limitation by quantifying anatomical fidelity across 95 cortical and subcortical regions.

This analysis reported in Section \ref{sec:region_based_res} reveals cases in which models with favorable global similarity scores nonetheless show reduced Dice overlap and increased regional intensity discrepancies. Importantly, these anatomical deficiencies are reflected in downstream brain age prediction performance, indicating that global similarity alone is insufficient to assess functional utility in clinically relevant tasks.

Overall, no single metric fully captures the behavior of the proposed model. While improvements are not uniformly large across all evaluation criteria, FlowLet consistently achieves competitive or improved performance across global, regional, and downstream evaluations. This pattern suggests that the contribution of the model lies in the combination of efficiency, controllability, and anatomically meaningful generation, rather than in a single dominant metric.
Region-based and downstream evaluations provide essential insight into model behavior that would otherwise remain hidden when relying exclusively on global similarity measures, and should be considered standard components of future assessments in 3D medical image synthesis.

\subsection{The role of explicit age conditioning}

A central finding of this work is that explicit age conditioning  is important for generating synthetic brain MRIs that remain coherent across age trajectories and useful for downstream age-related tasks. While several generative models can approximate the overall distribution of brain MRI intensities, our results show that unconditional synthesis is insufficient when the target application requires control over biologically meaningful attributes such as age.

In the absence of explicit conditioning, generative models tend to learn an average representation of the population, producing samples that appear globally realistic but lack consistent age-specific morphological patterns. This behavior is reflected in our experiments by the poor brain age prediction performance of unconditional models, including diffusion-based and flow-based baselines, despite their competitive global fidelity scores (as shown in Section \ref{sec:bap_res}).

This pattern is also reflected in the region-based evaluation, where the unconditional FlowLet variant remains broadly plausible overall but yields weaker anatomical scores than the age-conditioned variants.

The ablation analysis reported in Section \ref{sec:ablation_studies}, further highlights the importance of how conditioning information is injected into the model. Using either feature-wise modulation or spatial conditioning alone leads to degraded downstream and region-based performance, even when global metrics remain largely unaffected. Only the combination of feature-wise conditioning via FiLM and spatially adaptive conditioning via cross-attention consistently preserves age-relevant anatomical variations across the brain. This suggests that effective conditioning in volumetric neuroimaging requires both global control of feature statistics and localized modulation of spatial structures.

These observations are reinforced by the comparison between MOTFM and its age-conditioned variant MOTFMa. While the original MOTFM framework performs poorly on brain age prediction, applying the same age-conditioning strategy used in FlowLet yields a substantial improvement in downstream performance. This indicates that the observed gains are driven primarily by effective conditioning rather than by architectural differences alone.

Overall, our findings suggest that age-conditioned synthesis from noise remains underexplored in 3D neuroimaging and is particularly valuable for applications that require controlled generation along biologically interpretable dimensions. Explicit conditioning mechanisms therefore represent an important design direction for clinically meaningful MRI synthesis.

\subsection{Flow geometry matters more than solver accuracy}

Our ablation results indicate that the geometry of the learned transport path plays a more critical role in generative performance than the numerical accuracy of the ODE solver. While higher-order solvers are often assumed to improve sample quality in continuous-time generative models, our experiments show that increasing solver accuracy alone does not compensate for unfavorable flow geometry in high-dimensional anatomical settings.

In particular, replacing Euler integration with a fourth-order Runge--Kutta (RK4) solver for the Trigonometric formulation does not lead to improvements in either global fidelity or downstream performance. Instead, models based on curved interpolation paths exhibit increased instability and degraded anatomical consistency as the number of steps increases. This behavior suggests that the limitations observed for these formulations are primarily attributable to the structure of the velocity field rather than to discretization error.

In contrast, Rectified Flow Matching and Conditional Flow Matching, which rely on straight or near-straight interpolation paths, consistently provide stable and reliable performance across global, regional, and downstream evaluations (see Section \ref{sec:ablation_studies}). These findings are consistent with recent theoretical and empirical observations in the flow matching literature, which highlight the benefits of low-curvature transport paths in terms of training stability and expressiveness~\cite{DBLP:conf/iclr/LipmanCBNL23,DBLP:conf/icml/LeeKY23}.

Taken together, these results suggest that, in the context of volumetric medical image synthesis, the choice of flow formulation and interpolation geometry has a greater impact on anatomical plausibility and downstream utility than the choice of numerical solver. Prioritizing simple and stable transport paths therefore appears more effective than increasing solver order when designing efficient generative models for high-dimensional anatomical data.

\section{Limitations and future directions}

Despite the encouraging results, several limitations should be acknowledged. First, while region-based anatomical metrics and downstream brain age prediction provide stronger proxies than global similarity scores, they do not replace expert-driven clinical assessment. Future work should therefore incorporate structured evaluations by neuroradiologists to further validate anatomical realism.

Second, although FlowLet supports explicit conditioning, this work focuses exclusively on age. Extending the framework to multi-attribute conditioning (e.g., sex, pathology, cognitive scores) introduces additional challenges related to disentanglement and robustness, which require systematic investigation. In addition, the current evaluation matches synthetic and real subjects based solely on age. Other covariates, such as sex, are not consistently available across all datasets and are therefore not included in the matching procedure. While the preprocessing pipeline mitigates part of the inter-subject variability, incorporating multi-attribute matching represents an important direction for future work. 

Third, while the proposed preprocessing pipeline mitigates spurious correlations, comprehensive audits of fairness, robustness across demographic subgroups, and potential privacy leakage remain open challenges. Addressing these aspects will be essential for the deployment of generative models in sensitive clinical settings.

Finally, this study is limited to T1-weighted structural MRI. Future work will explore the application of FlowLet to other 3D imaging modalities and the integration of uncertainty estimation to better characterize the reliability of synthetic cohorts.

\section{Conclusion}

This work introduced FlowLet, a conditional generative framework for 3D brain MRI synthesis that combines flow matching with an invertible wavelet representation. By avoiding learned latent compression, FlowLet achieves efficient generation with few sampling steps while preserving anatomically meaningful structure. Extensive evaluation demonstrates that improvements in global fidelity translate into stronger regional anatomical plausibility and improved performance on a clinically relevant downstream task, brain age prediction. 

Our results highlight the importance of combining efficient generative modeling with explicit conditioning and anatomy-aware evaluation when synthesizing volumetric medical images. While limitations remain, FlowLet provides a scalable, open-source, and controllable approach to 3D MRI synthesis and represents a step toward more practical and clinically meaningful generative models in neuroimaging.

%% file: table_metrics_complete_overall.tex
\begin{table*}[t!]
\caption{{Overall mean values for synthetic sample quality. \textbf{Bold} and \underline{underlined} indicate the best and second-best models per metric. The \rlap{$^{*}$}~~marks results not significantly different from FlowLet-RFM 10 steps, based on pairwise Wilcoxon rank-sum tests with Bonferroni correction ($\alpha = 0.05$). Metrics are computed over 100 random bootstrap resamples of the full generated sets. Standard deviations ($\leq 10^{-3}$) are omitted for conciseness.}}
\centering

\begin{tabular}
{@{}p{0.48\textwidth}@{\hspace{0.04\textwidth}}p{0.48\textwidth}@{}}

\begin{minipage}[t]{\linewidth}

    % Table A
    \centering
    \subcaption{Ours (10 steps) vs. baselines}
    \label{tab:ours_vs_baselines}
        \setlength{\tabcolsep}{0.4mm} % Increased padding for better readability
    \resizebox{\linewidth}{!}{%
    \footnotesize
    \begin{tabular}
    {@{}ll@{\hspace{1mm}}c@{\hspace{7mm}}c@{\hspace{6mm}}c@{\hspace{5mm}}c@{}}

    \toprule
    & \textbf{Method} &\textbf{Steps} &\textbf{FID} $\downarrow$ & \textbf{MMD} $\downarrow$ & \textbf{MS-SSIM} $\downarrow$ \\
    \midrule
    \multirow{4}{*}{\rotatebox[]{90}{\scriptsize Ours}}
    &RFM &  10   & \underline{0.2981} & \underline{0.0119} & {0.9508}~~~ \\
    &CFM & 10    & 0.3098 & 0.0124 & 0.9707~~~ \\
    &VP &  10     & 0.3079 & 0.0123 & 0.9706~~~ \\
    &Trigon. & 10 & \textbf{0.2854} & \textbf{0.0114} & 0.9660~~~ \\
    \midrule
    \multirow{7}{*}{\rotatebox[]{90}{\scriptsize Baselines}}
    &WDM &1000   & 0.3073 & 0.0123 & 0.9456~~~ \\
    &{WDMa} &{1000}   & {0.3167} & {0.0123} & {\underline{0.9430}}~~~ \\
    &MD &1000    & 0.3843 & 0.0153 & 0.9595~~~ \\
    &MLDM &1000  & 0.3590 & 0.0144 & 0.9538~~~ \\
    &MOTFM &10   & 0.3696 & 0.0147 & 0.9676~~~ \\
    &MOTFMa &10 & 0.3747 & 0.0145 & 0.9528~~~ \\
    &{BS} &{--} & {0.3454} & {0.0138} & {\textbf{0.9346}}~~~ \\
    \bottomrule
    \end{tabular}%
    }

\includegraphics[width=\linewidth]{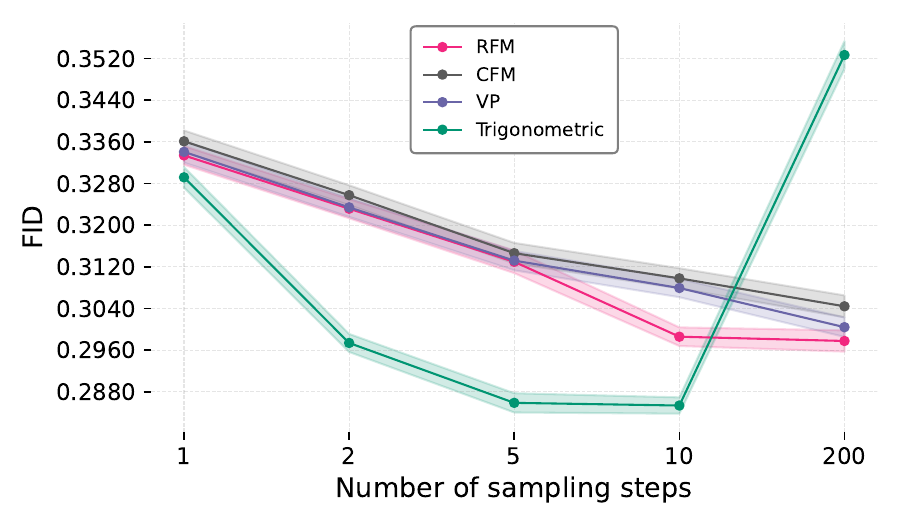}
\refstepcounter{figure}
{\sffamily\small\textbf{\color{scolor}Figure \thefigure:} FID vs. steps for FlowLet variants. The shaded bands indicate standard deviation.}
\label{fig:steps_ablation}
    
\end{minipage}
&
% --- Right Column: Table B ---
\begin{minipage}[t]{\linewidth}
    \centering
    \subcaption{Ablations of Ours (steps + conditioning)}
    \label{tab:ours_ablation}
    
    \resizebox{\linewidth}{!}{

    \begin{tabular}{lcccc}
    \toprule
    \textbf{Method} & \textbf{Steps} & \textbf{FID} $\downarrow$ & \textbf{MMD} $\downarrow$ & \textbf{MS-SSIM} $\downarrow$ \\
    \midrule
    \multirow{4}{*}{RFM}
     &1 & 0.3334 & 0.0133 & 0.9886 \\
     &2   & 0.3232 & 0.0129 & 0.9838 \\
     &5   & 0.3130 & 0.0125 & 0.9746 \\
     &200 & {0.2978}\rlap{$^{*}$} & \underline{0.0119}\rlap{$^{*}$} & \underline{0.9487} \\
     \midrule
     \multirow{1}{*}{{RFM DB4}}
     &{10} & {0.3141} & {0.0125} & {0.9663} \\
    \midrule
    \multirow{4}{*}{CFM}
     &1    & 0.3361 & 0.0134 & 0.9899 \\
     &2    & 0.3258 & 0.0130 & 0.9858 \\
     &5   & 0.3146 & 0.0126 & 0.9771 \\
     &200 & 0.3044 & 0.0122 & 0.9508\rlap{$^{*}$} \\
    \midrule
    \multirow{4}{*}{VP}
     &1     & 0.3341 & 0.0133 & 0.9898 \\
     &2    & 0.3234 & 0.0129 & 0.9858 \\
     &5     & 0.3132 & 0.0125 & 0.9771 \\
     &200   & 0.3004 & 0.0120 & 0.9513\rlap{$^{*}$} \\
    \midrule
    \multirow{4}{*}{Trigon.}
    &1 & 0.3292 & 0.0131 & \textbf{0.9211} \\
    &2 & 0.2974\rlap{$^{*}$} & 0.0119\rlap{$^{*}$} & 0.9521 \\
    &5 & \textbf{0.2859} & \textbf{0.0114} & 0.9680 \\
    &200 & 0.3527 & 0.0141 & 0.9557 \\
    \midrule
    \multirow{5}{*}{{Trigon. RK4}}
    &{1} & {\underline{0.2974}} & {\underline{0.0119}} & {0.9485} \\
    &{2} & {0.3112} & {0.0124} & {0.9576} \\
    &{5} & {0.3862} & {0.0154} & {0.9604} \\
    &{10} & {0.3820} & {0.0152} & {0.9579} \\
    &{200} & {0.3773} & {0.0151} & {0.9518} \\
    \midrule
    FiLM &10    & 0.3252 & 0.0130 & 0.9861 \\
    Spatial &10 & 0.3234 & 0.0129 & 0.9846 \\
    Uncond. &10 & 0.3181 & 0.0127 & 0.9803 \\
    \bottomrule
    \end{tabular}
    }
\end{minipage}
\end{tabular}
\label{tab:ours_sidebyside}
\end{table*}

%% file: TWOSIDED_TABLES_ROI_BAP.tex
\begin{table*}[t!]
\caption{{Downstream evaluations: (a) Brain Age Prediction (BAP); (b) Region-Based.}}
\centering

\begin{tabular}{@{}p{0.42\textwidth}@{\hspace{0.03\textwidth}}p{0.55\textwidth}@{}}

\begin{minipage}[t]{\linewidth}
    \centering
    \subcaption{{BAP for the Age $\geq$ 44 years group. Lower Mean Absolute Error (MAE, in years) = better accuracy. Ours: 10-step samples.}}
    \label{tab2a:brain_age_results_age_44}
    \vspace{0.5em}
    
    \setlength{\tabcolsep}{0.5mm} 
    \footnotesize
    \begin{tabular}{@{}ll@{\hspace{4mm}}c@{\hspace{3mm}}c@{}}
    \toprule
    & \textbf{Model} & \textbf{Train MAE} $\downarrow$ & \textbf{Test MAE} $\downarrow$ \\
    \midrule
    & Real Data  & $1.15 \pm 1.02$ & $4.91 \pm 3.92$ \\
    \midrule
    \multirow{4}{*}{\rotatebox[]{90}{\scriptsize Ours}}
    & RFM      & $1.46 \pm 0.59$ & $4.01 \pm 3.38$ \\
    & CFM      & $1.39 \pm 0.59$ & $4.06 \pm 3.37$ \\
    & VP       & $1.02 \pm 0.49$ & $4.68 \pm 3.78$ \\
    & Trigon.  & $1.09 \pm 0.48$ & $4.27 \pm 3.33$ \\
    \midrule
    \multirow{3}{*}{\rotatebox[]{90}{\scriptsize Ablat.}}
    & FiLM     & $0.57 \pm 0.51$ & $6.40 \pm 4.70$ \\
    & Spatial  & $0.87 \pm 0.54$ & $5.05 \pm 3.84$ \\
    & Uncond.  & $0.67 \pm 0.64$ & $5.65 \pm 3.74$ \\
    \midrule
    \multirow{7}{*}{\rotatebox[]{90}{\scriptsize Baselines}}
    & WDM      & $1.63 \pm 1.36$ & $\phantom{1}6.36 \pm 5.22$ \\
    & {WDMa}      & ${0.33 \pm 0.42}$ & ${\phantom{1}4.93 \pm 4.09}$ \\
    & MD       & $2.54 \pm 2.78$ & $\phantom{1}7.62 \pm 6.40$ \\
    & MLDM     & $0.98 \pm 0.47$ & $\phantom{1}5.30 \pm 3.86$ \\
    & MOTFM    & $2.10 \pm 2.82$ & $10.88 \pm 9.58$ \\
    & MOTFMa  & $0.85 \pm 0.53$ & $\phantom{1}4.90 \pm 3.57$ \\
    & {BS}      & ${0.90 \pm 0.40}$ & ${\phantom{1}4.16 \pm 3.38}$ \\
    \bottomrule
    \end{tabular}
\end{minipage}

&

\begin{minipage}[t]{\linewidth}
    \centering
    \subcaption{{Segmentation quality: lower is better for iMAE/KLD, higher for DICE.}}
    \label{tab:ROI_table}
    \vspace{0.5em}

    \setlength{\tabcolsep}{0.5mm} 
    \footnotesize
    \begin{tabular}{@{}ll@{\hspace{1mm}}c@{\hspace{2mm}}c@{\hspace{2mm}}c@{}}

    \toprule
    & \textbf{Model} & \textbf{iMAE} $\downarrow$ & \textbf{KLD} $\downarrow$ & \textbf{DICE} $\uparrow$ \\
    \midrule
    \multirow{5}{*}{\rotatebox[]{90}{\scriptsize Ours}}

    &RFM     & $37.68 \pm 10.22$ & $0.855 \pm 0.599$ & $0.420 \pm 0.169$ \\
    &CFM     & $42.93 \pm 11.19$ & $1.395 \pm 1.058$ & $0.424 \pm 0.171$ \\
    &VP      & $37.61 \pm 10.20$ & $0.865 \pm 0.615$ & $0.423 \pm 0.171$ \\
    &Trigon. & $43.35 \pm 12.06$ & $1.614 \pm 1.277$ & $0.379 \pm 0.172$ \\
    &{Uncond.} & {$39.95 \pm 10.14$} & {$1.188 \pm 1.094$} & {$0.409 \pm 0.159$} \\

    \midrule
    \multirow{7}{*}{\rotatebox[]{90}{\scriptsize Baselines}}
    &WDM     & $47.52 \pm \phantom{1}9.45$ & $1.088 \pm 0.781$ & $0.368 \pm 0.160$ \\
     &WDMa    & ${56.43 \pm 14.44}$ & ${2.112 \pm 1.090}$ & ${0.383 \pm 0.172}$ \\
    &MD      & $38.44 \pm 10.44$ & $0.863 \pm 0.593$ & $0.294 \pm 0.156$ \\
    &MLDM    & $46.93 \pm 11.62$ & $1.040 \pm 0.645$ & $0.331 \pm 0.154$ \\
    &MOTFM   & $41.67 \pm 11.12$ & $0.915 \pm 0.620$ & $0.409 \pm 0.163$ \\
    &MOTFMa & $42.93 \pm 11.18$ & $1.394 \pm 1.058$ & $0.391 \pm 0.162$ \\
    &{BS} & ${43.90 \pm 11.53}$ & ${0.862 \pm 0.589}$ & ${0.356 \pm 0.158}$ \\
    \bottomrule
    \end{tabular}
\end{minipage}

\end{tabular}

\label{tab:downstream_sidebyside}
\end{table*}

%% file: ablation_section.tex
\subsection{Ablation studies}
\label{sec:ablation_studies}

Ablation studies were conducted to assess how sensitive FlowLet is to key design choices. Specifically, we analyze the impact of the wavelet basis, the number of inference steps, the conditioning mechanisms, and the numerical solver. Furthermore, these experiments help disentangle which components drive sample quality and downstream utility; full results (including age-stratified evaluations and statistical tests) are reported in the Appendix.

\paragraph{Wavelet Selection.}
We analyze the impact of the wavelet basis used in the invertible representation on reconstruction fidelity and generative performance. Among the evaluated wavelet families (Haar, Daubechies-4, Symlet-4, Coiflet-2, and Biorthogonal 3.3), Haar consistently yields the lowest round-trip reconstruction error (mean MAE $6.08 \times 10^{-8}$) and the most stable generative behavior. Replacing Haar with smoother bases, such as Daubechies-4, degrades global image quality and downstream performance, as reflected by higher FID and reduced BAP accuracy (Tables~\ref{tab:ours_ablation}, \ref{tab:brain_age_results_age_44} , and ~\ref{tab:roi_additional_experiments}). Based on these results, Haar provides the most reliable trade-off between reconstruction fidelity, stability, and computational efficiency, and is adopted as the default wavelet basis. Detailed reconstruction analyses and extended quantitative comparisons are reported in Appendix  \ref{sec:wavelet}.

\paragraph{Step count and sampling efficiency.}
We evaluate the impact of the number of ODE sampling steps on generative quality and computational cost. Across the FlowLet variants, most gains are achieved within the first 10 steps. For RFM, CFM, and VP, image fidelity improves rapidly at low step counts and then largely saturates, indicating that longer trajectories provide limited practical benefit in these formulations (Figure~\ref{fig:steps_ablation} and Table~\ref{tab:ours_ablation}).

A different behavior is observed for the Trigonometric formulation. While it attains favorable global metrics at low step counts, its performance does not continue to improve at high step counts and instead degrades at 200 steps. This deterioration is most clearly visible in the quantitative metrics and is also reflected in downstream and region-based evaluation. Thus, the step-count study indicates that the high-step regime is not uniformly beneficial across flow formulations and that the Trigonometric path exhibits a distinct instability at large integration steps.

Consistent trends are observed in downstream evaluation. BAP performance improves rapidly from 1 to 10 steps and then saturates, with no meaningful benefit at higher step counts for the stable formulations (Table~\ref{tab:ours_ablation}). Notably, even low-step configurations outperform the real-only baseline, indicating that FlowLet captures age-relevant anatomical information without requiring long inference trajectories.

These results identify 10 steps as an effective trade-off between sampling efficiency and generative quality for FlowLet. Detailed timing analyses, extended step-wise evaluations, and solver-dependent behaviors are reported in Appendix~\ref{Efficiency_benchmarks} and Appendix~\ref{Additional_qualitative}.

\paragraph{Effect of conditioning mechanisms.}
We analyze the contribution of the conditioning strategy used to inject age information into FlowLet. Specifically, we compare the full model, which combines FiLM and spatial conditioning via cross-attention, against ablated variants using only FiLM, only spatial conditioning, or no conditioning.

Removing either conditioning component degrades performance across multiple evaluation axes. While globally conditioned variants retain competitive FID and MMD scores, downstream BAP performance and region-based anatomical metrics are substantially affected (Tables~\ref{tab:ours_ablation} and~\ref{tab2a:brain_age_results_age_44}). In particular, models using only spatial conditioning or only FiLM exhibit higher BAP error and reduced anatomical consistency compared to the full model, whereas the unconditional variant performs worst overall.

\paragraph{Solver analysis.}
We further investigate whether the observed high-step degradation can be attributed primarily to the numerical solver used for ODE integration. In addition to the default Euler method, we therefore evaluate a fourth-order Runge-Kutta (RK4) solver for the Trigonometric flow formulation.

Using RK4 does not resolve the Trigonometric failure mode. Instead, the degradation persists across evaluation axes: the overall FID worsens from $0.3112$ at 2 steps to $0.3774$ at 200 steps, the OpenBHB validation test MAE increases from $4.10 \pm 3.17$ to $4.43 \pm 3.83$, and the region-based metrics deteriorate from $37.21 \pm 9.96$ to $44.28 \pm 11.05$ in iMAE and from $0.915 \pm 0.668$ to $1.664 \pm 1.027$ in KLD. These results show that increasing solver accuracy alone does not restore stable behavior for the curved Trigonometric path.

Taken together, the solver comparison indicates that the limitations observed for the Trigonometric formulation are linked more strongly to the geometry of the learned transport path than to solver discretization alone (Tables~\ref{tab:ours_ablation}, \ref{tab:brain_age_results_age_44}, and \ref{tab:roi_additional_experiments}).

\input{table_brain_age_with_steps}

%% file: table_brain_age_with_steps.tex
\begin{table}[t]
\caption{Brain Age Prediction Performance for the Age $\geq$ 44 years group on the merged dataset. Lower values indicate better prediction accuracy. Results are reported as Mean Absolute Error (MAE, in years) and Standard Deviation.}
\centering
\setlength{\tabcolsep}{1.8mm} % Reduced column separation
\footnotesize
\sisetup{table-format=2.2(2), uncertainty-mode=separate}
\begin{tabular}{lc
                S[table-format=1.2(2)]
                S[table-format=2.2(2)]
                }
\toprule
\textbf{Model} & \textbf{Steps} & {\textbf{Train MAE} $\downarrow$} & {\textbf{Test MAE} $\downarrow$} \\
\midrule
Real Training Data &  & {1.15 $\pm$ 1.02} &{  4.91 $\pm$ 3.92 } \\
\midrule
\multirow{5}{*}{RFM}
& 1     & {1.32 $\pm$ 0.91}&{ 4.81 $\pm$ 4.43 }\\
& 2    &{ 1.03 $\pm$ 0.61 }&{ 4.74 $\pm$ 3.85 } \\
& 5    &{ 1.42 $\pm$ 0.46 }&{ 4.23 $\pm$ 3.52 }\\
& 10   &{ 1.46 $\pm$ 0.59 }&{ 4.01 $\pm$ 3.38 }\\
& 200  &{ 1.83 $\pm$ 0.40 }&{ 4.80 $\pm$ 3.91 }\\
\midrule
\multirow{5}{*}{{Trigon. RK4}}
& {1}     & {{1.09 $\pm$ 0.45}} & {{4.26 $\pm$ 3.40}} \\
& {2}    & {{0.33 $\pm$ 0.40}} & {{4.10 $\pm$ 3.17}} \\
& {5}    & {{0.93 $\pm$ 0.56}} & {{4.12 $\pm$ 3.73}} \\
& {10}   & {{0.76 $\pm$ 0.41}} & {{4.27 $\pm$ 3.49}} \\
& {200}  & {{0.63 $\pm$ 0.42}} & {{4.43 $\pm$ 3.83}} \\
\midrule
{RFM db4} & {10} & {{0.50 $\pm$ 0.35}} & {{4.61 $\pm$ 4.33}} \\
\bottomrule
\end{tabular}
\label{tab:brain_age_results_age_44}
\end{table}

\begin{table}[t]
\caption{Segmentation (ROI) quality metrics for the additional experiments. Lower values indicate better performance for iMAE and KLD; Higher values indicate better performance for DICE.}
\centering
\setlength{\tabcolsep}{1mm} 
\footnotesize
\sisetup{table-format=2.2, uncertainty-mode=separate}

\begin{tabular}{lc
                S[table-format=2.2(4)] % iMAE column
                S[table-format=1.3(2)] % KLD column
                S[table-format=1.3(2)] % DICE column
                }
\toprule
\textbf{Model} & \textbf{Steps} & {\textbf{iMAE} $\downarrow$} & {\textbf{KLD} $\downarrow$} & {\textbf{DICE} $\uparrow$} \\
\midrule
\multirow{5}{*}{Trigon. RK4}

& 1    & 40.14 \pm 10.53 & 1.186 \pm 0.676 & 0.038 \pm 0.099 \\
& 2    & 37.21 \pm 9.96  & 0.915 \pm 0.668 & 0.401 \pm 0.182 \\
& 5    & 37.43 \pm 10.27 & 0.868 \pm 0.628 & 0.407 \pm 0.174 \\
& 10   & 38.63 \pm 10.50 & 0.969 \pm 0.707 & 0.370 \pm 0.166 \\
& 200  & 44.28 \pm 11.05 & 1.664 \pm 1.027 & 0.394 \pm 0.172 \\
\midrule
RFM db4 & 10 & 40.48 \pm 10.85 & 1.065 \pm 0.751 & 0.427 \pm 0.171 \\
\bottomrule
\end{tabular}
\label{tab:roi_additional_experiments}
\end{table}

%% file: supplementary_corpus.tex
\appendix
\section{Appendix}
This section provides additional details and extended experiments supporting the main work, including Flow Matching implementations, hyperparameters and efficiency benchmarks, step-wise qualitative results, age-stratified metrics, wavelet analyses, and significance testing.
\label{Appendix}

\subsection{Flow Matching implementations}
\label{A_fm_implementation}
This subsection clarifies the implementation of the flow matching formulations. In all cases, the training objective is to minimize the MSE loss between a predicted velocity field \(v_\theta\) and a target velocity field \(v_{\mathrm{target}}\). For each training step, a time value \(t\) is sampled uniformly from \([0, 1]\). The data sample \(x_1\) corresponds to the variable \texttt{x1\_wavelet} in the code, and the noise sample \(x_0 \sim \mathcal{N}(0, I)\) is represented by variables named \texttt{x0\_wavelet}.

\paragraph{Rectified Flow Matching (RFM).}
The implementation directly translates the linear interpolation path and constant velocity from the paper. The path $x_t = (1 - t) x_0 + t x_1$ is computed as \texttt{xt = (1 - t\_broadcast) * x0\_wavelet + t\_broadcast * x1\_wavelet}. The target velocity \(v_{\mathrm{target}} = x_1 - x_0\) corresponds to the variable \texttt{v\_target = x1\_wavelet - x0\_wavelet}.

\paragraph{Conditional Flow Matching (CFM).}
The state-dependent target velocity \(v_{\mathrm{target}} = \frac{x_1 - x_t}{1 - t + \epsilon}\) is implemented as \texttt{v\_target = (x1\_wavelet - xt) / (1 - t\_broadcast + 1e-8)}. The term \(x_t\) is computed via the same linear interpolation as in RFM, and a small \(\epsilon = 10^{-8}\) is used for numerical stability.

\paragraph{Trigonometric Flow.}
The circular interpolation path \(x_t = \cos(\frac{\pi}{2} t) x_0 + \sin(\frac{\pi}{2} t) x_1\) is implemented as \texttt{xt = torch.cos(angle) * x0\_wavelet + torch.sin(angle) * x1\_wavelet}, where \texttt{angle} represents \(\frac{\pi}{2}t\). The corresponding velocity field \(v_{\mathrm{target}}\) is computed as its time derivative: \texttt{v\_target = -torch.sin(angle) * (pi/2) * x0\_wavelet + torch.cos(angle) * (pi/2) * x1\_wavelet}.

\input{Variance_preserving_derivation}

\input{table_hyperparameters}
\input{table_hyperparameters_bap}

\begin{figure*}[h]
\centering
\includegraphics[width=1\linewidth]{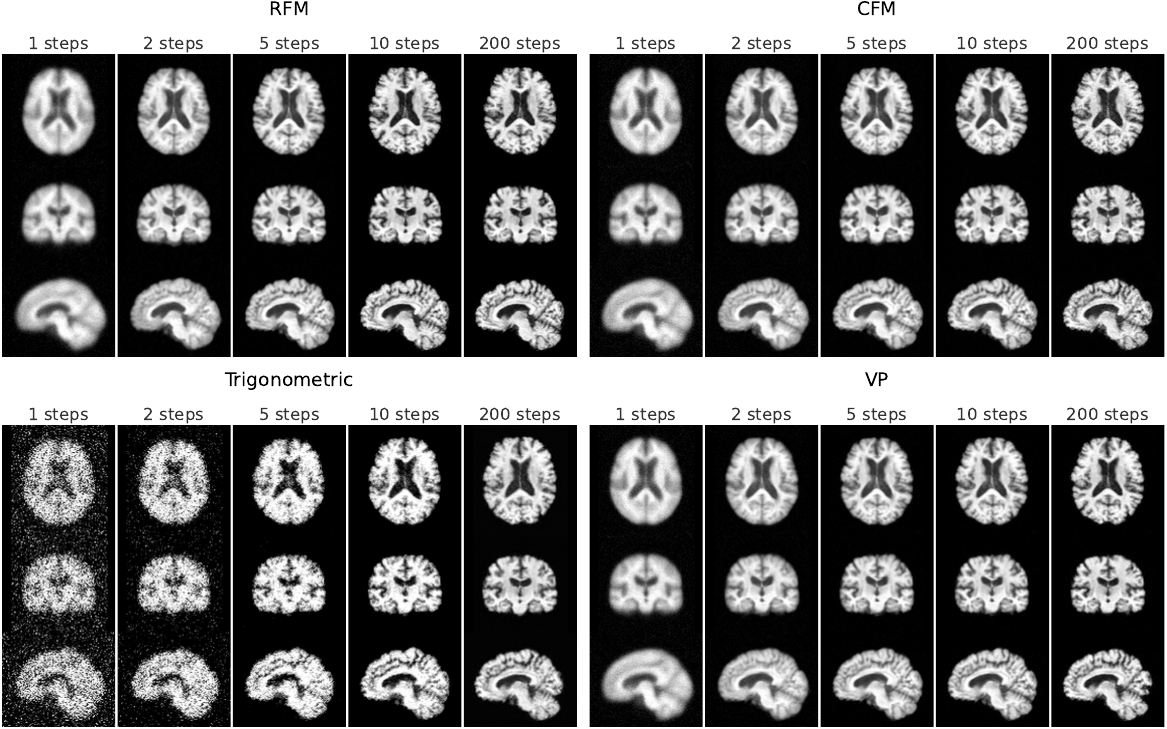}
\caption{Qualitative comparison of FlowLet flow matching formulation across different ODE step counts. Each column shows axial, coronal, and sagittal views for a given method and step count. All samples share the same noise seed and age condition, yielding anatomically consistent outputs with similar structural compartments across flow formulations.}\label{fig:Flow_comparison}
\end{figure*}

\subsection{Training hyperparameters}
\label{sec:hyperparams}
All models were trained using PyTorch\footnote{version 2.6.0
} on a system running Ubuntu 24.04, equipped with a single NVIDIA A6000 GPU (48GB).  Reproducibility was ensured  across the entire codebase by enforcing deterministic behavior across PyTorch, NumPy, and CuDNN. The training procedure was standardized across all Flow Matching formulations and baselines to ensure fair comparison. The training hyperparameters for FlowLet and the BAP model are reported in Table~\ref{tab:hyperparams} and Table~\ref{tab:hyperparams_bap}, respectively.

\paragraph{Data preparation and loading.}
The dataset was constructed from a centralized metadata file containing all MRI NIfTI file paths and associated subject metadata (e.g., age). Only subjects labeled as “Cognitively Normal” (CN) were retained. From this filtered dataset ($N=5{,}794$), a deterministic 80\%–20\% split was applied using a fixed random seed, resulting in 4,635 training and 1,159 validation subjects, ensuring no overlap across tasks. This split was used consistently across all training runs. The minimum and maximum age values were computed over the full dataset and used to normalize the age condition to the $[0, 1]$ range across all samples.

The preprocessing pipeline for generative modeling was applied to each 3D MRI volume and consisted of the following steps:
\begin{enumerate}
    \item Load the NIfTI volume and normalize intensities by clipping to the 0.5th and 99.5th percentiles, then scaling to the range $[-1, 1]$;
    \item Pad the volume to the required input size of $112 \times 112 \times 112$ using replication padding;
    \item For training samples, apply minimal, non-invasive data augmentation using the MONAI\footnote{\url{https://monai.io/} (1.4.0)} library, including random 3D rotations, intensity scaling, and Gaussian noise. These augmentations are designed to preserve anatomical structures while increasing data variance. No augmentations were applied to the validation set;
    \item Apply a 3D Haar DWT, implemented using the PyWavelets\footnote{\url{https://pywavelets.readthedocs.io/} (1.8.0)} library, to convert the single-channel input into an 8-channel tensor representing approximation and detail subbands. This tensor is directly used as input to the U-Net.
\end{enumerate}

\paragraph{Training dynamics.}
The model was trained for 200 epochs using the AdamW optimizer with an initial learning rate of $3 \times 10^{-6}$, decayed via cosine annealing to a minimum of $1 \times 10^{-7}$. Automatic mixed precision (AMP) training was enabled, using a \emph{GradScaler} to ensure numerical stability. Memory-efficient attention layers from the \texttt{xformers}\footnote{\url{https://github.com/facebookresearch/xformers} (0.0.29.post3)} library were used to reduce memory consumption. Gradients were clipped to a maximum L2 norm of 1.0. The model was trained using MSE loss between the predicted and target velocity fields in the wavelet domain.

Final model selection was based on the lowest validation MSE over the complete training epochs.

The full implementation is provided in the \texttt{FlowLet\_CODE} folder of the codebase.

\subsection{{Efficiency benchmarks}}\label{Efficiency_benchmarks}
{
To evaluate the computational efficiency of FlowLet, we report peak VRAM usage, training time, and inference throughput for all major baselines in Table~\ref{tab:resources_comparison}. FlowLet requires substantially less memory during training (approximately $22,\text{GB}$ with a batch size of 4), compared to over $40,\text{GB}$ for diffusion-based baselines such as WDM and MLDM. This reduced memory footprint allows FlowLet to be trained and deployed on consumer GPUs (e.g., RTX 3090/4090). Inference is also considerably faster: FlowLet produces a full $3$D sample in 1.57 seconds using 10 ODE steps, representing roughly a $45\times$ speedup over the conditional diffusion baseline WDMa (about 70 seconds for 1000 steps).}

\input{table_vram}

{
We further assess how FlowLet scales with increasing spatial resolution, as summarized in Table~\ref{tab:resolution_ablation}. FlowLet scales efficiently, requiring 42 GB of VRAM to train at $256^3$ with batch size 1. This behavior is largely due to operating in the wavelet domain. Sampling times remain stable across resolutions, increasing from roughly 1.6s per volume at $112^3$ to 6.8s at $256^3$ (10 ODE steps).}
\input{table_vram_resolutions}

\subsection{Additional qualitative assessment of Flow Variants}\label{Additional_qualitative}
To assess the generative behavior of different flow matching formulations, synthetic brain MRIs were generated using FlowLet with RFM, CFM, VP, and Trigonometric flows. For each method, volumes were sampled using 1, 2, 5, 10, and 200 ODE solver steps, with a fixed random seed and constant age condition (normalized age value $0.5\in[0,1]$, corresponding to 51 years). This controlled setup isolates the effect of the flow formulation and integration efficiency on the resulting images. Figure~\ref{fig:Flow_comparison} displays representative axial, coronal, and sagittal mid-slices for qualitative comparison. At high step counts (e.g., 200 steps), all methods converge toward anatomically plausible structures, confirming consistency in the asymptotic regime. In contrast, notable differences arise in low-step regimes: RFM, CFM, and VP produce stable and coherent volumes with as few as 2–5 steps, while the Trigonometric flow shows instability and structural artifacts. These results illustrate the trade-off between integration curvature and sampling stability, emphasizing the advantages of lower-curvature flows for efficient and anatomically faithful synthesis.

The sample generation process with FlowLet is implemented in \texttt{scripts/generate\_linear.py}.

\subsection{Additional age-stratified quantitative evaluation}\label{age_stratified_quantitative_evaluation}

\paragraph{Benchmark.} 
Three non-overlapping age bins were considered: 15--30, 40--55, and 65--80 years. 
For each generative model and each age bin, 200 synthetic brain MRI volumes were generated.
Age conditions were uniformly sampled within each bin.
All metrics (FID, MMD, and intra-set MS-SSIM) were computed independently for each age group using identical feature extractors and evaluation settings as in the main experiments.

Table~\ref{tab:benchmark_merged_all_ages} (overall,  15–30, 40–55 and 65–80 age groups) reports the complete quantitative results.  Cells marked with a dash ("–") under the unconditional (\emph{Uncond.}) baselines (RFM-Uncond., WDM, MOTFM and MD) indicate that age-stratified metrics could not be computed, as these models lack explicit age conditioning and therefore cannot be evaluated within the defined age range. 

The performance trend observed in the overall evaluation is consistently reflected across all age groups. Among the evaluated age ranges, the 65–80 group achieves the best performance in terms of both FID and intra-set MS-SSIM, indicating that FlowLet produces high-fidelity and diverse samples precisely in the demographic segment where data augmentation is most needed. This group exhibits the lowest MS-SSIM values across all age bins, suggesting increased anatomical variability in later adulthood, potentially reflecting diverse neuroanatomical aging trajectories~\cite{Bethlehem2022-em}. In contrast, higher MS-SSIM scores observed in younger age groups likely reflect more homogeneous structural patterns. 

To complement the results presented in the main text, Table \ref{tab:ablation_rfm_trigon} provides the complete quantitative results for the new ablation experiments, including the performance of FlowLet-RFM trained with the Daubechies-4 (db4) basis and the Trigonometric flow variant integrated using the 4th-Order Runge-Kutta (RK4) solver.
The performance degradation observed in the overall metrics for the RFM-db4 variant (FID 0.3142 vs. 0.2981 for Haar) is consistent across all age bins (Table \ref{tab:ablation_rfm_trigon}), confirming that the superior reconstruction fidelity of the Haar basis translates into better generalized generative performance, even in age-stratified contexts.
Similarly, the RK4 integration of the Trigonometric path, while intended to improve stability, failed to yield superior results compared to the simpler Euler-integrated RFM, often showing inferior FID/MMD across multiple step counts and age groups.

These findings underscore the importance of contextualizing intra-set MS-SSIM values with respect to the demographic characteristics of the generated data. All comparisons are reported within age-matched intervals to ensure fair and meaningful evaluation of generative diversity.

\begin{figure*}[t]
\centering
\includegraphics[width=1\linewidth]{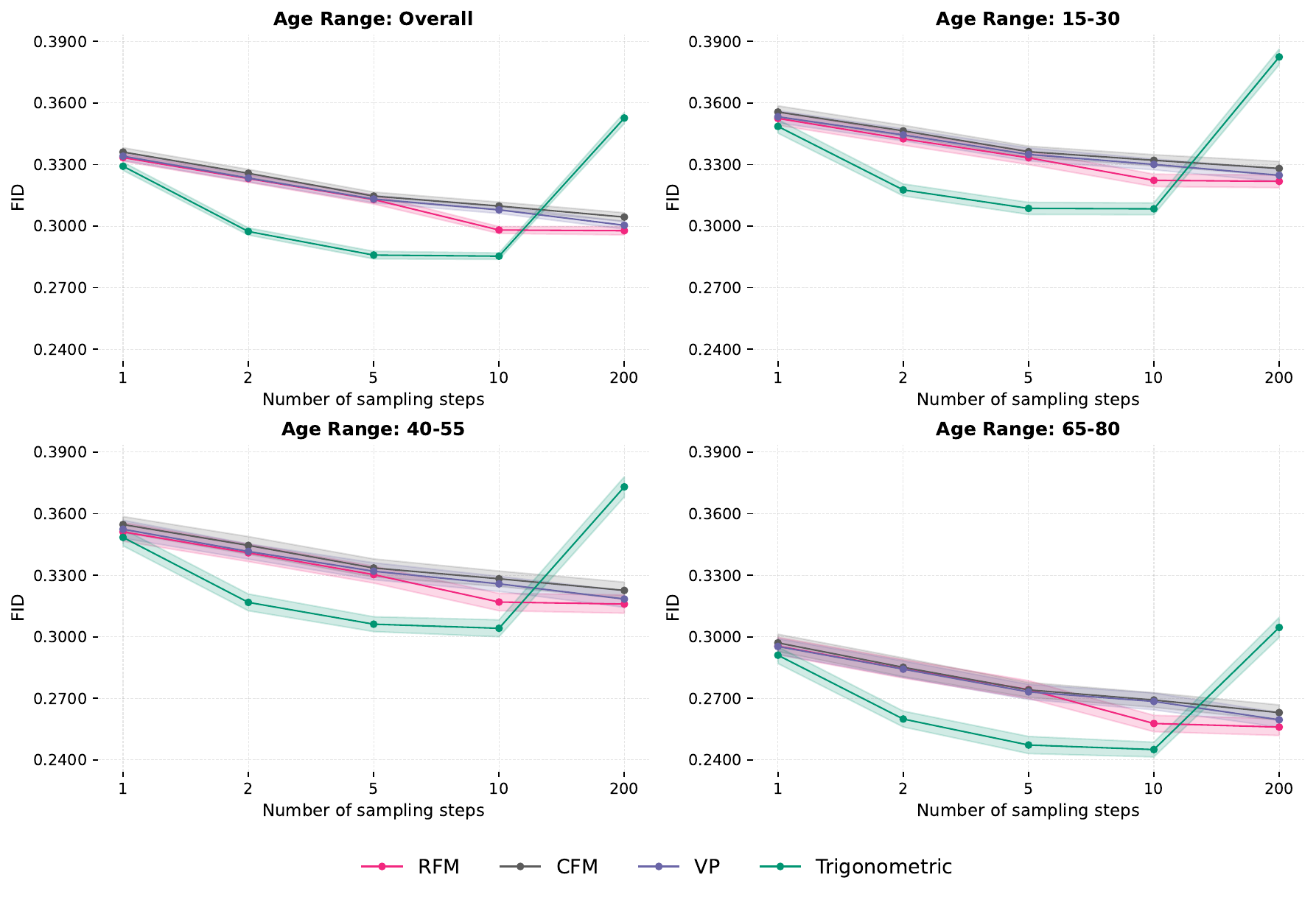}
\caption{{
FID vs. number of sampling steps for FlowLet variants calculated on overall (5.9-95.5 years) and by age range. We report FID as a function of the sampling budget for RFM, CFM, VP, and Trigonometric formulations. Shaded regions denote standard deviation across samples. All methods improve as steps increase, reaching marginal improvements after 10 steps. The Trigonometric variant consistently becomes unstable at 200 steps.}}
\label{fig:FIDvsSamples}
\end{figure*}

\paragraph{FID \emph{vs.} sampling steps.} Figure~\ref{fig:FIDvsSamples} plots the FID as a function of ODE steps for all FlowLet variants, stratified by age group.  All curves exhibit a consistent monotonic improvement with increasing step counts, in agreement with trends observed in the aggregate metrics. A performance plateau is observed between 10 and 200 steps, supporting the selection of 10 steps as an optimal balance between sampling efficiency and image fidelity across age groups. Notably, the Trigonometric flow, despite achieving competitive FID scores at low step counts, shows greater variability and degraded performance in high-step regimes.

\label{sec:age_stratified_metrics}

\begin{center}
\includegraphics[width=\columnwidth]{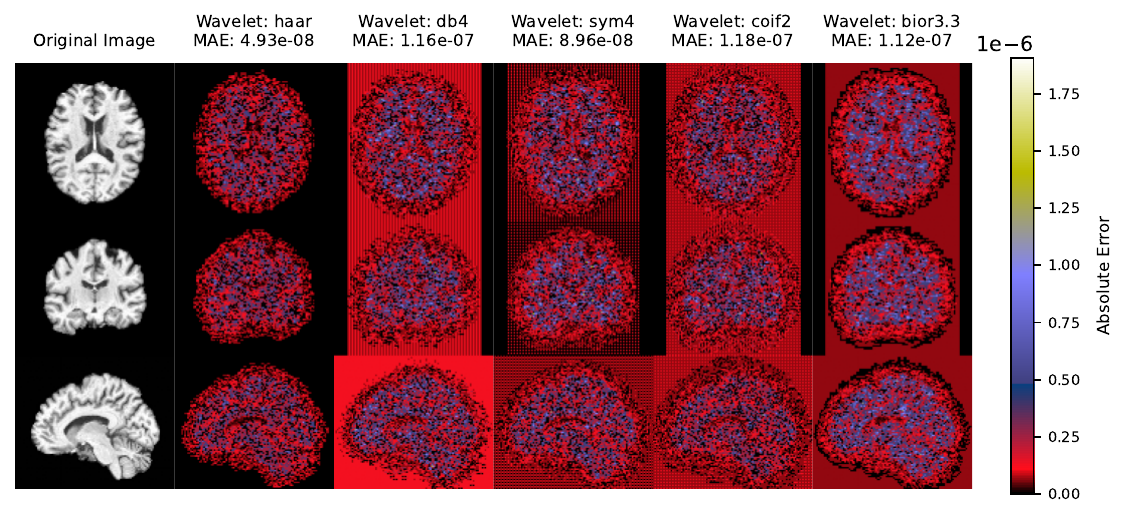}
\captionof{figure}{The leftmost column shows a real MRI volume in three views (axial, coronal, sagittal). Subsequent columns display the absolute reconstruction error for the same sample after a round-trip transform using different wavelet bases. The reported MAE values were computed over the entire set of voxel intensities in the 3D volume for this single instance. The color scale is adjusted to emphasize the narrow dynamic range of errors and facilitate visual comparison.}
\label{fig:Wavelet_comparison}
\end{center}

\subsection{{Significance testing and P-Value analysis.}}
\label{sec:pvalues}
{To assess the reliability of the performance differences observed in our study, we computed statistical significance relative to the reference configuration (FlowLet-RFM at 10 steps). These p-values correspond to the global quality metrics (FID, MMD, and MS-SSIM) reported in Table~\ref{tab:ablation_rfm_trigon} and Table~\ref{tab:benchmark_merged_all_ages} and are shown in Figure~\ref{fig:pvalue_heatmap}. In the heatmap, cells colored in red indicate results that are non-statistically significant ($p>0.05$). Notably, the prevalence of red cells in the 200-step rows for the RFM, CFM, and VP solvers confirms that extending the inference trajectory beyond 10 steps yields statistically negligible improvements in sample quality for these methods. Conversely, the remaining colored regions (green to yellow) represent statistically significant differences, validating the distinct performance gains achieved by the Trigonometric solver and the variations observed in the baseline comparisons.}

\input{Wavelet_comparison_table}
\input{table_metrics_rfmdb4_trigonrk4}

\subsection{Additional wavelet evaluation}
 \label{sec:wavelet}

Five wavelet families were compared, Haar, Daubechies-4 (db4), Symlet-4 (sym4), Coiflet-2 (coif2), and Biorthogonal 3.3 (bior3.3), to assess their suitability for 3D MRI volume reconstruction. The analysis focused on filter structure, computational efficiency, reconstruction accuracy, and the presence of voxel-domain artifacts. Detailed properties, including filter definitions and theoretical foundations, are available in~\cite{DBLP:journals/pami/Mallat89, DBLP:books/siam/92/D1992}. Table~\ref{tab:wavelet_recon_error} reports quantitative reconstruction errors, and Figure~\ref{fig:Wavelet_comparison} visualizes error distributions for a representative sample.

\paragraph{Haar (db1).} The Haar wavelet (Daubechies-1) is the simplest wavelet, defined by a step-function basis. It is discontinuous and uses a length-2 filter, providing the shortest support and highest computational efficiency. Haar achieved the lowest reconstruction error (mean MAE: $6.08\times10^{-8}$), with minimal boundary artifacts and numerically exact reconstruction. Although its constant basis functions can cause blockiness under compression, no such artifacts were observed under lossless reconstruction.

\paragraph{Daubechies (db4).} Daubechies-4 is an orthonormal wavelet with 4 vanishing moments and an 8-tap filter. Its smoother basis functions improve energy compaction and reduce blockiness compared to Haar. However, its longer support can introduce ringing near sharp transitions. In the experiments, db4 showed reconstruction errors roughly an order of magnitude higher than Haar (MAE $\sim$$10^{-7}$), likely due to boundary effects.

\paragraph{Symlets (sym4).} Sym4 is a near-symmetric variant of db4, preserving orthonormality and using an 8-tap filter. Its symmetry helps reduce phase distortion and shift sensitivity. Like db4, its reconstruction errors remained in the $10^{-7}$ range and were primarily localized in background voids.

\paragraph{Coiflets (coif2).} Coiflet-2 uses a 12-tap, near-symmetric filter with vanishing moments in both wavelet and scaling functions. It enables smooth intensity transitions at higher computational cost and boundary sensitivity. It achieved the second-lowest reconstruction error (MAE $\sim$$10^{-8}$), though some oscillatory artifacts appeared near high-contrast edges.

\paragraph{Biorthogonal (bior3.3).} Bior3.3 employs separate analysis and synthesis filters with three vanishing moments each and an 8-tap symmetric analysis filter. This biorthogonal design ensures linear phase and supports shift-invariant, edge-aligned reconstruction. The MAE was on par with other 8-tap wavelets ($\sim$$10^{-7}$), with mild structured artifacts near edges, likely due to non-orthogonality.

\paragraph{Qualitative reconstruction analysis.}
Figure~\ref{fig:Wavelet_comparison} presents the absolute voxel-wise reconstruction error for a representative MRI volume. This qualitative view highlights spatial error patterns following a single round-trip transform. Haar shows minimal, localized error within the brain, while the other bases introduce structured artifacts in background voids, reflecting longer filter supports or non-orthogonal behavior. These visual differences are consistent with the quantitative findings in Table~\ref{tab:wavelet_recon_error}.
Among all evaluated wavelets, Haar consistently achieved the best trade-off between reconstruction fidelity, computational cost, and artifact suppression, especially in low-signal regions, supporting its choice as the default basis for FlowLet.

\input{table_metrics_BIG_MERGE}
{
\clearpage
\onecolumn
\includegraphics[width=1\linewidth]{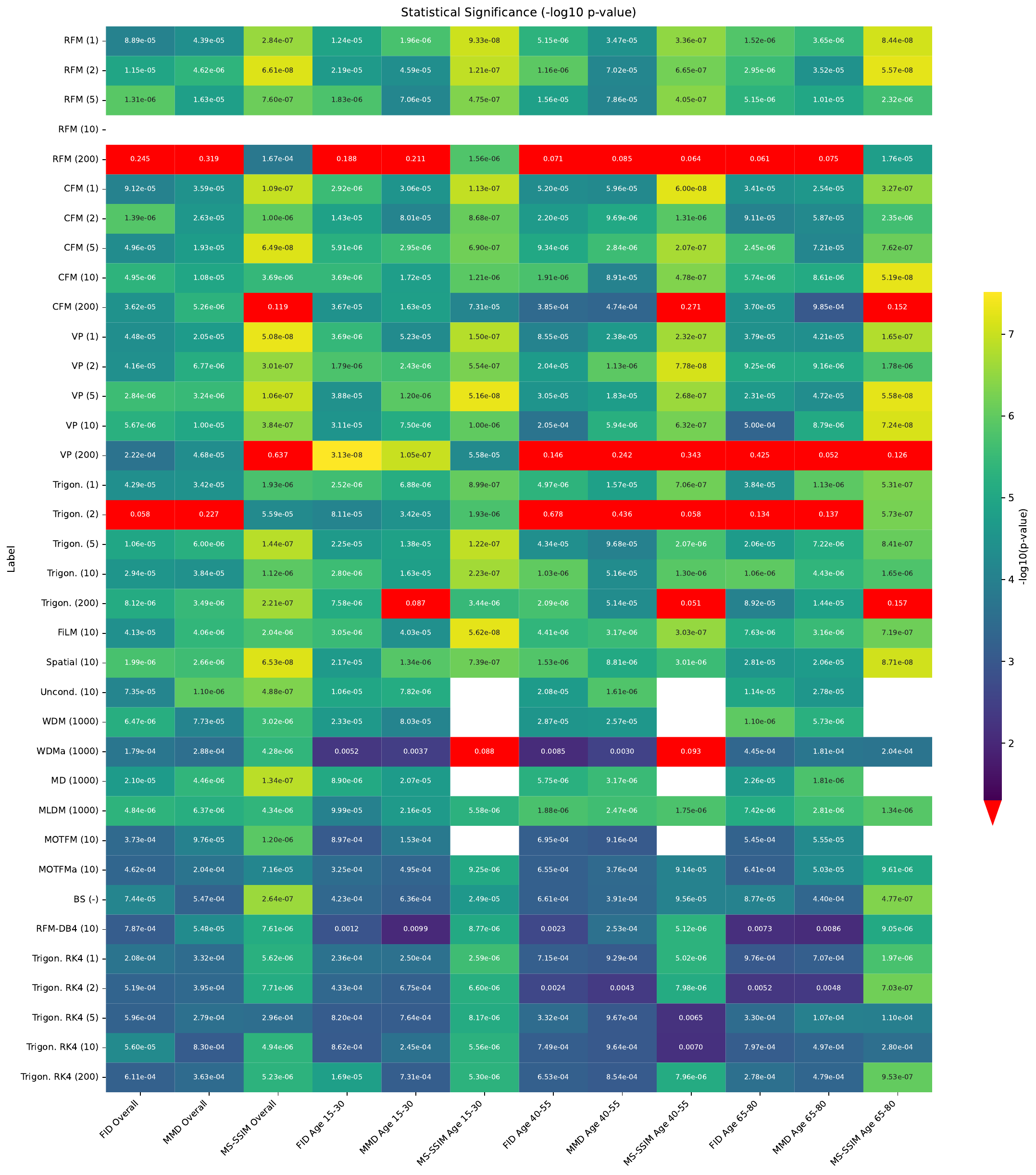}
\captionof{figure}{{Statistical Significance ($-\log_{10}$ p-value) of Pairwise Comparisons against FlowLet-RFM (10 Steps). The heatmap displays the results of the Bonferroni-corrected p-values obtained from two-sided Wilcoxon rank-sum tests for pairwise comparisons between the indicated models and the FlowLet-RFM (10 steps) baseline across various fidelity, diversity, and age-stratified metrics. Cells colored red indicate a non-statistically significant difference ($p > 0.05$) from the FlowLet-RFM (10 steps) baseline.}}
\label{fig:pvalue_heatmap}
\twocolumn
\clearpage
}

\input{rebuttal_experiments}

%% file: Variance_preserving_derivation.tex
\subsection{Variance-Preserving (VP) Diffusion Matching}
\label{A_VP_derivation}
This section presents the explanation of the VP Diffusion Matching implementation, establishing the connection between the high-level SDE formulation from the main text and the exact conditional velocity field used for training, as implemented in the codebase. The derivation summarises and follows rigorous treatments found in the literature, particularly~\cite{DBLP:conf/iclr/0011SKKEP21, DBLP:conf/iclr/LipmanCBNL23, DBLP:journals/corr/abs-2302-00482,gagneux2025a}.

\paragraph{Core definitions and time conventions.}
The VP formulation is defined over a continuous data-to-noise time interval \(t \in [0, 1]\), where \(t=0\) corresponds to clean data and \(t=1\) to pure noise. The process is governed by a linear variance schedule \(\beta(t)\):
{
\small
\begin{equation}
\beta(t) = \beta_{\min} + t(\beta_{\max} - \beta_{\min}),
\end{equation}
}
where our implementation uses the standard hyperparameters \(\beta_{\min}=0.1\) and \(\beta_{\max}=20.0\)~\cite{DBLP:conf/iclr/SongME21}. From this schedule, we define the signal and noise scaling coefficients, \(\bar{\alpha}(t)\) and \(\sigma(t)\), as follows:
{
\small
\begin{equation}
\bar{\alpha}(t) = \exp\left(-\frac{1}{2} \int_0^t \beta(s) \, ds\right), \quad \text{and} \quad \sigma(t) = \sqrt{1 - \bar{\alpha}(t)^2}.
\end{equation}
}
A noised sample \(x_t\) is generated from a real sample \(x_1\) via the interpolation \(x_t = \bar{\alpha}(t) x_1 + \sigma(t) \xi\), where \(\xi \sim \mathcal{N}(0, \mathbf{I})\).

Crucially, FlowLet training loop samples a time \(t_{\text{flow}} \in [0, 1]\) along the generative noise-to-data path. This is mapped to the theoretical data-to-noise time via the relation \(t = 1 - t_{\text{flow}}\). All formulas below use the theoretical time \(t\).

\paragraph{From stochastic SDE to deterministic ODE.}
To make explicit the link between the VP diffusion model and the velocity field
used in FlowLet, we first recall the full reverse-time SDE associated with the
VP forward process ~\cite{DBLP:conf/iclr/0011SKKEP21}:
{
\small
\begin{equation}
dx_t = \left[
-\tfrac12 \beta(t) x_t - \beta(t)\, \nabla_{x_t}\log p_t(x_t) \right] dt
+ \sqrt{\beta(t)}\, d\bar W_t.
\end{equation}
}

For a generic diffusion SDE of the form 
\(
dx_t = f(x_t,t)\,dt + g(t)\,dW_t,
\)
the corresponding Probability Flow ODE that evolves the same marginals $p_t$
is ~\cite{DBLP:conf/iclr/SongME21}:
{
\small
\begin{equation}
\frac{dx_t}{dt}
=f(x_t,t) - \tfrac12\, g(t)g(t)^\top\, \nabla_{x_t}\log p_t(x_t).
\end{equation}
}

Applying this identity to the VP reverse SDE, with $g(t)=\sqrt{\beta(t)}I$, 
yields the deterministic VP flow:
{
\small
\begin{equation}
v_{\mathrm{ODE}}(x_t, t) = -\frac{1}{2}\beta(t)x_t - \frac{1}{2}\beta(t)\nabla_{x_t} \log p_t(x_t).
\label{eq:ode_velocity_supp}
\end{equation}
}
which explains the factor $1/2$ multiplying the score term. This is the 
velocity field that FlowLet is trained to approximate.

\paragraph{Deriving the computable target velocity.}
{Directly using Eq.~\ref{eq:ode_velocity_supp} for training is intractable because the true score, \(\nabla_{x_t} \log p_t(x_t)\), is unknown. In the conditional flow matching framework, we circumvent this by computing the velocity conditioned on the target data sample \(x_1\). This is achieved by substituting the intractable score with the analytical conditional score, \(\nabla_{x_t} \log p_t(x_t|x_1)\), whose closed form is given by Tweedie's formula~\cite{Efron2011-ui}:}
{
\small
\begin{equation}
\nabla_{x_t} \log p_t(x_t|x_1) = -\frac{x_t - \bar{\alpha}(t)x_1}{\sigma(t)^2}.
\end{equation}
}
{Substituting this into our ODE velocity field (Eq.~\ref{eq:ode_velocity_supp}) gives the final, computable target velocity that our network learns to predict. This connection is rigorously established in recent flow matching literature~\cite{DBLP:conf/iclr/LipmanCBNL23, DBLP:journals/corr/abs-2302-00482, gagneux2025a}:}
{
\small
\begin{align}
v_{\mathrm{target}}(x_t, t \mid x_1) 
&= -\frac{1}{2}\beta(t)x_t - \frac{1}{2}\beta(t)\left(-\frac{x_t - \bar{\alpha}(t)x_1}{\sigma(t)^2}\right) \notag \\
&= -\frac{1}{2}\beta(t)\left(x_t - \frac{x_t - \bar{\alpha}(t)x_1}{1 - \bar{\alpha}(t)^2}\right) \notag \\
&=  -\frac{1}{2}\beta(t) \frac{\bar{\alpha}(t)^2 x_t - \bar{\alpha}(t) x_1}{1-\bar{\alpha}(t)^2}.
\label{eq:final_target_supp}
\end{align}
}

\paragraph{Equivalence to the code implementation.}
{The target velocity in Eq.~\ref{eq:final_target_supp} is mathematically equivalent to our implementation, which adopts a numerically stable reformulation of the same expression.  We recall that $\bar{\alpha}(t) = e^{-\frac{1}{2}T(t)}$ and $\bar{\alpha}(t)^2 = e^{-T(t)}$. Additionally, in the generative training loop, the time variable is reversed as $ t_{\text{flow}} = 1 - t$. 
Finally, Eq.~\ref{eq:final_target_supp} takes the following form, as implemented in FlowLet.}
{\small
\begin{equation}
v_{\text{target}}(x_t, t \mid x_1) = -\frac{1}{2}\beta(1-t)\frac{e^{-T(1-t)} x_t - e^{-\frac{1}{2}T(1-t)} x_1}{1 - e^{-T(1-t)}}.
\end{equation}
}

All the implemented formulations are available in \\\texttt{/flowlet/models/flow\_matching.py}.

%% file: table_hyperparameters.tex
\begin{table}[t]
\caption{Training Hyperparameters for FlowLet Models.}
\centering
\footnotesize
\begin{tabular}{ll}
\toprule
\textbf{Parameter} & \textbf{Value} \\ 
\midrule
\multicolumn{2}{c}{\textit{Optimizer \& Scheduler}} \\
Optimizer & AdamW \\
Learning Rate & 3e-6 \\
Weight Decay & 1e-5 \\
Adam $\beta_1, \beta_2$ & 0.9, 0.999 \\
LR Scheduler & CosineAnnealingLR \\
Scheduler \texttt{eta\_min} & 1e-7 \\ 
\midrule
\multicolumn{2}{c}{\textit{Training}} \\
Epochs & 200 \\
Batch Size & 4 \\
Gradient Clipping Norm & 1.0 \\
\midrule
\multicolumn{2}{c}{\textit{Model \& Architecture}} \\
U-Net Base Channels & 128 \\
U-Net Channel Multipliers & (1, 2, 4, 8) \\
U-Net Attention Resolutions & (4, 8) \\
U-Net Attention Heads & 8 \\
Condition Embedding Dim & 512 \\
\texttt{xformers} Attention & Enabled \\ 
Mixed Precision & Enabled \\
\bottomrule
\end{tabular}
\label{tab:hyperparams}
\end{table}

%% file: table_hyperparameters_bap.tex
\begin{table}[t]
\caption{Training Hyperparameters for the BAP Model.}
\footnotesize
\centering
\begin{tabular}{ll}
\toprule
\textbf{Parameter} & \textbf{Value} \\ 
\midrule
\multicolumn{2}{c}{\textit{Optimizer \& Scheduler}} \\
Optimizer & SGD \\
Initial Learning Rate & 0.01 \\
LR Scheduler & CosineAnnealingWarmRestarts \\
Scheduler \texttt{T\_0} & 17 \\
Scheduler \texttt{T\_mult} & 2 \\
Scheduler \texttt{eta\_min} & 1e-5 \\
\midrule
\multicolumn{2}{c}{\textit{Training}} \\
Epochs & 100 \\
Batch Size & 16 \\
Loss Function & L1 Loss (MAE) \\
\midrule
\multicolumn{2}{c}{\textit{Model \& Architecture}} \\
Model Architecture & DenseNet-121 \\
Input Channels & 1 \\
Output Classes & 1 (Age) \\
\midrule
\multicolumn{2}{c}{\textit{Data Preprocessing}} \\
Normalization Method & Min-Max Scaling \\
Scaling Min Value & 5th Percentile (of training set) \\
Scaling Max Value & 95th Percentile (of training set) \\
\bottomrule
\end{tabular}
\label{tab:hyperparams_bap}
\end{table}

%% file: table_vram.tex
\begin{table*}[t]
\color{black}{

\caption{{Computational Resources Analysis. Training times are for 200 epochs unless specified otherwise in parentheses. Inference time is measured for generating 3000 samples.}}
\centering
\footnotesize
\setlength{\tabcolsep}{1.8mm}
\begin{tabular}{l l c c c c}
\toprule
\textbf{Method} & \textbf{Stage / Component} & \textbf{Batch size} & \textbf{VRAM (GB)} & \textbf{Training Time} & \textbf{Inference Time} \\
\midrule
\textbf{FlowLet (Ours)} & Single Stage & 4 & 22 & 80 h & 1 h 18 min \\
\midrule
\multirow{2}{*}{MLDM} 
& Stage 1 & 2 & 39 & 120 h & \multirow{2}{*}{10 h} \\
& Stage 2 & 2 & 14 & 70 h (350 epochs) & \\
\midrule
MOTFM & Single Stage & 1 & 25 & 315 h & 1 h 40 min \\
\midrule
MOTFMa & Single Stage & 1 & 25 & 315 h & 1 h 40 min \\
\midrule
\multirow{4}{*}{MD} 
& Stage 1 & 4 & 16 & 74 h & \multirow{4}{*}{13 h 40 min} \\
& Stage 2 & 2 & 39 & 200 h & \\
& Stage 3 & 4 & 10 & 1 h & \\
& Stage 4 & 16 & 11 & 16 h 40 min & \\
\midrule
WDM & Single Stage & 1 & 40 & 96 h & 58 h 20 min \\
\midrule
WDMa & Single Stage & 1 & 43 & 101 h & 58 h 20 min \\
\midrule
\multirow{3}{*}{BS} 
& Stage 1 & 16 & 7 & 205 h (500 epochs) & \multirow{3}{*}{4 h 18 min} \\
& Stage 2 (Extraction) & -- & 4 & 1 h & \\
& Stage 3 & 2 & 5 & 27 h (500 epochs) & \\
\bottomrule
\end{tabular}
\label{tab:resources_comparison}
}
\end{table*}

%% file: table_vram_resolutions.tex
\begin{table}[h]
\caption{Ablation study on spatial resolution, memory consumption, and speed for FlowLet model. Peak VRAM usage is reported for Batch Size 1 and 4. Inference time is measured per sample. The entry "--" indicates required VRAM exceeds 48 GB.}
\centering
\footnotesize
\setlength{\tabcolsep}{1mm}
\begin{tabular}{c c c c}
\toprule
\textbf{Resolution} &
\textbf{\shortstack{Peak VRAM\\(Batch size 1)}} &
\textbf{\shortstack{Peak VRAM\\(Batch size 4)}} &
\textbf{Inference Time} \\
\midrule
$112^3$ & 18 GB & 22 GB & 1.57 s \\
$128^3$ & 23 GB & 31 GB & 2.12 s \\
$256^3$ & 42 GB & --    & 6.80 s \\
\bottomrule
\end{tabular}
\label{tab:resolution_ablation}
\end{table}

%% file: Wavelet_comparison_table.tex
\begin{table}
\centering
\caption{
Mean Absolute Error (MAE) and Standard Deviation (Std) computed over the entire training set ($N = 5{,}794$) for each wavelet. The error reflects voxel-wise intensity differences between original and reconstructed volumes.
}
\label{tab:wavelet_recon_error}
\small
\begin{tabular}{lc}
\toprule
\textbf{Wavelet Type} & \textbf{MAE} $\pm$ \textbf{Std} $\downarrow$ \\
\midrule
{Haar} & \textbf{\num{6.08e-8} $\pm$ \num{1.60e-8}} \\
Daubechies (db4) & \num{1.03e-7} $\pm$ \num{2.30e-8} \\
Symlet (sym4) & \num{1.11e-7} $\pm$ \num{2.81e-8} \\
Coiflet (coif2) & \num{9.98e-8} $\pm$ \num{2.24e-8} \\
Biorthogonal (bior3.3) & \num{1.32e-7} $\pm$ \num{3.70e-8} \\
\bottomrule
\end{tabular}
\end{table}

%% file: table_metrics_rfmdb4_trigonrk4.tex
\begin{table*}[h]
  \caption{Comparison of synthetic sample quality metrics for RFM DB4 and Trigon. RK4 variants. The metrics are computed over 100 bootstrap resamples.}
  \color{black}
  \centering
  \scriptsize
  \setlength{\tabcolsep}{1mm}

  % --- First Table (Overall + Age 15-30) ---
  \begin{tabular}{l l c | c c c | c c c}
    \toprule
    \textbf{Method} & \textbf{Type} & \textbf{Steps} &
    \multicolumn{3}{c|}{\textbf{Overall}} &
    \multicolumn{3}{c}{\textbf{Age 15–30}} \\
    & & & \textbf{FID} $\downarrow$ & \textbf{MMD} $\downarrow$ & \textbf{MS-SSIM} $\downarrow$ &
      \textbf{FID} $\downarrow$ & \textbf{MMD} $\downarrow$ & \textbf{MS-SSIM} $\downarrow$ \\
    \midrule
    Ours & RFM DB4 & 10 & 0.3142$~\pm~$0.0018 & 0.0125$~\pm~$0.0001 & 0.9663$~\pm~$0.0153 & 0.3376$~\pm~$0.0031 & 0.0136$~\pm~$0.0001 & 0.9822$~\pm~$0.0015 \\
    \midrule
    \multirow{5}{*}{Ours} & \multirow{5}{*}{Trigon. RK4} & 1 & 0.2974$~\pm~$0.0017 & 0.0119$~\pm~$0.0001 & 0.9485$~\pm~$0.0149 & 0.3202$~\pm~$0.0029 & 0.0129$~\pm~$0.0001 & 0.9632$~\pm~$0.0008 \\
      & & 2 & 0.3112$~\pm~$0.0022 & 0.0124$~\pm~$0.0001 & 0.9576$~\pm~$0.0132 & 0.3359$~\pm~$0.0031 & 0.0135$~\pm~$0.0001 & 0.9721$~\pm~$0.0009 \\
      & & 5 & 0.3862$~\pm~$0.0021 & 0.0154$~\pm~$0.0001 & 0.9604$~\pm~$0.0114 & 0.4145$~\pm~$0.0031 & 0.0166$~\pm~$0.0001 & 0.9743$~\pm~$0.0016 \\
      & & 10 & 0.3820$~\pm~$0.0022 & 0.0152$~\pm~$0.0001 & 0.9579$~\pm~$0.0129 & 0.4061$~\pm~$0.0035 & 0.0163$~\pm~$0.0001 & 0.9742$~\pm~$0.0028 \\
      & & 200 & 0.3774$~\pm~$0.0023 & 0.0151$~\pm~$0.0001 & 0.9518$~\pm~$0.0172 & 0.4061$~\pm~$0.0040 & 0.0163$~\pm~$0.0002 & 0.9728$~\pm~$0.0042 \\
    \bottomrule
  \end{tabular}

  \vspace{1mm}

  % --- Second Table (Age 40-55 + Age 65-80) ---
  \begin{tabular}{l l c | c c c | c c c}
    \toprule
    \textbf{Method} & \textbf{Type} & \textbf{Steps} &
    \multicolumn{3}{c|}{\textbf{Age 40–55}} &
    \multicolumn{3}{c}{\textbf{Age 65–80}} \\
    & & & \textbf{FID} $\downarrow$ & \textbf{MMD} $\downarrow$ & \textbf{MS-SSIM} $\downarrow$ &
      \textbf{FID} $\downarrow$ & \textbf{MMD} $\downarrow$ & \textbf{MS-SSIM} $\downarrow$ \\
    \midrule
    Ours & RFM DB4 & 10 & 0.3338$~\pm~$0.0043 & 0.0134$~\pm~$0.0002 & 0.9804$~\pm~$0.0027 & 0.2739$~\pm~$0.0039 & 0.0109$~\pm~$0.0002 & 0.9780$~\pm~$0.0035 \\
    \midrule
    \multirow{5}{*}{Ours} & \multirow{5}{*}{Trigon. RK4} & 1 & 0.3175$~\pm~$0.0041 & 0.0127$~\pm~$0.0002 & 0.9642$~\pm~$0.0010 & 0.2563$~\pm~$0.0037 & 0.0102$~\pm~$0.0001 & 0.9611$~\pm~$0.0018 \\
      & & 2 & 0.3307$~\pm~$0.0040 & 0.0132$~\pm~$0.0002 & 0.9716$~\pm~$0.0012 & 0.2685$~\pm~$0.0040 & 0.0107$~\pm~$0.0002 & 0.9676$~\pm~$0.0021 \\
      & & 5 & 0.4080$~\pm~$0.0049 & 0.0163$~\pm~$0.0002 & 0.9717$~\pm~$0.0021 & 0.3405$~\pm~$0.0048 & 0.0136$~\pm~$0.0002 & 0.9665$~\pm~$0.0029 \\
      & & 10 & 0.4003$~\pm~$0.0057 & 0.0160$~\pm~$0.0002 & 0.9697$~\pm~$0.0041 & 0.3394$~\pm~$0.0054 & 0.0136$~\pm~$0.0002 & 0.9614$~\pm~$0.0056 \\
      & & 200 & 0.3974$~\pm~$0.0050 & 0.0159$~\pm~$0.0002 & 0.9627$~\pm~$0.0095 & 0.3310$~\pm~$0.0045 & 0.0132$~\pm~$0.0002 & 0.9483$~\pm~$0.0129 \\
    \bottomrule
  \end{tabular}

  \label{tab:ablation_rfm_trigon}
\end{table*}

%% file: table_metrics_BIG_MERGE.tex
\clearpage
\onecolumn
\begin{center}
\captionsetup{type=table}
\captionof{table}{Synthetic sample quality metrics across different age groups. Bold and underlined values are best and second-best models. The $^*$ marks results not significantly different from FlowLet-RFM at 10 steps ($p > 0.05$). The metrics are computed over 100 bootstrap resamples. Values reported as "–" indicate configurations where age-specific metrics could not be computed, such as unconditional models lacking explicit age control.}
  \centering
  \scriptsize
  \setlength{\tabcolsep}{1mm}

  % --- First Table (Overall + Age 15-30) ---
  \begin{tabular}{l l c | c c c | c c c}
    \toprule
    \textbf{Method} & \textbf{Type} & \textbf{Steps} &
    \multicolumn{3}{c|}{\textbf{Overall}} &
    \multicolumn{3}{c}{\textbf{Age 15–30}} \\
    & & & \textbf{FID} $\downarrow$ & \textbf{MMD} $\downarrow$ & \textbf{MS-SSIM} $\downarrow$ &
      \textbf{FID} $\downarrow$ & \textbf{MMD} $\downarrow$ & \textbf{MS-SSIM} $\downarrow$ \\
    \midrule
    \multirow{5}{*}{Ours} & \multirow{5}{*}{RFM} & 1 & 0.3334$~\pm~$0.0018 & 0.0133$~\pm~$0.0001 & 0.9886$~\pm~$0.0102 & 0.3525$~\pm~$0.0033 & 0.0142$~\pm~$0.0001 & 0.9981$~\pm~$0.0002 \\
      & & 2 & 0.3232$~\pm~$0.0018 & 0.0129$~\pm~$0.0001 & 0.9838$~\pm~$0.0112 & 0.3425$~\pm~$0.0031 & 0.0138$~\pm~$0.0001 & 0.9942$~\pm~$0.0004 \\
      & & 5 & 0.3130$~\pm~$0.0022 & 0.0125$~\pm~$0.0001 & 0.9746$~\pm~$0.0121 & 0.3333$~\pm~$0.0033 & 0.0134$~\pm~$0.0001 & 0.9863$~\pm~$0.0008 \\
      & & 10 & 0.2981$~\pm~$0.0017 & \underline{0.0119}$~\pm~$0.0001 & 0.9508$~\pm~$0.0195 & 0.3223$~\pm~$0.0030 & \underline{0.0130}$~\pm~$0.0001 & 0.9698$~\pm~$0.0043 \\
      & & 200 & \underline{0.2978}\rlap{$^{*}$}$~\pm~$0.0020 & \underline{0.0119}\rlap{$^{*}$}$~\pm~$0.0001 & 0.9487$~\pm~$0.0206 & \underline{0.3218}\rlap{$^{*}$}$~\pm~$0.0031 & \underline{0.0130}\rlap{$^{*}$}$~\pm~$0.0001 & 0.9687$~\pm~$0.0047 \\
    \midrule
    \multirow{5}{*}{Ours} & \multirow{5}{*}{CFM} & 1 & 0.3361$~\pm~$0.0021 & 0.0134$~\pm~$0.0001 & 0.9899$~\pm~$0.0093 & 0.3556$~\pm~$0.0030 & 0.0143$~\pm~$0.0001 & {0.9978}$~\pm~$0.0005 \\
      & & 2 & 0.3258$~\pm~$0.0019 & 0.0130$~\pm~$0.0001 & 0.9858$~\pm~$0.0104 & 0.3464$~\pm~$0.0027 & 0.0139$~\pm~$0.0001 & 0.9945$~\pm~$0.0006 \\
      & & 5 & 0.3146$~\pm~$0.0020 & 0.0126$~\pm~$0.0001 & 0.9771$~\pm~$0.0111 & 0.3363$~\pm~$0.0027 & 0.0135$~\pm~$0.0001 & 0.9870$~\pm~$0.0009 \\
      & & 10 & 0.3098$~\pm~$0.0019 & 0.0124$~\pm~$0.0001 & 0.9707$~\pm~$0.0117 & 0.3321$~\pm~$0.0027 & 0.0134$~\pm~$0.0001 & 0.9815$~\pm~$0.0014 \\
      & & 200 & 0.3044$~\pm~$0.0021 & 0.0122$~\pm~$0.0001 & 0.9508\rlap{$^{*}$}$~\pm~$0.0182 & 0.3281$~\pm~$0.0034 & 0.0132$~\pm~$0.0001 &{0.9675}$~\pm~$0.0055 \\
    \midrule
    \multirow{5}{*}{Ours} & \multirow{5}{*}{VP} & 1 & 0.3341$~\pm~$0.0021 & 0.0133$~\pm~$0.0001 & 0.9898$~\pm~$0.0092 & 0.3533$~\pm~$0.0030 & 0.0142$~\pm~$0.0001 & 0.9979$~\pm~$0.0003 \\
      & & 2 & 0.3234$~\pm~$0.0018 & 0.0129$~\pm~$0.0001 & 0.9858$~\pm~$0.0101 & 0.3445$~\pm~$0.0029 & 0.0138$~\pm~$0.0001 & 0.9948$~\pm~$0.0004 \\
      & & 5 & 0.3132$~\pm~$0.0019 & 0.0125$~\pm~$0.0001 & 0.9771$~\pm~$0.0109 & 0.3349$~\pm~$0.0032 & 0.0135$~\pm~$0.0001 & 0.9871$~\pm~$0.0008 \\
      & & 10 & 0.3079$~\pm~$0.0018 & 0.0123$~\pm~$0.0001 & 0.9706$~\pm~$0.0115 & 0.3301$~\pm~$0.0026 & 0.0133$~\pm~$0.0001 & 0.9817$~\pm~$0.0013 \\
      & & 200 & 0.3004$~\pm~$0.0019 & 0.0120$~\pm~$0.0001 & 0.9513\rlap{$^{*}$}$~\pm~$0.0183 & 0.3248$~\pm~$0.0032 & 0.0131$~\pm~$0.0001 & 0.9685\rlap{$^{*}$}$~\pm~$0.0057 \\
    \midrule
    \multirow{5}{*}{Ours} & \multirow{5}{*}{Trigon.} & 1 & 0.3292$~\pm~$0.0020 & 0.0131$~\pm~$0.0001 & \textbf{0.9211}$~\pm~$0.0084 & 0.3486$~\pm~$0.0033 & 0.0140$~\pm~$0.0001 & \textbf{0.9292}$~\pm~$0.0004 \\
      & & 2 & 0.2974\rlap{$^{*}$}$~\pm~$0.0017 & 0.0119\rlap{$^{*}$}$~\pm~$0.0001 & 0.9521$~\pm~$0.0112 & 0.3176$~\pm~$0.0029 & 0.0128$~\pm~$0.0001 & 0.9624$~\pm~$0.0004 \\
      & & 5 & 0.2859$~\pm~$0.0019 & \textbf{0.0114}$~\pm~$0.0001 & 0.9680$~\pm~$0.0130 & 0.3086$~\pm~$0.0030 & \textbf{0.0124}$~\pm~$0.0001 & 0.9799$~\pm~$0.0008 \\
      & & 10 & \textbf{0.2854}$~\pm~$0.0016 & \textbf{0.0114}$~\pm~$0.0001 & 0.9660$~\pm~$0.0119 & \textbf{0.3084}$~\pm~$0.0029 & \textbf{0.0124}$~\pm~$0.0001 & 0.9775$~\pm~$0.0011 \\
      & & 200 & 0.3527$~\pm~$0.0027 & 0.0141$~\pm~$0.0001 & 0.9557$~\pm~$0.0165 & 0.3824$~\pm~$0.0038 & 0.0154$~\pm~$0.0002 & 0.9723\rlap{$^{*}$}$~\pm~$0.0039 \\
    \midrule
    & FiLM     & 10 & 0.3252$~\pm~$0.0020 & 0.0130$~\pm~$0.0001 & 0.9861$~\pm~$0.0008 & 0.3469$~\pm~$0.0029 & 0.0139$~\pm~$0.0001 & 0.9862$~\pm~$0.0007 \\
    Ours     & Spatial  & 10 & 0.3234$~\pm~$0.0017 & 0.0129$~\pm~$0.0001 & 0.9846$~\pm~$0.0022 & 0.3489$~\pm~$0.0030 & 0.0140$~\pm~$0.0001 & 0.9849$~\pm~$0.0011 \\
         & Uncond.  & 10 & 0.3181$~\pm~$0.0021 & 0.0127$~\pm~$0.0001 & 0.9803$~\pm~$0.0014 & 0.3384$~\pm~$0.0011 & 0.0136$~\pm~$0.0000 & -- \\
    \midrule
    WDM  & Uncond. & 1000 & 0.3073$~\pm~$0.0018 & 0.0123$~\pm~$0.0001 & {0.9456}$~\pm~$0.0248 & 0.3284$~\pm~$0.0013 & 0.0132$~\pm~$0.0001 & -- \\
    {WDMa}  & {Cond.} & {1000} & {0.3166}$~\pm~${0.0018} & {0.0127}$~\pm~${0.0001} & {{0.9431}}$~\pm~${0.0253} & {0.3315}$~\pm~${0.0030} & {0.0133}$~\pm~${0.0001} & {0.9694}\rlap{$^{*}$}$~\pm~${0.0005} \\
    MD   & Uncond. & 1000 & 0.3843$~\pm~$0.0026 & 0.0153$~\pm~$0.0001 & 0.9595$~\pm~$0.0289 & 0.4072$~\pm~$0.0024 & 0.0163$~\pm~$0.0001 & -- \\
    MLDM & Cond.   & 1000 & 0.3590$~\pm~$0.0021 & 0.0144$~\pm~$0.0001 & 0.9538$~\pm~$0.0259 & 0.3733$~\pm~$0.0032 & 0.0150$~\pm~$0.0001 & 0.9784$~\pm~$0.0024 \\
   MOTFM & Uncond.   & 10 & 0.3692$~\pm~$0.0024 & 0.0147$~\pm~$0.0001 & 0.9677$~\pm~$0.0105 & 0.3926$~\pm~$0.0021 & 0.0158$~\pm~$0.0001 & --\\
  MOTFMa & Cond.   & 10 & 0.3747$~\pm~$0.0027 & 0.0150$~\pm~$0.0001 & 0.9529$~\pm~$0.0203 & 0.3539$~\pm~$0.0033 & 0.0142$~\pm~$0.0001 & 0.9775$~\pm~$0.0028\\
  {BS}  & {Cond.} & {--} & {0.3454}$~\pm~${0.0020} & {0.0138}$~\pm~${0.0001} & {\underline{0.9346}}$~\pm~${0.0281} & {0.3495}$~\pm~${0.0035} & {0.0140}$~\pm~${0.0001} & {\underline{0.9600}}$~\pm~${0.0097} \\
    \bottomrule
  \end{tabular}

  \vspace{1mm} % Vertical space between the two tables

  % --- Second Table (Age 40-55 + Age 65-80) ---
  \begin{tabular}{l l c | c c c | c c c}
    \toprule
    \textbf{Method} & \textbf{Type} & \textbf{Steps} &
    \multicolumn{3}{c|}{\textbf{Age 40–55}} &
    \multicolumn{3}{c}{\textbf{Age 65–80}} \\
    & & & \textbf{FID} $\downarrow$ & \textbf{MMD} $\downarrow$ & \textbf{MS-SSIM} $\downarrow$ &
      \textbf{FID} $\downarrow$ & \textbf{MMD} $\downarrow$ & \textbf{MS-SSIM} $\downarrow$ \\
    \midrule
    \multirow{5}{*}{Ours} & \multirow{5}{*}{RFM} & 1 & 0.3511$~\pm~$0.0045 & 0.0140$~\pm~$0.0002 & 0.9966$~\pm~$0.0018 & 0.2955$~\pm~$0.0043 & 0.0118$~\pm~$0.0002 & 0.9958$~\pm~$0.0019 \\
      & & 2 & 0.3409$~\pm~$0.0043 & 0.0136$~\pm~$0.0002 & 0.9927$~\pm~$0.0020 & 0.2844$~\pm~$0.0045 & 0.0113$~\pm~$0.0002 & 0.9911$~\pm~$0.0026 \\
      & & 5 & 0.3303$~\pm~$0.0041 & 0.0132$~\pm~$0.0002 & 0.9838$~\pm~$0.0031 & 0.2743$~\pm~$0.0043 & 0.0109$~\pm~$0.0002 & 0.9812$~\pm~$0.0037 \\
      & & 10 & 0.3169$~\pm~$0.0042 & \underline{0.0127}$~\pm~$0.0002 & 0.9580$~\pm~$0.0129 & 0.2578$~\pm~$0.0040 & 0.0103$~\pm~$0.0002 & 0.9465$~\pm~$0.0145 \\
      & & 200 & \underline{0.3160}\rlap{$^{*}$}$~\pm~$0.0044 & \underline{0.0127}\rlap{$^{*}$}$~\pm~$0.0002 & {0.9560}\rlap{$^{*}$}$~\pm~$0.0138 & \underline{0.2560}\rlap{$^{*}$}$~\pm~$0.0041 & \underline{0.0102}\rlap{$^{*}$}$~\pm~$0.0002 & {0.9433}$~\pm~$0.0155 \\
    \midrule
    \multirow{5}{*}{Ours} & \multirow{5}{*}{CFM} & 1 & 0.3547$~\pm~$0.0039 & 0.0142$~\pm~$0.0002 & 0.9973$~\pm~$0.0009 & 0.2971$~\pm~$0.0042 & 0.0118$~\pm~$0.0002 & 0.9956$~\pm~$0.0021 \\
      & & 2 & 0.3445$~\pm~$0.0043 & 0.0138$~\pm~$0.0002 & 0.9940$~\pm~$0.0010 & 0.2851$~\pm~$0.0046 & 0.0114$~\pm~$0.0002 & 0.9915$~\pm~$0.0027 \\
      & & 5 & 0.3335$~\pm~$0.0044 & 0.0133$~\pm~$0.0002 & 0.9858$~\pm~$0.0018 & 0.2741$~\pm~$0.0036 & 0.0109$~\pm~$0.0002 & 0.9827$~\pm~$0.0033 \\
      & & 10 & 0.3283$~\pm~$0.0038 & 0.0131$~\pm~$0.0002 & 0.9793$~\pm~$0.0030 & 0.2692$~\pm~$0.0035 & 0.0107$~\pm~$0.0001 & 0.9754$~\pm~$0.0043 \\
      & & 200 & 0.3226$~\pm~$0.0041 & 0.0129$~\pm~$0.0002 & {0.9579}\rlap{$^{*}$}$~\pm~$0.0114 & 0.2631$~\pm~$0.0038 & 0.0105$~\pm~$0.0002 & 0.9474\rlap{$^{*}$}$~\pm~$0.0149 \\
    \midrule
    \multirow{5}{*}{Ours} & \multirow{5}{*}{VP} & 1 & 0.3524$~\pm~$0.0044 & 0.0141$~\pm~$0.0002 & {0.9969}$~\pm~$0.0017 & 0.2953$~\pm~$0.0038 & 0.0118$~\pm~$0.0002 & 0.9959$~\pm~$0.0019 \\
      & & 2 & 0.3416$~\pm~$0.0037 & 0.0137$~\pm~$0.0002 & 0.9937$~\pm~$0.0020 & 0.2842$~\pm~$0.0040 & 0.0113$~\pm~$0.0002 & 0.9919$~\pm~$0.0024 \\
      & & 5 & 0.3320$~\pm~$0.0041 & 0.0133$~\pm~$0.0002 & 0.9854$~\pm~$0.0028 & 0.2731$~\pm~$0.0037 & 0.0109$~\pm~$0.0002 & 0.9829$~\pm~$0.0031 \\
      & & 10 & 0.3258$~\pm~$0.0037 & 0.0130$~\pm~$0.0002 & 0.9788$~\pm~$0.0041 & 0.2685$~\pm~$0.0041 & 0.0107$~\pm~$0.0002 & 0.9752$~\pm~$0.0041 \\
      & & 200 & 0.3184\rlap{$^{*}$}$~\pm~$0.0040 & \underline{0.0127}\rlap{$^{*}$}$~\pm~$0.0002 & 0.9584\rlap{$^{*}$}$~\pm~$0.0134 & 0.2596\rlap{$^{*}$}$~\pm~$0.0036 & 0.0104\rlap{$^{*}$}$~\pm~$0.0001 & {0.9472}\rlap{$^{*}$}$~\pm~$0.0151 \\
    \midrule
    \multirow{5}{*}{Ours} & \multirow{5}{*}{Trigon.} & 1 & 0.3484$~\pm~$0.0041 & 0.0139$~\pm~$0.0002 & \textbf{0.9281}$~\pm~$0.0011 & 0.2910$~\pm~$0.0041 & 0.0116$~\pm~$0.0002 & \textbf{0.9239}$~\pm~$0.0011 \\
      & & 2 & 0.3168\rlap{$^{*}$}$~\pm~$0.0041 & 0.0127\rlap{$^{*}$}$~\pm~$0.0002 & 0.9611\rlap{$^{*}$}$~\pm~$0.0015 & 0.2599\rlap{$^{*}$}$~\pm~$0.0039 & 0.0104\rlap{$^{*}$}$~\pm~$0.0002 & 0.9584$~\pm~$0.0016 \\
      & & 5 & {0.3062}$~\pm~$0.0036 & {0.0123}$~\pm~$0.0002 & 0.9789$~\pm~$0.0023 & {0.2473}$~\pm~$0.0042 & {0.0099}$~\pm~$0.0002 & 0.9756$~\pm~$0.0023 \\
      & & 10 & \textbf{0.3042}$~\pm~$0.0041 & \textbf{0.0122}$~\pm~$0.0002 & 0.9758$~\pm~$0.0028 & \textbf{0.2451}$~\pm~$0.0036 & \textbf{0.0098}$~\pm~$0.0001 & 0.9715$~\pm~$0.0031 \\
      & & 200 & 0.3731$~\pm~$0.0049 & 0.0149$~\pm~$0.0002 & 0.9629\rlap{$^{*}$}$~\pm~$0.0115 & 0.3046$~\pm~$0.0049 & 0.0122$~\pm~$0.0002 & 0.9506\rlap{$^{*}$}$~\pm~$0.0131 \\
    \midrule
     & FiLM     & 10 & 0.3455$~\pm~$0.0040 & 0.0138$~\pm~$0.0002 & 0.9860$~\pm~$0.0009 & 0.2855$~\pm~$0.0049 & 0.0114$~\pm~$0.0002 & 0.9862$~\pm~$0.0008 \\
     Ours    & Spatial  & 10 & 0.3450$~\pm~$0.0042 & 0.0138$~\pm~$0.0002 & 0.9861$~\pm~$0.0008 & 0.2809$~\pm~$0.0039 & 0.0112$~\pm~$0.0002 & {0.9874}$~\pm~$0.0007 \\
         & Uncond.  & 10 & 0.3383$~\pm~$0.0026 & 0.0135$~\pm~$0.0001 & -- & 0.2787$~\pm~$0.0017 & 0.0111$~\pm~$0.0001 & -- \\
    \midrule
    WDM  & Uncond. & 1000 & 0.3277$~\pm~$0.0020 & 0.0131$~\pm~$0.0001 & -- & 0.2694$~\pm~$0.0017 & 0.0108$~\pm~$0.0001 & -- \\
    {WDMa}  & {Cond.} & {1000} & {0.3335}$~\pm~${0.0044} & {0.0134}$~\pm~${0.0002} & {{0.9566}}\rlap{$^{*}$}$~\pm~${0.0160} & {0.2825}$~\pm~${0.0039} & {0.0113}$~\pm~${0.0002} & {0.9401}$~\pm~${0.0208} \\
    MD   & Uncond. & 1000 & 0.4074$~\pm~$0.0038 & 0.0163$~\pm~$0.0002 & -- & 0.3426$~\pm~$0.0024 & 0.0137$~\pm~$0.0001 & -- \\
    MLDM & Cond.   & 1000 & 0.3681$~\pm~$0.0043 & 0.0147$~\pm~$0.0002 & 0.9776$~\pm~$0.0034 & 0.3311$~\pm~$0.0054 & 0.0132$~\pm~$0.0002 & 0.9478$~\pm~$0.0242 \\
   MOTFM & Uncond.   & 10 & 0.3914$~\pm~$0.0027 & 0.0157$~\pm~$0.0001 & -- & 0.3269$~\pm~$0.0021 & 0.0131$~\pm~$0.0001 & --\\
  MOTFMa & Cond.   & 10 & 0.4001$~\pm~$0.0048 & 0.0160$~\pm~$0.0002 & 0.9697$~\pm~$0.0068 & 0.3685$~\pm~$0.0049 & 0.0147$~\pm~$0.0002 & 0.9630$~\pm~$0.0101\\
  {BS}  & {Cond.} & {--} & {0.3601}$~\pm~${0.0055} & {0.0144}$~\pm~${0.0002} & {\underline{0.9463}}$~\pm~${0.0286} & {0.3188}$~\pm~${0.0050} & {0.0127}$~\pm~${0.0002} & {\underline{0.9382}}$~\pm~${0.0234} \\
    \bottomrule
  \end{tabular}

  \label{tab:benchmark_merged_all_ages}
\end{center}
\clearpage
\twocolumn

%% file: rebuttal_experiments.tex
\begin{table*}[t]
    \centering
    \caption{Comparison between the two RFM-based low-frequency ablations. The first row corresponds to a FlowLet model trained and sampled using only the approximation band. The second row corresponds to a full-wavelet model reconstructed after zeroing the generated high-frequency bands at inference time.}
    \label{tab:lll_ablation_results_appendix}
    \begingroup
    \scriptsize
    \setlength{\tabcolsep}{1.6mm}
    \resizebox{\textwidth}{!}{%
    \begin{tabular}{@{}lcccccccc@{}}
        \toprule
        & \multicolumn{3}{c}{Global Metrics} & \multicolumn{2}{c}{BAP Test MAE $\downarrow$} & \multicolumn{3}{c}{ROI} \\
        \cmidrule(lr){2-4} \cmidrule(lr){5-6} \cmidrule(lr){7-9}
        Setting & FID $\downarrow$ & MMD $\downarrow$ & MS-SSIM $\downarrow$ & DLBS & \makecell[c]{Held-out\\OpenBHB} & iMAE $\downarrow$ & KLD $\downarrow$ & DICE $\uparrow$ \\
        \midrule
        \makecell[l]{$LLL$-only} & $0.3195 \pm 0.0019$ & $0.0127 \pm 0.0001$ & $0.9656 \pm 0.0134$ & $5.62 \pm 4.44$ & $4.85 \pm 3.14$ & $37.95 \pm 10.14$ & $1.189 \pm 1.094$ & $0.424 \pm 0.172$ \\
        \makecell[l]{HF-zeroed} & $0.3145 \pm 0.0022$ & $0.0126 \pm 0.0001$ & $0.9582 \pm 0.0205$ & $5.32 \pm 4.12$ & $4.54 \pm 3.01$ & $37.38 \pm 10.68$ & $1.190 \pm 1.113$ & $0.428 \pm 0.165$ \\
        \bottomrule
    \end{tabular}%
    }
    \endgroup
\end{table*}

\subsection{Low-frequency ablations}
\label{sec:lll_only_ablation_appendix}
To assess how much of the current evaluation signal can already be explained by coarse low-frequency structure, we considered two complementary RFM-based $LLL$ ablations. In the first setting, FlowLet was trained and sampled using only the approximation band $LLL$, with all seven high-frequency subbands removed throughout both training and inference. In the second setting, the standard full-wavelet FlowLet checkpoints were used, but the generated high-frequency bands were set to zero only at inference time before reconstruction. These two experiments separate the effect of learning exclusively from coarse structure from the effect of suppressing high-frequency content only at reconstruction time.

Table~\ref{tab:lll_ablation_results_appendix} summarizes the results across the full evaluation setup, including global metrics, brain age prediction, and region-based anatomical measures. Both ablations remain competitive under the present evaluation protocol, confirming that coarse anatomical structure carries a substantial fraction of the measurable signal. At the same time, the model trained on full wavelets and evaluated after high-frequency suppression remains consistently stronger than the model trained exclusively on $LLL$. Relative to the dedicated $LLL$-only model, it improves the global metrics ($FID$: $0.3145$ vs.\ $0.3195$; $MMD$: $0.0126$ vs.\ $0.0127$), the external DLBS age-restricted test MAE ($5.32 \pm 4.12$ vs.\ $5.62 \pm 4.44$), the held-out OpenBHB validation test MAE ($4.54 \pm 3.01$ vs.\ $4.85 \pm 3.14$), and the region-based overlap score ($DICE$: $0.428 \pm 0.165$ vs.\ $0.424 \pm 0.172$). These results indicate that coarse structure explains a substantial part of the current metric behavior, but they do not imply that the high-frequency bands are irrelevant. Rather, they show that full-wavelet training still improves the retained coarse-scale reconstruction even when the generated high-frequency coefficients are removed before reconstruction.

\begin{figure*}[t]
    \centering
    \includegraphics[width=0.94\textwidth,height=0.78\textheight,keepaspectratio]{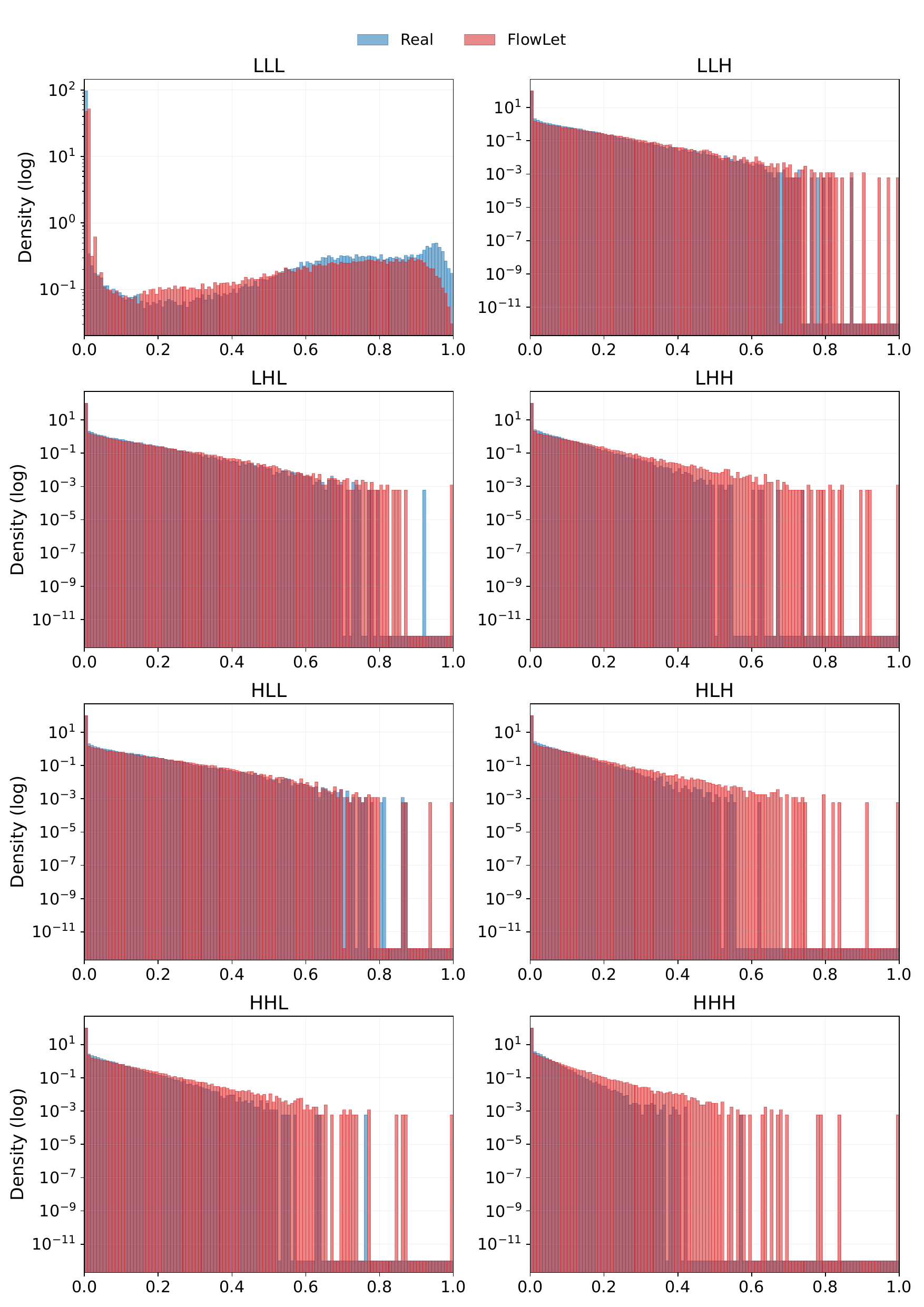}
    \caption{Absolute-value wavelet coefficient histograms for the real cohort and FlowLet, pooled over all subjects and shown for all eight subbands in model space. The detail bands remain populated and do not collapse toward zero. The strongest overlap with the real cohort is observed in the one-high-pass bands ($LLH$, $LHL$, and $HLL$), whereas the mismatch grows in the two-high-pass bands and is largest in the fully diagonal $HHH$ band.}
    \label{fig:wavelet_abs_montage_appendix}
\end{figure*}

\begin{figure*}[t]
    \centering
    \includegraphics[width=0.82\textwidth]{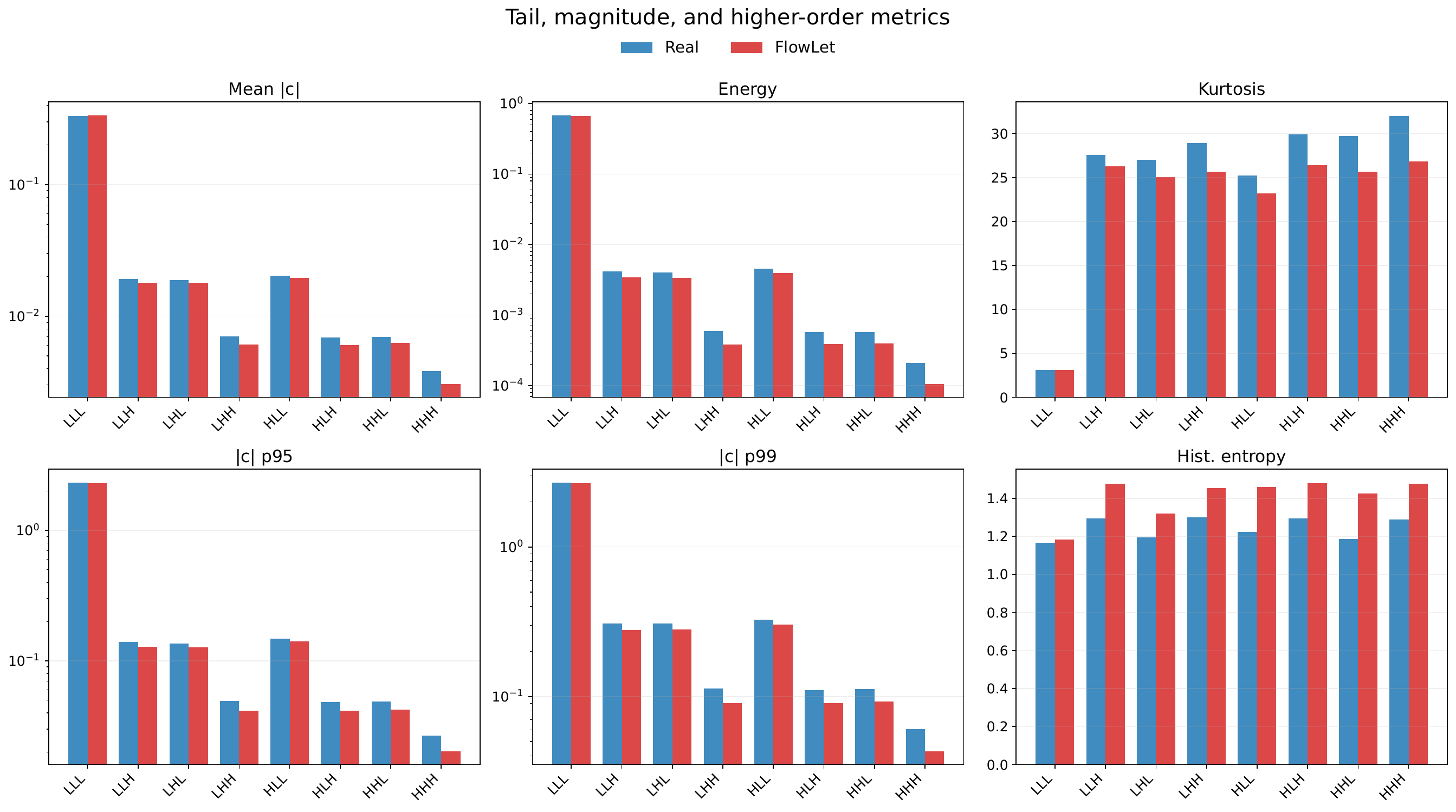}
    \caption{Band-wise comparison of magnitude, tail, and higher-order summary metrics in model space. FlowLet remains close to the real cohort in the one-high-pass bands, while the attenuation becomes stronger in the two-high-pass bands and is largest in $HHH$, where mean $|c|$, energy, upper-tail statistics, and kurtosis are reduced, while histogram entropy is increased.}
    \label{fig:wavelet_tail_metrics_appendix}
\end{figure*}

\subsection{High-frequency wavelet coefficient analysis}
\label{sec:high_frequency_wavelet_analysis_appendix}
We additionally evaluated FlowLet directly in wavelet-coefficient space to characterize how closely the generated subbands match the corresponding real-cohort distributions. The analysis used the same 3D Haar transform employed by the model and compared the combined 5,793 real MRIs against 3,000 generated FlowLet samples in model space. Statistics were computed for all eight bands separately.

The Figure~\ref{fig:wavelet_abs_montage_appendix} shows that the seven detail bands remain populated and do not collapse toward zero. The one-high-pass bands retain the strongest overlap with the real cohort, whereas the two-high-pass and fully diagonal bands become progressively less well matched. The $HHH$ is the sparsest subband, making it the most challenging band in the visual comparison. This pattern indicates that FlowLet retains structured high-frequency content across all detail bands, while the degree of alignment varies naturally with band complexity.

Table~\ref{tab:flowlet_wavelet_summary_compact} reports the exact band-wise statistics as mean $\pm$ standard deviation across volumes, while Figure~\ref{fig:wavelet_tail_metrics_appendix} provides a compact graphical summary of the same quantities on a logarithmic scale to facilitate inspection of small relative differences.

All reported quantities are computed from the wavelet coefficients $c$ in each subband, where $|c|$ denotes their absolute value. Mean $|c|$ describes the coefficient magnitude within a band and summarizes how strongly that band is populated on average. Energy measures the overall signal content by giving greater weight to larger coefficients, making it sensitive to the presence of stronger responses. 
Kurtosis characterizes the sharpness and tail-heaviness of the distribution, helping to distinguish more concentrated coefficient distributions from more diffuse ones. The statistics $|c|$ p95 and $|c|$ p99 denote the 95th and 99th percentiles of the absolute coefficient magnitudes, respectively, and are included to describe the upper tail of the distribution without relying on extreme maxima. Histogram entropy measures how spread the coefficients are across bins, providing a complementary view of distributional diversity. Together, these descriptors allow us to assess both the average strength of each band and the overall shape of its coefficient distribution relative to the real cohort.

These results show that FlowLet preserves the overall wavelet structure, while the higher-frequency statistics shift in a band-dependent way. The approximation band remains close to the real cohort, and the detail bands stay populated throughout the hierarchy. Across the seven detail bands, mean $|c|$ decreases by 9.78\%, energy by 27.98\%, $|c|$ p95 by 12.32\%, and $|c|$ p99 by 15.53\%, while histogram entropy increases by 14.96\%. At the band level, the one-high-pass bands remain closest to the real cohort, whereas the most finely structured band shows the largest change, with mean $|c|$ decreasing by 20.19\%, energy by 49.82\%, $|c|$ p95 by 24.33\%, and $|c|$ p99 by 28.65\%.

These results show that the high-frequency mismatch is neither a simple collapse nor a uniform global shift. The analysis shows that nonzero detail coefficients are retained across all subbands. Figure~\ref{fig:wavelet_tail_metrics_appendix} and Table~\ref{tab:flowlet_wavelet_summary_compact} show that the coefficient typical amplitude, tail strength, and kurtosis remain closer to the real cohort in the easier bands and become progressively more challenging in the more diagonal ones. The higher-order scatter further shows that this is not just a rescaling effect: the generated high-frequency distributions shift toward lower energy, lower kurtosis, and higher histogram entropy, which is consistent with slightly weaker but more diffuse coefficient distributions.

\begin{table*}[t]
\centering
\caption{Summary of overall high-frequency wavelet statistics for real data and FlowLet. Each value is reported as mean $\pm$ standard deviation across volumes.}
\label{tab:flowlet_wavelet_summary_compact}
\begingroup
\footnotesize
\setlength{\tabcolsep}{3.0mm}

\resizebox{\textwidth}{!}{%
\begin{tabular}{lcccccc}
\toprule
& \multicolumn{2}{c}{Mean $|c|$} & \multicolumn{2}{c}{Energy} & \multicolumn{2}{c}{Kurtosis} \\
\cmidrule(lr){2-3}\cmidrule(lr){4-5}\cmidrule(lr){6-7}
Band & Real & FlowLet & Real & FlowLet & Real & FlowLet \\
\midrule
LLH & $0.0192 \pm 0.0017$ & $0.0180 \pm 0.0015$ & $0.0042 \pm 0.0010$ & $0.0034 \pm 0.0007$ & $27.598 \pm \phantom{1}9.454$ & $26.298 \pm 3.474$ \\
LHL & $0.0188 \pm 0.0016$ & $0.0180 \pm 0.0015$ & $0.0040 \pm 0.0009$ & $0.0034 \pm 0.0007$ & $26.995 \pm \phantom{1}6.435$ & $25.058 \pm 3.243$ \\
LHH & $0.0070 \pm 0.0017$ & $0.0061 \pm 0.0015$ & $0.0006 \pm 0.0004$ & $0.0004 \pm 0.0002$ & $28.902 \pm 13.936$ & $25.631 \pm 3.902$ \\
HLL & $0.0202 \pm 0.0016$ & $0.0196 \pm 0.0015$ & $0.0045 \pm 0.0008$ & $0.0040 \pm 0.0007$ & $25.203 \pm \phantom{1}6.642$ & $23.179 \pm 3.205$ \\
HLH & $0.0069 \pm 0.0016$ & $0.0061 \pm 0.0015$ & $0.0006 \pm 0.0003$ & $0.0004 \pm 0.0002$ & $29.905 \pm 17.790$ & $26.409 \pm 3.239$ \\
HHL & $0.0069 \pm 0.0015$ & $0.0063 \pm 0.0015$ & $0.0006 \pm 0.0003$ & $0.0004 \pm 0.0002$ & $29.704 \pm \phantom{1}9.402$ & $25.652 \pm 3.532$ \\
HHH & $0.0038 \pm 0.0018$ & $0.0031 \pm 0.0014$ & $0.0002 \pm 0.0002$ & $0.0001 \pm 0.0001$ & $32.007 \pm 21.929$ & $26.859 \pm 5.031$ \\
\bottomrule
\end{tabular}
}

\resizebox{\textwidth}{!}{%
\begin{tabular}{lcccccc}
\toprule
& \multicolumn{2}{c}{$|c|$ p95} & \multicolumn{2}{c}{$|c|$ p99} & \multicolumn{2}{c}{Hist. entropy} \\
\cmidrule(lr){2-3}\cmidrule(lr){4-5}\cmidrule(lr){6-7}
Band & Real & FlowLet & Real & FlowLet & Real & FlowLet \\
\midrule
LLH & $0.1399 \pm 0.0155$ & $0.1289 \pm 0.0121$ & $0.3071 \pm 0.0414$ & $0.2784 \pm 0.0319$ & $1.293 \pm 0.168$ & $1.476 \pm 0.151$ \\
LHL & $0.1363 \pm 0.0137$ & $0.1273 \pm 0.0111$ & $0.3070 \pm 0.0411$ & $0.2805 \pm 0.0329$ & $1.195 \pm 0.107$ & $1.320 \pm 0.143$ \\
LHH & $0.0496 \pm 0.0119$ & $0.0415 \pm 0.0099$ & $0.1129 \pm 0.0353$ & $0.0905 \pm 0.0250$ & $1.301 \pm 0.168$ & $1.453 \pm 0.154$ \\
HLL & $0.1478 \pm 0.0150$ & $0.1409 \pm 0.0125$ & $0.3256 \pm 0.0336$ & $0.3041 \pm 0.0298$ & $1.224 \pm 0.109$ & $1.459 \pm 0.148$ \\
HLH & $0.0484 \pm 0.0115$ & $0.0417 \pm 0.0102$ & $0.1108 \pm 0.0335$ & $0.0906 \pm 0.0249$ & $1.293 \pm 0.168$ & $1.479 \pm 0.153$ \\
HHL & $0.0486 \pm 0.0108$ & $0.0425 \pm 0.0099$ & $0.1122 \pm 0.0317$ & $0.0927 \pm 0.0243$ & $1.185 \pm 0.118$ & $1.426 \pm 0.144$ \\
HHH & $0.0269 \pm 0.0125$ & $0.0203 \pm 0.0096$ & $0.0605 \pm 0.0327$ & $0.0432 \pm 0.0219$ & $1.288 \pm 0.182$ & $1.477 \pm 0.174$ \\
\bottomrule
\end{tabular}
}
\endgroup
\end{table*}

\begin{figure*}[t]
    \centering
    \includegraphics[width=\textwidth]{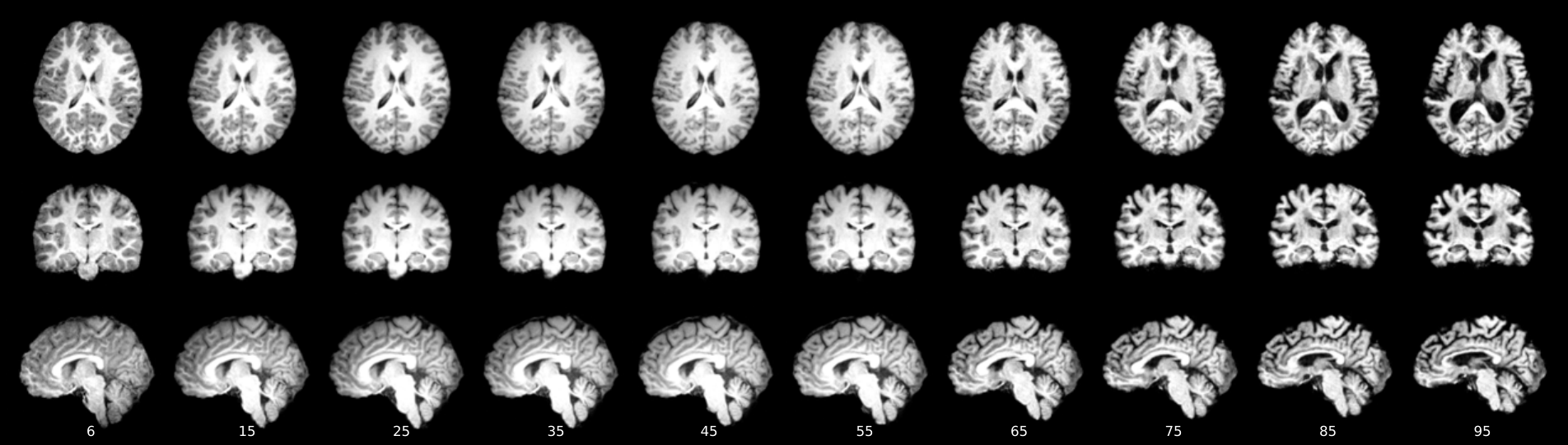}
    \caption{Fixed-seed age-conditioning trajectory generated with FlowLet-RFM. Each column corresponds to a different target age, while the seed is kept fixed across the full sequence. Axial, coronal, and sagittal views are shown for multiple ages.}
    \label{fig:fixed_seed_age_trajectory_appendix}
\end{figure*}

\subsection{Fixed-seed age-conditioning trajectory}
\label{sec:fixed_seed_age_trajectory_appendix}
Figure~\ref{fig:fixed_seed_age_trajectory_appendix} shows the effect of isolated age conditioning. We generated a controlled age trajectory with a single fixed seed (42) and a fixed target-age grid of 6, 15, 25, 35, 45, 55, 65, 75, 85, and 95 years. Only the conditioning variable was changed across columns, while the initialization and sampling steps (10) were held constant. The resulting sequence shows coherent age-dependent morphological variation, including progressive ventricular enlargement, sulcal widening, and cortical thinning in later decades.

\subsection{Dallas Lifespan Brain Study}
\label{sec:dlbs_dataset_appendix}
We further evaluated FlowLet on an external cohort derived from the Dallas Lifespan Brain Study (DLBS), a longitudinal adult-lifespan study designed to jointly characterize brain and cognition across healthy adulthood\footnote{Park et al., \emph{The Dallas Lifespan Brain Study: A Comprehensive Adult Lifespan Data Set of Brain and Cognitive Aging}, \emph{Scientific Data}, 2025. DOI: 10.1038/s41597-025-04847-7.}. For this analysis work, we extracted the T1-weighted structural MRI subset, retained only healthy subjects, and applied the same minimal preprocessing pipeline adopted for the main experiments. The resulting processed subset comprises 956 scans, with mean age $57\pm17$ years, minimum age 21 years, and maximum age 89 years. In this form, DLBS serves as an independent adult-lifespan validation cohort with stronger coverage of later adulthood than OpenBHB, making it a useful complement to the merged training distribution used in the main study. This contrast is illustrated in Figure~\ref{fig:Age_Distribution_external}, where the held-out OpenBHB subset remains concentrated in younger ages, whereas DLBS provides broader support across middle and older adulthood.

\begin{center}
\includegraphics[width=\columnwidth]{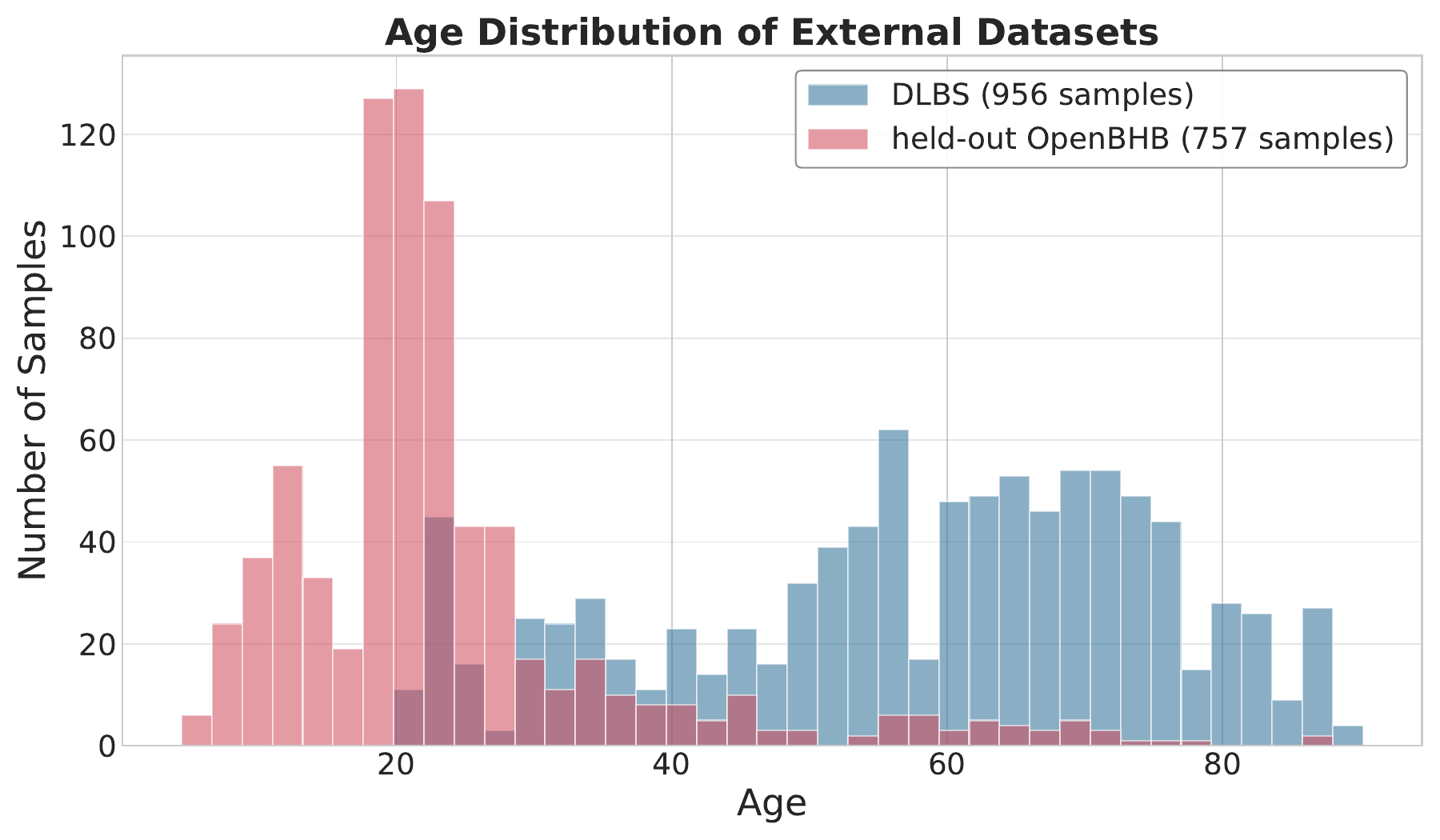}
\captionof{figure}{Age distribution of the external evaluation cohorts. The held-out OpenBHB validation subset is concentrated among younger adults, while the DLBS subset spans the adult lifespan, with greater coverage in middle to late adulthood.}
\label{fig:Age_Distribution_external}
\end{center}

\subsection{Real-to-real global metric references}
\label{sec:real_to_real_references_appendix}
To better contextualize the global similarity metrics, we computed real-to-real references using the same evaluation pipeline adopted for the synthetic comparisons. Specifically, we compared two independent real cohorts against the merged real-cohort reference: (i) the held-out OpenBHB validation subset reserved for downstream testing, and (ii) the processed DLBS subset introduced above. 
The held-out OpenBHB comparison remains close to the merged real cohort overall, while still showing non-zero differences, particularly in the older age bin, which reflects the expected shift between real subsets drawn from a heterogeneous lifespan distribution (Table~\ref{tab:openbhb_real_to_real_appendix}). The DLBS comparison is also close to the merged real cohort (Table~\ref{tab:dlbs_real_to_real_appendix}), indicating that after the same preprocessing and healthy-subject filtering, DLBS occupies a compatible but not identical region of the real-data distribution. Taken together, these real-to-real references provide a practical dataset-level scale for interpreting the synthetic global metrics, showing that non-zero FID, MMD, and MS-SSIM values also arise across genuine real cohorts due to natural biological variability and residual inter-cohort differences. While the synthetic-to-real values remain higher than the real-to-real references, these baselines confirm that non-zero metric values naturally arise even between genuine real cohorts, providing a practical scale for interpreting the reported results.

\begin{table}[t]
    \centering
    \caption{Real-to-real metric reference obtained by comparing the held-out OpenBHB validation subset with the merged real cohort.}
    \label{tab:openbhb_real_to_real_appendix}
    \begingroup
    \small
    \begin{tabular}{lcccc}
        \hline
        Age Group & Samples & FID $\downarrow$ & MMD $\downarrow$ & MS-SSIM $\downarrow$\\
        \hline
        overall & 757 & 0.0069 & 0.0104 & 0.9233 \\
        15--30 & 497 & 0.0037 & 0.0074 & 0.9497 \\
        40--55 & 35 & 0.0030 & 0.0066 & 0.9467 \\
        65--80 & 17 & 0.0186 & 0.0220 & 0.8883 \\
        \hline
    \end{tabular}
    \endgroup
\end{table}

\begin{table}[t]
    \centering
    \caption{Real-to-real global metrics for the DLBS subset versus the merged real cohort.}
    \label{tab:dlbs_real_to_real_appendix}
    \begingroup
    \small
    \begin{tabular}{lcccc}
        \hline
        Age Group & Samples & FID $\downarrow$ & MMD $\downarrow$ & MS-SSIM $\downarrow$\\
        \hline
        overall & 956 & 0.0018 & 0.0014 & 0.9194 \\
        15--30 & 100 & 0.0011 & 0.0037 & 0.9348 \\
        40--55 & 200 & 0.0009 & 0.0027 & 0.9294 \\
        65--80 & 200 & 0.0003 & 0.0031 & 0.9212 \\
        \hline
    \end{tabular}
    \endgroup
\end{table}

\subsection{External DLBS validation}
\label{sec:dlbs_external_validation_appendix}

We next evaluated the downstream augmentation pipelines on DLBS using the same age-restricted BAP setting adopted in the main paper for subjects aged 44 years and older. Results are shown in Table~\ref{tab:dlbs_external_bap_appendix}. On this external cohort, age-conditioned augmentation remained beneficial relative to the real-only baseline. The real-only model reached test MAE $6.38 \pm 4.97$, whereas the FlowLet variants improved this value to $5.24 \pm 4.38$ (RFM), $5.29 \pm 4.26$ (CFM), $5.27 \pm 4.49$ (VP), and $5.26 \pm 4.48$ (Trigonometric). The strongest non-FlowLet baseline among the reported methods was MOTFMa with test MAE $5.53 \pm 4.48$, followed by BrainSynth with $5.55 \pm 4.33$ and WDMa with $5.60 \pm 4.58$. These findings indicate that the benefit of age-conditioned synthetic augmentation relative to the real-only training baseline extends to an independent external adult-lifespan cohort.

\begin{table}[t]
    \centering
    \caption{External DLBS Brain Age Prediction performance for the Age $\geq$ 44 years group. Lower Mean Absolute Error (MAE, in years) = better accuracy.}
    \label{tab:dlbs_external_bap_appendix}
    \vspace{0.5em}
    
    \begingroup
    \setlength{\tabcolsep}{0.5mm}
    \footnotesize
    \begin{tabular}{@{}ll@{\hspace{4mm}}c@{\hspace{3mm}}c@{}}
    \toprule
    & \textbf{Model} & \textbf{Train MAE} $\downarrow$ & \textbf{Test MAE} $\downarrow$ \\
    \midrule
    & Real Data  & $1.15 \pm 1.02$ & $6.38 \pm 4.97$ \\
    \midrule
    \multirow{4}{*}{\rotatebox[]{90}{\scriptsize Ours}}
    & RFM      & $1.46 \pm 0.59$ & $5.24 \pm 4.38$ \\
    & CFM      & $1.39 \pm 0.59$ & $5.29 \pm 4.26$ \\
    & VP       & $1.02 \pm 0.49$ & $5.27 \pm 4.49$ \\
    & Trigon.  & $1.09 \pm 0.48$ & $5.26 \pm 4.48$ \\
    \midrule
    \multirow{3}{*}{\rotatebox[]{90}{\scriptsize Ablat.}}
    & FiLM     & $0.57 \pm 0.51$ & $6.37 \pm 5.06$ \\
    & Spatial  & $0.87 \pm 0.54$ & $5.79 \pm 4.53$ \\
    & Uncond.  & $1.46 \pm 1.06$ & $5.97 \pm 4.83$ \\
    \midrule
    \multirow{6}{*}{\rotatebox[]{90}{\scriptsize Baselines}}
    & WDM      & $1.63 \pm 1.36$ & $7.36 \pm 6.20$ \\
    & {WDMa}   & ${0.33 \pm 0.42}$ & ${5.60 \pm 4.58}$ \\
    & MD       & $2.54 \pm 2.78$ & $5.81 \pm 4.37$ \\
    & MLDM     & $0.98 \pm 0.47$ & $5.97 \pm 4.75$ \\
    & MOTFMa   & $0.85 \pm 0.53$ & $5.53 \pm 4.48$ \\
    & {BS}     & ${0.90 \pm 0.40}$ & ${5.55 \pm 4.33}$ \\
    \bottomrule
    \end{tabular}
    \endgroup
\end{table}

%% file: acknowledgments.tex
\section{Acknowledgement.} This work was partially supported by the following projects:“LIFE: the itaLian system wIde Frailty nEtwork”; DEMETRA: “Development of anensemble learning-based, multidimensional sensory impairment score to predict cognitive impairment in an elderly cohort of Southern Italy” (CUP D99J22001970006)Missione 6/componente 2/Investimento: 2.1 “Rafforzamento e potenziamento della ricerca biomedica del SSN”, funded by European Commission NextGenerationEU; We acknowledge the CINECA award under the ISCRA initiative (Project IsCc1 SynBrain), (Project IsCd1 FlowMRI), (Project IsCd3 EMBRAIN) for the availability of high performance computing resources and support”; This work has been carried out while \textit{Matteo Attimonelli} was enrolled in the Italian National Doctorate on Artificial Intelligence run by Sapienza University of Rome in collaboration with \textit{Politecnico Di Bari}.